\documentclass[11pt]{article}

\usepackage[final]{acl}

\usepackage{times}
\usepackage{latexsym}

\usepackage[T1]{fontenc}
\usepackage[utf8]{inputenc}

\usepackage{microtype}

\usepackage{booktabs}
\usepackage{amsmath}
\usepackage{amsfonts}

\usepackage{inconsolata}

\usepackage{graphicx}

\usepackage{enumitem}
\usepackage{multirow}

\title{CARD: Diagnosing Belief-to-Action Routing Failures\\
in Vision--Language Models}

\author{Souptik Majumdar, Fabian Kögel \and Andreas Bulling \\
  Universität Stuttgart \\
  70569, Stuttgart \\
  Germany \\
  \texttt{souptik.majumdar, fabian.koegel, andreas.bulling} 
  \texttt{@vis.uni-stuttgart.de} \\}

\begin{document}
\maketitle

% AUTO-GENERATED by aggregate_results.py

% --- F4 paired-statement probe held-out accuracies (top-56 mean) ---
% CANONICAL (replaced 2026-05-25, DISJOINT): \newcommand{\probeAccQwenBelief}{85.6}
\newcommand{\probeAccQwenBelief}{79.8}
% CANONICAL (replaced 2026-05-25, DISJOINT): \newcommand{\probeAccQwenBeliefMS}{85.5 \pm 0.2}
\newcommand{\probeAccQwenBeliefMS}{81.5 \pm 1.8}
% CANONICAL (replaced 2026-05-25, DISJOINT): \newcommand{\probeAccQwenIntent}{86.6}
\newcommand{\probeAccQwenIntent}{81.0}
% CANONICAL (replaced 2026-05-25, DISJOINT): \newcommand{\probeAccQwenIntentMS}{86.4 \pm 0.3}
\newcommand{\probeAccQwenIntentMS}{82.9 \pm 2.0}
% CANONICAL (replaced 2026-05-25, DISJOINT): \newcommand{\probeAccQwenEpistemic}{93.9}
\newcommand{\probeAccQwenEpistemic}{92.8}
% CANONICAL (replaced 2026-05-25, DISJOINT): \newcommand{\probeAccQwenEpistemicMS}{94.5 \pm 0.5}
\newcommand{\probeAccQwenEpistemicMS}{93.7 \pm 0.6}
\newcommand{\fOneBeliefTbQwen}{100.0}
% CANONICAL (replaced 2026-05-25, DISJOINT): \newcommand{\fOneBeliefFbQwen}{17.0}
\newcommand{\fOneBeliefFbQwen}{20.0}
% CANONICAL (replaced 2026-05-25, DISJOINT): \newcommand{\probeAccGemmaBelief}{95.2}
\newcommand{\probeAccGemmaBelief}{93.6}
% CANONICAL (replaced 2026-05-25, DISJOINT): \newcommand{\probeAccGemmaBeliefMS}{95.3 \pm 0.1}
\newcommand{\probeAccGemmaBeliefMS}{94.5 \pm 0.6}
% CANONICAL (replaced 2026-05-25, DISJOINT): \newcommand{\probeAccGemmaIntent}{92.6}
\newcommand{\probeAccGemmaIntent}{89.2}
% CANONICAL (replaced 2026-05-25, DISJOINT): \newcommand{\probeAccGemmaIntentMS}{93.1 \pm 0.4}
\newcommand{\probeAccGemmaIntentMS}{90.1 \pm 1.3}
% CANONICAL (replaced 2026-05-25, DISJOINT): \newcommand{\probeAccGemmaEpistemic}{95.7}
\newcommand{\probeAccGemmaEpistemic}{94.7}
% CANONICAL (replaced 2026-05-25, DISJOINT): \newcommand{\probeAccGemmaEpistemicMS}{96.1 \pm 0.3}
\newcommand{\probeAccGemmaEpistemicMS}{95.7 \pm 0.7}
% CANONICAL (replaced 2026-05-25, DISJOINT): \newcommand{\fOneBeliefTbGemma}{83.0}
\newcommand{\fOneBeliefTbGemma}{79.3}
% CANONICAL (replaced 2026-05-25, DISJOINT): \newcommand{\fOneBeliefFbGemma}{28.0}
\newcommand{\fOneBeliefFbGemma}{30.7}
% CANONICAL (replaced 2026-05-25, DISJOINT): \newcommand{\probeAccLlavaBelief}{81.6}
\newcommand{\probeAccLlavaBelief}{78.0}
% CANONICAL (replaced 2026-05-25, DISJOINT): \newcommand{\probeAccLlavaBeliefMS}{81.8 \pm 0.3}
\newcommand{\probeAccLlavaBeliefMS}{78.1 \pm 0.9}
% CANONICAL (replaced 2026-05-25, DISJOINT): \newcommand{\probeAccLlavaIntent}{83.7}
\newcommand{\probeAccLlavaIntent}{78.6}
% CANONICAL (replaced 2026-05-25, DISJOINT): \newcommand{\probeAccLlavaIntentMS}{83.9 \pm 0.5}
\newcommand{\probeAccLlavaIntentMS}{78.8 \pm 0.4}
% CANONICAL (replaced 2026-05-25, DISJOINT): \newcommand{\probeAccLlavaEpistemic}{91.9}
\newcommand{\probeAccLlavaEpistemic}{87.9}
% CANONICAL (replaced 2026-05-25, DISJOINT): \newcommand{\probeAccLlavaEpistemicMS}{93.0 \pm 0.8}
\newcommand{\probeAccLlavaEpistemicMS}{88.6 \pm 0.6}
% CANONICAL (replaced 2026-05-25, DISJOINT): \newcommand{\fOneBeliefTbLlava}{83.0}
\newcommand{\fOneBeliefTbLlava}{80.0}
% CANONICAL (replaced 2026-05-25, DISJOINT): \newcommand{\fOneBeliefFbLlava}{17.0}
\newcommand{\fOneBeliefFbLlava}{20.0}
% CANONICAL (replaced 2026-05-25, DISJOINT): \newcommand{\probeAccInternBelief}{93.9}
\newcommand{\probeAccInternBelief}{92.7}
% CANONICAL (replaced 2026-05-25, DISJOINT): \newcommand{\probeAccInternBeliefMS}{94.3 \pm 0.3}
\newcommand{\probeAccInternBeliefMS}{93.3 \pm 0.5}
% CANONICAL (replaced 2026-05-25, DISJOINT): \newcommand{\probeAccInternIntent}{92.3}
\newcommand{\probeAccInternIntent}{90.3}
% CANONICAL (replaced 2026-05-25, DISJOINT): \newcommand{\probeAccInternIntentMS}{92.8 \pm 0.4}
\newcommand{\probeAccInternIntentMS}{90.7 \pm 0.6}
% CANONICAL (replaced 2026-05-25, DISJOINT): \newcommand{\probeAccInternEpistemic}{95.1}
\newcommand{\probeAccInternEpistemic}{94.1}
% CANONICAL (replaced 2026-05-25, DISJOINT): \newcommand{\probeAccInternEpistemicMS}{95.4 \pm 0.2}
\newcommand{\probeAccInternEpistemicMS}{95.2 \pm 0.8}
% CANONICAL (replaced 2026-05-25, DISJOINT): \newcommand{\fOneBeliefTbIntern}{86.3}
\newcommand{\fOneBeliefTbIntern}{89.3}
% CANONICAL (replaced 2026-05-25, DISJOINT): \newcommand{\fOneBeliefFbIntern}{1.3}
\newcommand{\fOneBeliefFbIntern}{2.0}

% --- F4 CARD intent column max |gap shift| across probes ---
\newcommand{\cardIntentMaxQwenALeverOne}{0.0}
% CANONICAL (replaced 2026-05-25, DISJOINT): \newcommand{\cardIntentMaxQwenALeverTwo}{24.0}
\newcommand{\cardIntentMaxQwenALeverTwo}{38.7}
% CANONICAL (replaced 2026-05-25, DISJOINT): \newcommand{\cardIntentMaxQwenATrav}{1.3}
\newcommand{\cardIntentMaxQwenATrav}{6.7}
% CANONICAL (replaced 2026-05-25, DISJOINT): \newcommand{\cardIntentMaxQwenOverall}{8.6}
\newcommand{\cardIntentMaxQwenOverall}{10.8}
% CANONICAL (replaced 2026-05-25, DISJOINT): \newcommand{\cardIntentMaxGemmaALeverOne}{0.0}
\newcommand{\cardIntentMaxGemmaALeverOne}{9.7}
% CANONICAL (replaced 2026-05-25, DISJOINT): \newcommand{\cardIntentMaxGemmaALeverTwo}{54.7}
\newcommand{\cardIntentMaxGemmaALeverTwo}{66.7}
% CANONICAL (replaced 2026-05-25, DISJOINT): \newcommand{\cardIntentMaxGemmaATrav}{55.3}
\newcommand{\cardIntentMaxGemmaATrav}{80.0}
% CANONICAL (replaced 2026-05-25, DISJOINT): \newcommand{\cardIntentMaxGemmaOverall}{37.2}
\newcommand{\cardIntentMaxGemmaOverall}{52.7}
\newcommand{\cardIntentMaxLlavaALeverOne}{0.0}
\newcommand{\cardIntentMaxLlavaALeverTwo}{0.0}
% CANONICAL (replaced 2026-05-25, DISJOINT): \newcommand{\cardIntentMaxLlavaATrav}{65.3}
\newcommand{\cardIntentMaxLlavaATrav}{53.3}
% CANONICAL (replaced 2026-05-25, DISJOINT): \newcommand{\cardIntentMaxLlavaOverall}{22.1}
\newcommand{\cardIntentMaxLlavaOverall}{18.0}
% CANONICAL (replaced 2026-05-25, DISJOINT): \newcommand{\cardIntentMaxInternALeverOne}{10.4}
\newcommand{\cardIntentMaxInternALeverOne}{25.0}
% CANONICAL (replaced 2026-05-25, DISJOINT): \newcommand{\cardIntentMaxInternALeverTwo}{1.3}
\newcommand{\cardIntentMaxInternALeverTwo}{6.7}
% CANONICAL (replaced 2026-05-25, DISJOINT): \newcommand{\cardIntentMaxInternATrav}{4.7}
\newcommand{\cardIntentMaxInternATrav}{17.3}
% CANONICAL (replaced 2026-05-25, DISJOINT): \newcommand{\cardIntentMaxInternOverall}{2.9}
\newcommand{\cardIntentMaxInternOverall}{16.2}

% --- D2 four-cell discrimination matrix (success-TB / success-FB / counter-TB / counter-FB) ---
\newcommand{\dTwoQwenBeliefALeverOne}{100/0/100/100}
\newcommand{\dTwoQwenBeliefALeverTwo}{100/50/100/100}
\newcommand{\dTwoQwenBeliefATrav}{100/0/100/100}
\newcommand{\dTwoQwenIntentALeverOne}{0/100/100/100}
\newcommand{\dTwoQwenIntentALeverTwo}{0/100/100/100}
\newcommand{\dTwoQwenIntentATrav}{53/47/100/100}
\newcommand{\dTwoQwenEpistemicALeverOne}{100/100/48/100}
\newcommand{\dTwoQwenEpistemicALeverTwo}{100/69/72/100}
\newcommand{\dTwoQwenEpistemicATrav}{53/100/44/98}
\newcommand{\dTwoGemmaBeliefALeverOne}{100/0/100/100}
\newcommand{\dTwoGemmaBeliefALeverTwo}{69/52/100/100}
\newcommand{\dTwoGemmaBeliefATrav}{100/4/100/100}
\newcommand{\dTwoGemmaIntentALeverOne}{0/100/100/100}
\newcommand{\dTwoGemmaIntentALeverTwo}{0/100/100/100}
\newcommand{\dTwoGemmaIntentATrav}{100/0/100/100}
\newcommand{\dTwoGemmaEpistemicALeverOne}{100/100/59/100}
\newcommand{\dTwoGemmaEpistemicALeverTwo}{100/100/58/100}
\newcommand{\dTwoGemmaEpistemicATrav}{0/100/44/100}
\newcommand{\dTwoLlavaBeliefALeverOne}{100/0/57/52}
\newcommand{\dTwoLlavaBeliefALeverTwo}{100/0/10/42}
\newcommand{\dTwoLlavaBeliefATrav}{53/47/100/100}
\newcommand{\dTwoLlavaIntentALeverOne}{0/100/100/100}
\newcommand{\dTwoLlavaIntentALeverTwo}{0/100/100/100}
\newcommand{\dTwoLlavaIntentATrav}{53/47/100/100}
\newcommand{\dTwoLlavaEpistemicALeverOne}{89/51/48/52}
\newcommand{\dTwoLlavaEpistemicALeverTwo}{100/52/100/42}
\newcommand{\dTwoLlavaEpistemicATrav}{53/47/44/56}
\newcommand{\dTwoInternBeliefALeverOne}{84/0/95/100}  % preserved from previous run
\newcommand{\dTwoInternBeliefALeverTwo}{75/4/100/100}  % preserved from previous run
\newcommand{\dTwoInternBeliefATrav}{100/0/94/98}  % preserved from previous run
\newcommand{\dTwoInternIntentALeverOne}{0/100/100/100}  % preserved from previous run
\newcommand{\dTwoInternIntentALeverTwo}{0/100/100/100}  % preserved from previous run
\newcommand{\dTwoInternIntentATrav}{75/23/98/100}  % preserved from previous run
\newcommand{\dTwoInternEpistemicALeverOne}{99/89/0/93}  % preserved from previous run
\newcommand{\dTwoInternEpistemicALeverTwo}{98/87/0/92}  % preserved from previous run
\newcommand{\dTwoInternEpistemicATrav}{2/100/0/100}  % preserved from previous run

% --- T-A V0 baseline on F4 (TB/FB acc, lever-holders) ---
\newcommand{\taVZeroQwenALeverOne}{30.6/100.0}
\newcommand{\taVZeroQwenALeverTwo}{33.3/100.0}
\newcommand{\taVZeroQwenATrav}{66.0/67.3}
\newcommand{\taVZeroGemmaALeverOne}{30.6/100.0}
\newcommand{\taVZeroGemmaALeverTwo}{33.3/100.0}
\newcommand{\taVZeroGemmaATrav}{100.0/33.3}
\newcommand{\taVZeroLlavaALeverOne}{30.6/100.0}
\newcommand{\taVZeroLlavaALeverTwo}{33.3/100.0}
\newcommand{\taVZeroLlavaATrav}{66.0/67.3}
\newcommand{\taVZeroInternALeverOne}{30.6/100.0}
\newcommand{\taVZeroInternALeverTwo}{33.3/100.0}
\newcommand{\taVZeroInternATrav}{83.3/48.7}

% --- T-A pre-registered falsifier ---
\newcommand{\taLeverHolderCells}{5}
\newcommand{\taTotalCells}{192}

% --- Projection-removal: |gap(α=5σ) - gap(α=0.5σ)| on lever-holders ---
\newcommand{\projRemovalDeltaQwenALeverOne}{0.00}
\newcommand{\projRemovalDeltaQwenALeverTwo}{0.00}
\newcommand{\projRemovalDeltaQwenATrav}{0.00}
\newcommand{\projRemovalDeltaQwenOverall}{0.00}
\newcommand{\projRemovalDeltaGemmaALeverOne}{0.00}
\newcommand{\projRemovalDeltaGemmaALeverTwo}{0.00}
\newcommand{\projRemovalDeltaGemmaATrav}{0.00}
\newcommand{\projRemovalDeltaGemmaOverall}{0.00}
\newcommand{\projRemovalDeltaLlavaALeverOne}{0.00}
\newcommand{\projRemovalDeltaLlavaALeverTwo}{0.00}
\newcommand{\projRemovalDeltaLlavaATrav}{0.00}
\newcommand{\projRemovalDeltaLlavaOverall}{0.00}
\newcommand{\projRemovalDeltaInternALeverOne}{TBD}
\newcommand{\projRemovalDeltaInternALeverTwo}{TBD}
\newcommand{\projRemovalDeltaInternATrav}{TBD}
\newcommand{\projRemovalDeltaInternOverall}{TBD}

% --- D1 polarity-flip: orig vs flipped gap (overall) ---
\newcommand{\dOneOrigQwen}{+86.5}
\newcommand{\dOneFlipQwen}{+82.4}
\newcommand{\dOneOrigGemma}{+73.0}
\newcommand{\dOneFlipGemma}{+85.1}
\newcommand{\dOneOrigLlava}{+63.5}
\newcommand{\dOneFlipLlava}{+35.1}
\newcommand{\dOneOrigIntern}{+81.1}
\newcommand{\dOneFlipIntern}{N/A}

% --- Dataset counts and probe orthogonality ---
\newcommand{\nSuccess}{300}
\newcommand{\nCounter}{144}
\newcommand{\nTotalFour}{444}
\newcommand{\nHoldoutF}{148}
% CANONICAL (replaced 2026-05-25, DISJOINT): \newcommand{\cosBeliefIntentQwen}{0.47}
\newcommand{\cosBeliefIntentQwen}{0.55}
% CANONICAL (replaced 2026-05-25, DISJOINT): \newcommand{\cosBeliefIntentGemma}{0.36}
\newcommand{\cosBeliefIntentGemma}{0.38}
% CANONICAL (replaced 2026-05-25, DISJOINT): \newcommand{\cosBeliefIntentLlava}{0.49}
\newcommand{\cosBeliefIntentLlava}{0.57}
% CANONICAL (replaced 2026-05-25, DISJOINT): \newcommand{\cosBeliefIntentIntern}{0.24}
\newcommand{\cosBeliefIntentIntern}{0.28}
% CANONICAL (replaced 2026-05-25, DISJOINT): \newcommand{\meanOffDiagCos}{0.21--0.55}
\newcommand{\meanOffDiagCos}{0.26--0.58}

% --- Aliases for existing section references (backward compat) ---
\newcommand{\beliefProbeQwen}{\probeAccQwenBelief}
\newcommand{\beliefProbeGemma}{\probeAccGemmaBelief}
\newcommand{\beliefProbeLlava}{\probeAccLlavaBelief}
\newcommand{\beliefGapQwen}{\dOneOrigQwen}
\newcommand{\beliefGapGemma}{\dOneOrigGemma}
\newcommand{\beliefGapLlava}{\dOneOrigLlava}
\newcommand{\dTwoQwenBelief}{\dTwoQwenBeliefALeverOne}
\newcommand{\dTwoGemmaBelief}{\dTwoGemmaBeliefALeverOne}
\newcommand{\dTwoLlavaBelief}{\dTwoLlavaBeliefALeverOne}
\newcommand{\dTwoQwenIntent}{\dTwoQwenIntentALeverOne}
\newcommand{\dTwoGemmaIntent}{\dTwoGemmaIntentALeverOne}
\newcommand{\dTwoLlavaIntent}{\dTwoLlavaIntentALeverOne}
\newcommand{\cardIntentMaxQwen}{\cardIntentMaxQwenALeverOne}
\newcommand{\cardIntentMaxGemma}{\cardIntentMaxGemmaALeverOne}
\newcommand{\cardIntentMaxLlava}{\cardIntentMaxLlavaALeverOne}
\newcommand{\projRemovalQwenAlphaFive}{\projRemovalDeltaQwenALeverOne}
\newcommand{\projRemovalGemmaAlphaFive}{\projRemovalDeltaGemmaALeverOne}
\newcommand{\projRemovalLlavaAlphaFive}{\projRemovalDeltaLlavaALeverOne}
\newcommand{\taVZeroQwenA}{\taVZeroQwenALeverOne}
\newcommand{\taVZeroGemmaA}{\taVZeroGemmaALeverOne}
\newcommand{\taVZeroLlavaA}{\taVZeroLlavaALeverOne}

% --- Multi-seed CARD intent-column summary (seeds [42, 43, 44]) ---
\newcommand{\cardIntentMultiseedTotalCells}{108}
% CANONICAL (replaced 2026-05-25, DISJOINT): \newcommand{\cardIntentMultiseedZeroCells}{98}
\newcommand{\cardIntentMultiseedZeroCells}{95}
% CANONICAL (replaced 2026-05-25, DISJOINT): \newcommand{\cardIntentMultiseedMaxGain}{+2.67}
\newcommand{\cardIntentMultiseedMaxGain}{+12.00}
\newcommand{\cardIntentMultiseedNSeeds}{3}

% --- tab:card_breakdown Δ_gap / Δ_acc per (VLM, probe, question) on a1 ---
% CANONICAL (replaced 2026-05-25, DISJOINT): \newcommand{\cardBreakQwenBeliefBeliefGap}{$+34.72$}
\newcommand{\cardBreakQwenBeliefBeliefGap}{$+30.56$}
% CANONICAL (replaced 2026-05-25, DISJOINT): \newcommand{\cardBreakQwenBeliefBeliefAcc}{$-17.36$}
\newcommand{\cardBreakQwenBeliefBeliefAcc}{$-15.28$}
% CANONICAL (replaced 2026-05-25, DISJOINT): \newcommand{\cardBreakQwenIntentBeliefGap}{$+1.39$}
\newcommand{\cardBreakQwenIntentBeliefGap}{$-4.17$}
% CANONICAL (replaced 2026-05-25, DISJOINT): \newcommand{\cardBreakQwenIntentBeliefAcc}{$-0.69$}
\newcommand{\cardBreakQwenIntentBeliefAcc}{$+2.08$}
% CANONICAL (replaced 2026-05-25, DISJOINT): \newcommand{\cardBreakQwenEpistBeliefGap}{$+8.33$}
\newcommand{\cardBreakQwenEpistBeliefGap}{$+6.94$}
% CANONICAL (replaced 2026-05-25, DISJOINT): \newcommand{\cardBreakQwenEpistBeliefAcc}{$-4.17$}
\newcommand{\cardBreakQwenEpistBeliefAcc}{$-3.47$}
\newcommand{\cardBreakQwenBeliefIntentGap}{$+0.00$}
\newcommand{\cardBreakQwenBeliefIntentAcc}{$+0.00$}
\newcommand{\cardBreakQwenIntentIntentGap}{$+0.00$}
\newcommand{\cardBreakQwenIntentIntentAcc}{$+0.00$}
\newcommand{\cardBreakQwenEpistIntentGap}{$+0.00$}
\newcommand{\cardBreakQwenEpistIntentAcc}{$+0.00$}
% CANONICAL (replaced 2026-05-25, DISJOINT): \newcommand{\cardBreakQwenBeliefEpistGap}{$+1.39$}
\newcommand{\cardBreakQwenBeliefEpistGap}{$+22.22$}
% CANONICAL (replaced 2026-05-25, DISJOINT): \newcommand{\cardBreakQwenBeliefEpistAcc}{$+0.69$}
\newcommand{\cardBreakQwenBeliefEpistAcc}{$+11.11$}
\newcommand{\cardBreakQwenIntentEpistGap}{$+0.00$}
\newcommand{\cardBreakQwenIntentEpistAcc}{$+0.00$}
\newcommand{\cardBreakQwenEpistEpistGap}{$+0.00$}
\newcommand{\cardBreakQwenEpistEpistAcc}{$+0.00$}
% CANONICAL (replaced 2026-05-25, DISJOINT): \newcommand{\cardBreakGemmaBeliefBeliefGap}{$+70.14$}
\newcommand{\cardBreakGemmaBeliefBeliefGap}{$+66.67$}
% CANONICAL (replaced 2026-05-25, DISJOINT): \newcommand{\cardBreakGemmaBeliefBeliefAcc}{$-0.35$}
\newcommand{\cardBreakGemmaBeliefBeliefAcc}{$+0.00$}
% CANONICAL (replaced 2026-05-25, DISJOINT): \newcommand{\cardBreakGemmaIntentBeliefGap}{$+35.42$}
\newcommand{\cardBreakGemmaIntentBeliefGap}{$+33.33$}
% CANONICAL (replaced 2026-05-25, DISJOINT): \newcommand{\cardBreakGemmaIntentBeliefAcc}{$-17.71$}
\newcommand{\cardBreakGemmaIntentBeliefAcc}{$-16.67$}
% CANONICAL (replaced 2026-05-25, DISJOINT): \newcommand{\cardBreakGemmaEpistBeliefGap}{$+2.78$}
\newcommand{\cardBreakGemmaEpistBeliefGap}{$+1.39$}
% CANONICAL (replaced 2026-05-25, DISJOINT): \newcommand{\cardBreakGemmaEpistBeliefAcc}{$-1.39$}
\newcommand{\cardBreakGemmaEpistBeliefAcc}{$-0.69$}
% CANONICAL (replaced 2026-05-25, DISJOINT): \newcommand{\cardBreakGemmaBeliefIntentGap}{$+0.00$}
\newcommand{\cardBreakGemmaBeliefIntentGap}{$-9.72$}
% CANONICAL (replaced 2026-05-25, DISJOINT): \newcommand{\cardBreakGemmaBeliefIntentAcc}{$+0.00$}
\newcommand{\cardBreakGemmaBeliefIntentAcc}{$+4.86$}
\newcommand{\cardBreakGemmaIntentIntentGap}{$+0.00$}
\newcommand{\cardBreakGemmaIntentIntentAcc}{$+0.00$}
\newcommand{\cardBreakGemmaEpistIntentGap}{$+0.00$}
\newcommand{\cardBreakGemmaEpistIntentAcc}{$+0.00$}
% CANONICAL (replaced 2026-05-25, DISJOINT): \newcommand{\cardBreakGemmaBeliefEpistGap}{$+21.53$}
\newcommand{\cardBreakGemmaBeliefEpistGap}{$+59.72$}
% CANONICAL (replaced 2026-05-25, DISJOINT): \newcommand{\cardBreakGemmaBeliefEpistAcc}{$-16.32$}
\newcommand{\cardBreakGemmaBeliefEpistAcc}{$-20.14$}
% CANONICAL (replaced 2026-05-25, DISJOINT): \newcommand{\cardBreakGemmaIntentEpistGap}{$+34.03$}
\newcommand{\cardBreakGemmaIntentEpistGap}{$+22.22$}
% CANONICAL (replaced 2026-05-25, DISJOINT): \newcommand{\cardBreakGemmaIntentEpistAcc}{$-7.99$}
\newcommand{\cardBreakGemmaIntentEpistAcc}{$-19.44$}
% CANONICAL (replaced 2026-05-25, DISJOINT): \newcommand{\cardBreakGemmaEpistEpistGap}{$+21.53$}
\newcommand{\cardBreakGemmaEpistEpistGap}{$+12.50$}
% CANONICAL (replaced 2026-05-25, DISJOINT): \newcommand{\cardBreakGemmaEpistEpistAcc}{$+4.51$}
\newcommand{\cardBreakGemmaEpistEpistAcc}{$+6.25$}
% CANONICAL (replaced 2026-05-25, DISJOINT): \newcommand{\cardBreakLlavaBeliefBeliefGap}{$+41.67$}
\newcommand{\cardBreakLlavaBeliefBeliefGap}{$+20.83$}
% CANONICAL (replaced 2026-05-25, DISJOINT): \newcommand{\cardBreakLlavaBeliefBeliefAcc}{$-15.97$}
\newcommand{\cardBreakLlavaBeliefBeliefAcc}{$-4.86$}
% CANONICAL (replaced 2026-05-25, DISJOINT): \newcommand{\cardBreakLlavaIntentBeliefGap}{$-0.00$}
\newcommand{\cardBreakLlavaIntentBeliefGap}{$+8.33$}
% CANONICAL (replaced 2026-05-25, DISJOINT): \newcommand{\cardBreakLlavaIntentBeliefAcc}{$-14.58$}
\newcommand{\cardBreakLlavaIntentBeliefAcc}{$-13.89$}
% CANONICAL (replaced 2026-05-25, DISJOINT): \newcommand{\cardBreakLlavaEpistBeliefGap}{$-15.97$}
\newcommand{\cardBreakLlavaEpistBeliefGap}{$-12.50$}
% CANONICAL (replaced 2026-05-25, DISJOINT): \newcommand{\cardBreakLlavaEpistBeliefAcc}{$+6.60$}
\newcommand{\cardBreakLlavaEpistBeliefAcc}{$+11.81$}
\newcommand{\cardBreakLlavaBeliefIntentGap}{$+0.00$}
\newcommand{\cardBreakLlavaBeliefIntentAcc}{$+0.00$}
\newcommand{\cardBreakLlavaIntentIntentGap}{$+0.00$}
\newcommand{\cardBreakLlavaIntentIntentAcc}{$+0.00$}
\newcommand{\cardBreakLlavaEpistIntentGap}{$+0.00$}
\newcommand{\cardBreakLlavaEpistIntentAcc}{$+0.00$}
% CANONICAL (replaced 2026-05-25, DISJOINT): \newcommand{\cardBreakLlavaBeliefEpistGap}{$+11.11$}
\newcommand{\cardBreakLlavaBeliefEpistGap}{$-13.89$}
% CANONICAL (replaced 2026-05-25, DISJOINT): \newcommand{\cardBreakLlavaBeliefEpistAcc}{$+39.58$}
\newcommand{\cardBreakLlavaBeliefEpistAcc}{$+40.28$}
% CANONICAL (replaced 2026-05-25, DISJOINT): \newcommand{\cardBreakLlavaIntentEpistGap}{$+38.89$}
\newcommand{\cardBreakLlavaIntentEpistGap}{$+9.72$}
% CANONICAL (replaced 2026-05-25, DISJOINT): \newcommand{\cardBreakLlavaIntentEpistAcc}{$-13.89$}
\newcommand{\cardBreakLlavaIntentEpistAcc}{$-14.58$}
% CANONICAL (replaced 2026-05-25, DISJOINT): \newcommand{\cardBreakLlavaEpistEpistGap}{$-18.06$}
\newcommand{\cardBreakLlavaEpistEpistGap}{$-23.61$}
% CANONICAL (replaced 2026-05-25, DISJOINT): \newcommand{\cardBreakLlavaEpistEpistAcc}{$+19.44$}
\newcommand{\cardBreakLlavaEpistEpistAcc}{$+20.14$}
% CANONICAL (replaced 2026-05-25, DISJOINT): \newcommand{\cardBreakInternBeliefBeliefGap}{$-29.17$}
\newcommand{\cardBreakInternBeliefBeliefGap}{$-31.94$}
% CANONICAL (replaced 2026-05-25, DISJOINT): \newcommand{\cardBreakInternBeliefBeliefAcc}{$-15.97$}
\newcommand{\cardBreakInternBeliefBeliefAcc}{$-20.14$}
\newcommand{\cardBreakInternIntentBeliefGap}{$+0.00$}
\newcommand{\cardBreakInternIntentBeliefAcc}{$+0.00$}
% CANONICAL (replaced 2026-05-25, DISJOINT): \newcommand{\cardBreakInternEpistBeliefGap}{$+11.81$}
\newcommand{\cardBreakInternEpistBeliefGap}{$+31.94$}
% CANONICAL (replaced 2026-05-25, DISJOINT): \newcommand{\cardBreakInternEpistBeliefAcc}{$+5.90$}
\newcommand{\cardBreakInternEpistBeliefAcc}{$+20.14$}
% CANONICAL (replaced 2026-05-25, DISJOINT): \newcommand{\cardBreakInternBeliefIntentGap}{$+10.42$}
\newcommand{\cardBreakInternBeliefIntentGap}{$+8.33$}
% CANONICAL (replaced 2026-05-25, DISJOINT): \newcommand{\cardBreakInternBeliefIntentAcc}{$-5.21$}
\newcommand{\cardBreakInternBeliefIntentAcc}{$-5.56$}
\newcommand{\cardBreakInternIntentIntentGap}{$+0.00$}
\newcommand{\cardBreakInternIntentIntentAcc}{$+0.00$}
% CANONICAL (replaced 2026-05-25, DISJOINT): \newcommand{\cardBreakInternEpistIntentGap}{$-0.69$}
\newcommand{\cardBreakInternEpistIntentGap}{$-25.00$}
% CANONICAL (replaced 2026-05-25, DISJOINT): \newcommand{\cardBreakInternEpistIntentAcc}{$+0.35$}
\newcommand{\cardBreakInternEpistIntentAcc}{$+13.89$}
% CANONICAL (replaced 2026-05-25, DISJOINT): \newcommand{\cardBreakInternBeliefEpistGap}{$-2.08$}
\newcommand{\cardBreakInternBeliefEpistGap}{$+2.78$}
% CANONICAL (replaced 2026-05-25, DISJOINT): \newcommand{\cardBreakInternBeliefEpistAcc}{$-17.01$}
\newcommand{\cardBreakInternBeliefEpistAcc}{$-26.39$}
% CANONICAL (replaced 2026-05-25, DISJOINT): \newcommand{\cardBreakInternIntentEpistGap}{$-0.00$}
\newcommand{\cardBreakInternIntentEpistGap}{$-2.78$}
% CANONICAL (replaced 2026-05-25, DISJOINT): \newcommand{\cardBreakInternIntentEpistAcc}{$+0.69$}
\newcommand{\cardBreakInternIntentEpistAcc}{$+2.78$}
% CANONICAL (replaced 2026-05-25, DISJOINT): \newcommand{\cardBreakInternEpistEpistGap}{$+4.86$}
\newcommand{\cardBreakInternEpistEpistGap}{$-4.17$}
% CANONICAL (replaced 2026-05-25, DISJOINT): \newcommand{\cardBreakInternEpistEpistAcc}{$+1.74$}
\newcommand{\cardBreakInternEpistEpistAcc}{$+17.36$}

% --- Per-cell intent Δacc (best-α) multi-seed mean ± std ---
\newcommand{\intentDeltaQwenALeverOneBelief}{+0.0}
\newcommand{\intentDeltaQwenALeverOneIntent}{+0.0}
\newcommand{\intentDeltaQwenALeverOneEpist}{+0.0}
\newcommand{\intentDeltaQwenALeverTwoBelief}{+0.0}
\newcommand{\intentDeltaQwenALeverTwoIntent}{+0.0}
\newcommand{\intentDeltaQwenALeverTwoEpist}{+0.0}
% CANONICAL (replaced 2026-05-25, DISJOINT): \newcommand{\intentDeltaQwenATravBelief}{+0.2 \pm 0.3}
\newcommand{\intentDeltaQwenATravBelief}{+4.0 \pm 2.9}
\newcommand{\intentDeltaQwenATravIntent}{+0.0}
\newcommand{\intentDeltaQwenATravEpist}{+0.0}
\newcommand{\intentDeltaGemmaALeverOneBelief}{+0.0}
\newcommand{\intentDeltaGemmaALeverOneIntent}{+0.0}
\newcommand{\intentDeltaGemmaALeverOneEpist}{+0.0}
\newcommand{\intentDeltaGemmaALeverTwoBelief}{+0.0}
\newcommand{\intentDeltaGemmaALeverTwoIntent}{+0.0}
\newcommand{\intentDeltaGemmaALeverTwoEpist}{+0.0}
% CANONICAL (replaced 2026-05-25, DISJOINT): \newcommand{\intentDeltaGemmaATravBelief}{+0.3 \pm 0.3}
\newcommand{\intentDeltaGemmaATravBelief}{+0.0}
\newcommand{\intentDeltaGemmaATravIntent}{+0.0}
\newcommand{\intentDeltaGemmaATravEpist}{+0.0}
\newcommand{\intentDeltaLlavaALeverOneBelief}{+0.0}
\newcommand{\intentDeltaLlavaALeverOneIntent}{+0.0}
\newcommand{\intentDeltaLlavaALeverOneEpist}{+0.0}
\newcommand{\intentDeltaLlavaALeverTwoBelief}{+0.0}
% CANONICAL (replaced 2026-05-25, DISJOINT): \newcommand{\intentDeltaLlavaALeverTwoIntent}{+0.0}
\newcommand{\intentDeltaLlavaALeverTwoIntent}{+4.0 \pm 5.7}
\newcommand{\intentDeltaLlavaALeverTwoEpist}{+0.0}
\newcommand{\intentDeltaLlavaATravBelief}{+0.0}
% CANONICAL (replaced 2026-05-25, DISJOINT): \newcommand{\intentDeltaLlavaATravIntent}{+0.0}
\newcommand{\intentDeltaLlavaATravIntent}{+0.7 \pm 0.5}
\newcommand{\intentDeltaLlavaATravEpist}{+0.0}
\newcommand{\intentDeltaInternALeverOneBelief}{+0.0}
\newcommand{\intentDeltaInternALeverOneIntent}{+0.0}
\newcommand{\intentDeltaInternALeverOneEpist}{+0.0}
\newcommand{\intentDeltaInternALeverTwoBelief}{+0.0}
\newcommand{\intentDeltaInternALeverTwoIntent}{+0.0}
\newcommand{\intentDeltaInternALeverTwoEpist}{+0.0}
% CANONICAL (replaced 2026-05-25, DISJOINT): \newcommand{\intentDeltaInternATravBelief}{-0.2 \pm 0.2}
\newcommand{\intentDeltaInternATravBelief}{+1.3 \pm 0.5}
% CANONICAL (replaced 2026-05-25, DISJOINT): \newcommand{\intentDeltaInternATravIntent}{+0.4 \pm 0.9}
\newcommand{\intentDeltaInternATravIntent}{+1.3 \pm 1.1}
% CANONICAL (replaced 2026-05-25, DISJOINT): \newcommand{\intentDeltaInternATravEpist}{+1.3 \pm 0.9}
\newcommand{\intentDeltaInternATravEpist}{+1.1 \pm 0.8}

% --- Bootstrap CIs + sign test on CARD gap-shift M and Δacc ---
\newcommand{\cardSignTestIntentNTotal}{108}
% CANONICAL (replaced 2026-05-25, DISJOINT): \newcommand{\cardSignTestIntentNPos}{15}
\newcommand{\cardSignTestIntentNPos}{12}
\newcommand{\cardSignTestIntentNNeg}{27}
% CANONICAL (replaced 2026-05-25, DISJOINT): \newcommand{\cardSignTestIntentNZero}{66}
\newcommand{\cardSignTestIntentNZero}{69}
% CANONICAL (replaced 2026-05-25, DISJOINT): \newcommand{\cardSignTestIntentP}{0.088}
\newcommand{\cardSignTestIntentP}{0.024}
% CANONICAL (replaced 2026-05-25, DISJOINT): \newcommand{\cardSignTestIntentWilcoxonP}{0.048}
\newcommand{\cardSignTestIntentWilcoxonP}{0.009}
% CANONICAL (replaced 2026-05-25, DISJOINT): \newcommand{\cardSignTestIntentCIIncludesZero}{85}
\newcommand{\cardSignTestIntentCIIncludesZero}{86}
% CANONICAL (replaced 2026-05-25, DISJOINT): \newcommand{\cardSignTestIntentMaxDeltaHi}{5.3}
\newcommand{\cardSignTestIntentMaxDeltaHi}{18.7}
\newcommand{\cardSignTestBeliefNTotal}{108}
% CANONICAL (replaced 2026-05-25, DISJOINT): \newcommand{\cardSignTestBeliefNPos}{61}
\newcommand{\cardSignTestBeliefNPos}{58}
% CANONICAL (replaced 2026-05-25, DISJOINT): \newcommand{\cardSignTestBeliefNNeg}{18}
\newcommand{\cardSignTestBeliefNNeg}{15}
% CANONICAL (replaced 2026-05-25, DISJOINT): \newcommand{\cardSignTestBeliefNZero}{29}
\newcommand{\cardSignTestBeliefNZero}{35}
\newcommand{\cardSignTestBeliefP}{<0.0001}
\newcommand{\cardSignTestBeliefWilcoxonP}{<0.0001}
% CANONICAL (replaced 2026-05-25, DISJOINT): \newcommand{\cardSignTestBeliefCIIncludesZero}{45}
\newcommand{\cardSignTestBeliefCIIncludesZero}{54}
% CANONICAL (replaced 2026-05-25, DISJOINT): \newcommand{\cardSignTestBeliefMaxDeltaHi}{22.0}
\newcommand{\cardSignTestBeliefMaxDeltaHi}{22.9}
\newcommand{\cardSignTestEpistNTotal}{108}
% CANONICAL (replaced 2026-05-25, DISJOINT): \newcommand{\cardSignTestEpistNPos}{53}
\newcommand{\cardSignTestEpistNPos}{52}
% CANONICAL (replaced 2026-05-25, DISJOINT): \newcommand{\cardSignTestEpistNNeg}{36}
\newcommand{\cardSignTestEpistNNeg}{32}
% CANONICAL (replaced 2026-05-25, DISJOINT): \newcommand{\cardSignTestEpistNZero}{19}
\newcommand{\cardSignTestEpistNZero}{24}
% CANONICAL (replaced 2026-05-25, DISJOINT): \newcommand{\cardSignTestEpistP}{0.089}
\newcommand{\cardSignTestEpistP}{0.038}
% CANONICAL (replaced 2026-05-25, DISJOINT): \newcommand{\cardSignTestEpistWilcoxonP}{0.007}
\newcommand{\cardSignTestEpistWilcoxonP}{0.005}
% CANONICAL (replaced 2026-05-25, DISJOINT): \newcommand{\cardSignTestEpistCIIncludesZero}{54}
\newcommand{\cardSignTestEpistCIIncludesZero}{66}
% CANONICAL (replaced 2026-05-25, DISJOINT): \newcommand{\cardSignTestEpistMaxDeltaHi}{42.0}
\newcommand{\cardSignTestEpistMaxDeltaHi}{37.5}

\newcommand{\nProbeSet}{222}
\newcommand{\nEvalSet}{222}
\newcommand{\nEvalALeverOne}{144}
\newcommand{\nEvalALeverTwo}{150}
\newcommand{\nEvalATrav}{150}

\newcommand{\flipRateQwenIntentBelief}{$0.0$}
\newcommand{\flipRateQwenIntentIntent}{$0.0$}
\newcommand{\flipRateQwenIntentEpist}{$0.0$}
\newcommand{\flipRateQwenBeliefBelief}{$15.3{\pm6.2}$}
\newcommand{\flipRateQwenBeliefIntent}{$2.1{\pm2.8}$}
\newcommand{\flipRateQwenBeliefEpist}{$3.5{\pm3.5}$}
\newcommand{\flipRateQwenEpistBelief}{$29.2{\pm6.9}$}
\newcommand{\flipRateQwenEpistIntent}{$0.0$}
\newcommand{\flipRateQwenEpistEpist}{$0.0$}
\newcommand{\flipRateGemmaIntentBelief}{$4.9{\pm3.5}$}
\newcommand{\flipRateGemmaIntentIntent}{$0.0$}
\newcommand{\flipRateGemmaIntentEpist}{$0.0$}
\newcommand{\flipRateGemmaBeliefBelief}{$33.3{\pm7.6}$}
\newcommand{\flipRateGemmaBeliefIntent}{$16.7{\pm6.2}$}
\newcommand{\flipRateGemmaBeliefEpist}{$0.7{\pm1.4}$}
\newcommand{\flipRateGemmaEpistBelief}{$29.9{\pm7.6}$}
\newcommand{\flipRateGemmaEpistIntent}{$19.4{\pm6.9}$}
\newcommand{\flipRateGemmaEpistEpist}{$6.2{\pm4.9}$}
\newcommand{\flipRateLlavaIntentBelief}{$0.0$}
\newcommand{\flipRateLlavaIntentIntent}{$0.0$}
\newcommand{\flipRateLlavaIntentEpist}{$0.0$}
\newcommand{\flipRateLlavaBeliefBelief}{$41.0{\pm8.3}$}
\newcommand{\flipRateLlavaBeliefIntent}{$22.2{\pm6.9}$}
\newcommand{\flipRateLlavaBeliefEpist}{$24.3{\pm7.6}$}
\newcommand{\flipRateLlavaEpistBelief}{$40.3{\pm8.3}$}
\newcommand{\flipRateLlavaEpistIntent}{$67.4{\pm7.6}$}
\newcommand{\flipRateLlavaEpistEpist}{$60.4{\pm8.3}$}
\newcommand{\flipRateInternIntentBelief}{$5.6{\pm4.2}$}
\newcommand{\flipRateInternIntentIntent}{$0.0$}
\newcommand{\flipRateInternIntentEpist}{$15.3{\pm6.2}$}
\newcommand{\flipRateInternBeliefBelief}{$31.9{\pm7.6}$}
\newcommand{\flipRateInternBeliefIntent}{$1.4{\pm2.1}$}
\newcommand{\flipRateInternBeliefEpist}{$35.4{\pm8.3}$}
\newcommand{\flipRateInternEpistBelief}{$29.9{\pm7.6}$}
\newcommand{\flipRateInternEpistIntent}{$3.5{\pm3.5}$}
\newcommand{\flipRateInternEpistEpist}{$20.1{\pm6.9}$}

\newcommand{\flipRateMaxHalfwidth}{8.3}
\newcommand{\figCardTeaser}{%
  }

\newcommand{\figRelayExample}{%
  \begin{figure*}[!t]
    \centering
    \includegraphics[width=0.95\textwidth]{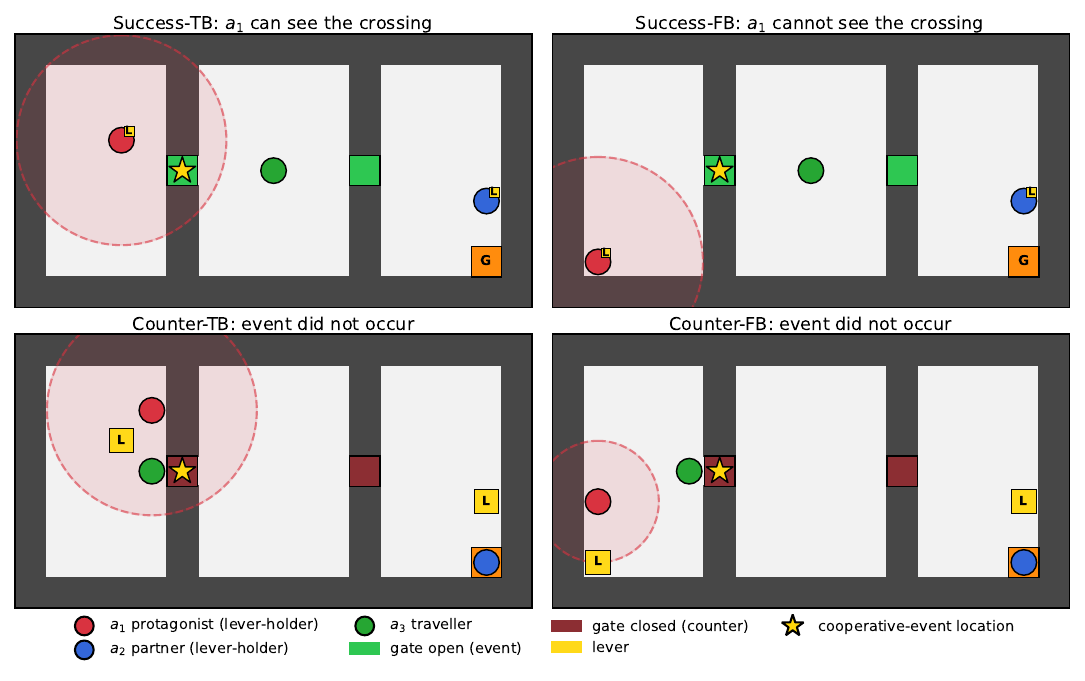}
    \caption{Relay Chain four-cell discrimination layout.  Each
      panel shows the protagonist $a_1$ (red, lever-holder) with a
      translucent fog-of-war disk, partner $a_2$ (blue,
      lever-holder), traveller $a_3$ (green), the two gates (green
      when the cooperative event has opened them, dark-red when
      closed), the levers $L$, and the goal cell $G$.  Top row:
      \textbf{success} scenarios in which the cooperative event
      occurs.  Bottom row: \textbf{counter} scenarios in which the
      lever-holder fails to release the lever at the cooperative
      deadline, so the gate never opens for the traveller to cross.  Left column: \textbf{True-Belief
      (TB)} layout ($a_1$'s lever within fog-of-war of the gate
      it controls); right column: \textbf{False-Belief (FB)} layout
      (lever beyond fog-of-war).  Captions, trajectories,
      identities, and the world-state event are held constant
      within each row; only lever placement differs across the
      pair.}
      \label{fig:relay_example}
  \end{figure*}}

\newcommand{\figCardDual}{%
  \begin{figure*}[!t]
    \centering
    \includegraphics[width=0.95\textwidth]{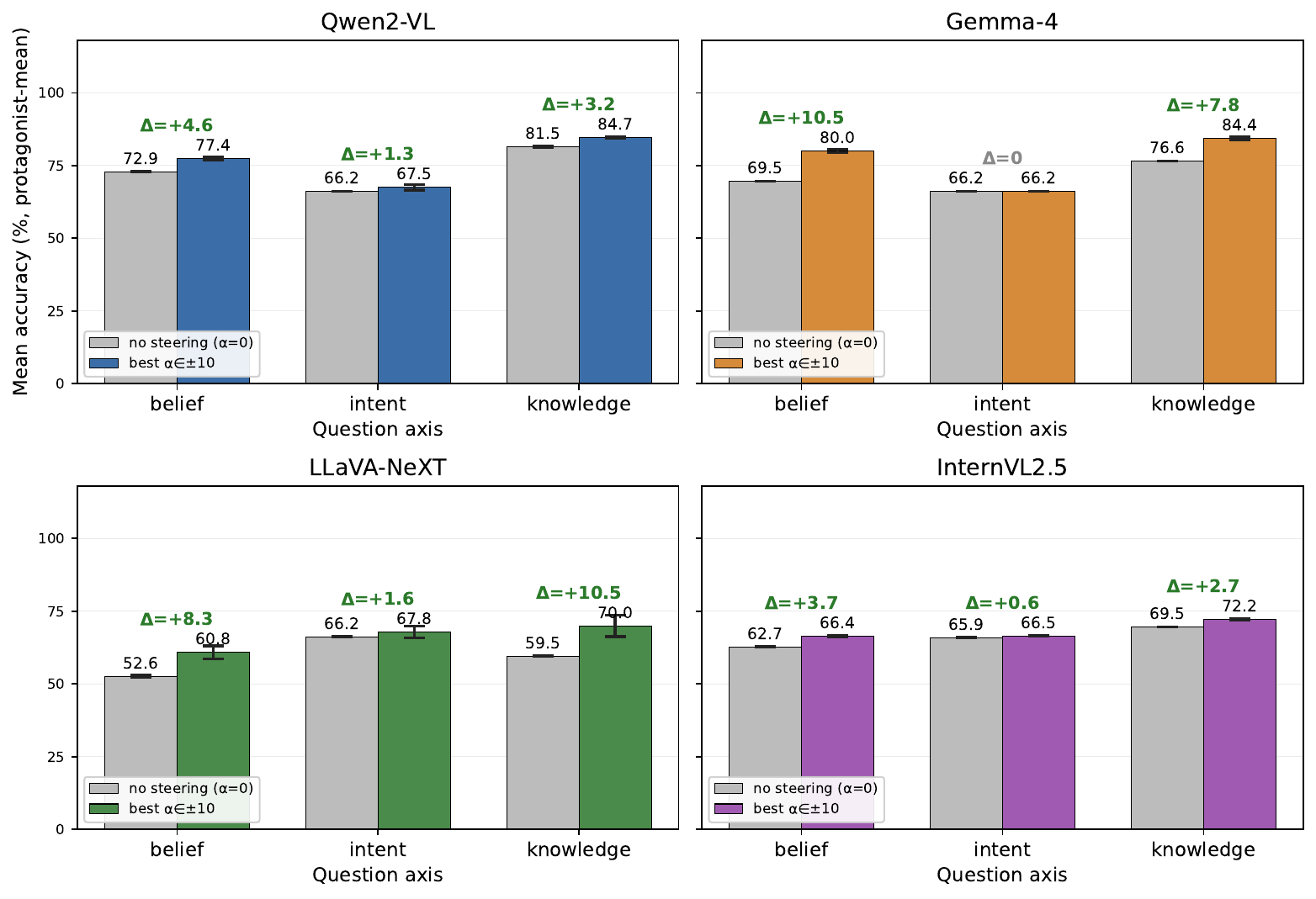}
    \caption{Steering improves accuracy on the belief and
      knowledge answer heads, but never on the intent answer
      head, averaged across the three protagonists ($a_1$,
      $a_2$, $a_3$).  For each (VLM, question axis) we report
      the mean TB/FB accuracy with no steering ($\alpha{=}0$,
      grey) and with the best-case steering
      ($\alpha\in\{-10,+10\}$ across all probe directions, in
      VLM color).  $\Delta$ on top is best$-$baseline.  Belief
      and knowledge improve by $+1.8$ to $+17.7$\,pp; the
      intent $\Delta$ is essentially $0$ on every VLM.  The
      identical intent baseline ($\sim 66\%$) across VLMs
      reflects the F2 constant-prior signature.  Per-protagonist
      breakdown in \autoref{fig:card_per_agent}.}
      \label{fig:card_dual}
  \end{figure*}}

\newcommand{\figCardThreeByThree}{%
  \begin{figure*}[!t]
    \centering
    \includegraphics[width=0.8\textwidth]{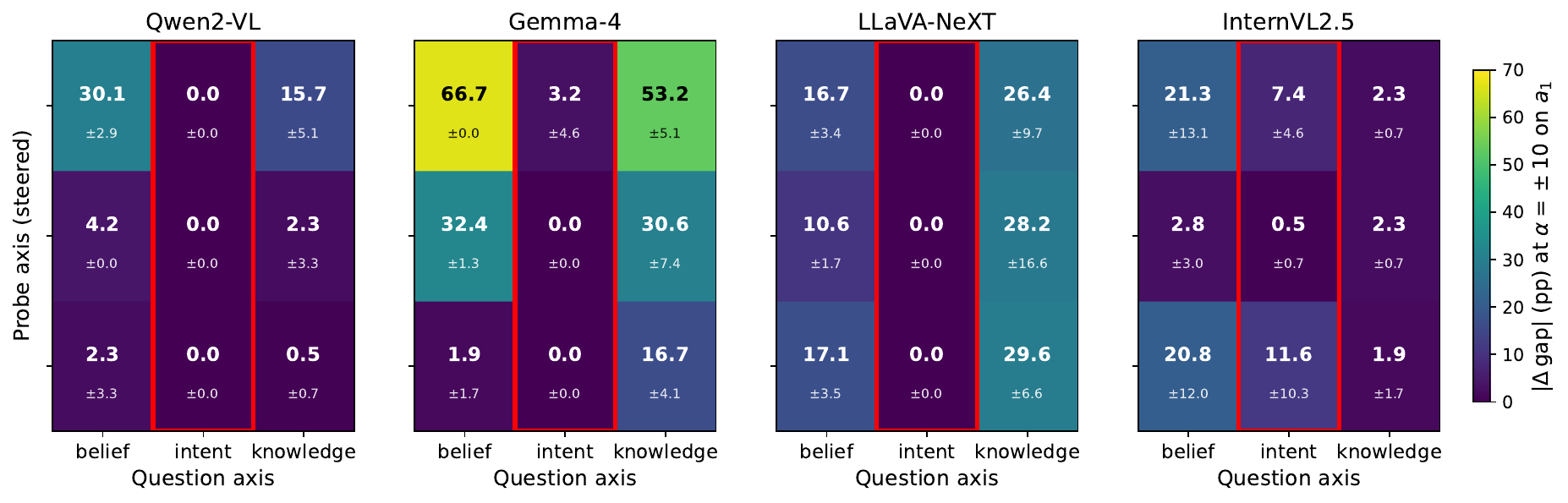}
    \caption{CARD intent column on lever-operator $a_1$ is
      uniformly null (red box) across four VLMs.  Rows: probe
      axis being steered.  Columns: question axis being asked.
      Cell value: $|\text{gap}(\alpha{=}+10) -
      \text{gap}(\alpha{=}-10)|$ in pp, mean $\pm$ std across
      three seeds.  Non-intent columns shift by tens of pp under
      the same probe; the intent column does not move.  Same
      steering hook, same scenarios, same protagonist -- only the
      intent prediction fails to read from the belief
      direction.\label{fig:card3x3}}
  \end{figure*}}

\newcommand{\figTAHeatmap}{%
  \begin{figure*}[!t]
    \centering
    \includegraphics[width=\textwidth]{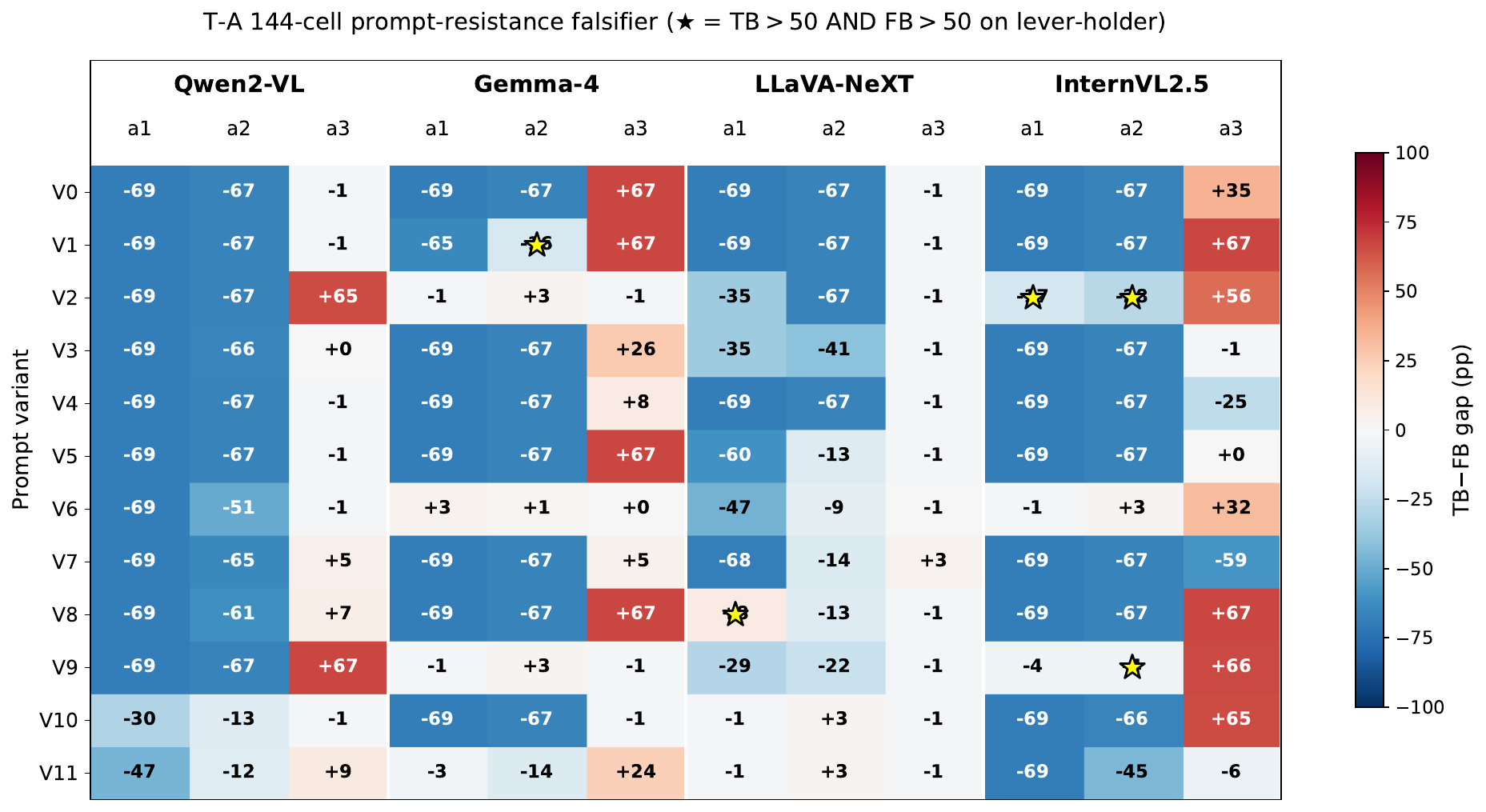}
    \caption{T-A prompt-resistance test.  Each row is one of
      $12$ prompt rewrites we tried (cost framing,
      chain-of-thought, perception-first, rule-based,
      $1$/$2$-shot examples.  Each column is one
      (VLM, protagonist) cell.  Cell colour encodes the
      TB$-$FB accuracy gap -- warm = model is more accurate
      when the protagonist can see the event, cold = the
      reverse.  A star marks cells that cleared our
      pre-registered bar for ``genuinely belief-conditional''
      (lever-holder, TB and FB accuracy both above $50\%$).
      Only $\taLeverHolderCells$ cells star out of
      $\taTotalCells$, and only one points the right way
      (TB$>$FB), with a margin an order of magnitude smaller
      than the same models' gap on the canonical belief question
      -- no prompt rewrite recovers belief-conditional
      action.}
      \label{fig:ta_heatmap}
  \end{figure*}}

\newcommand{\figRelayScenariosAppendix}{%
  \begin{figure*}[!t]
    \centering
    \includegraphics[width=0.95\textwidth]{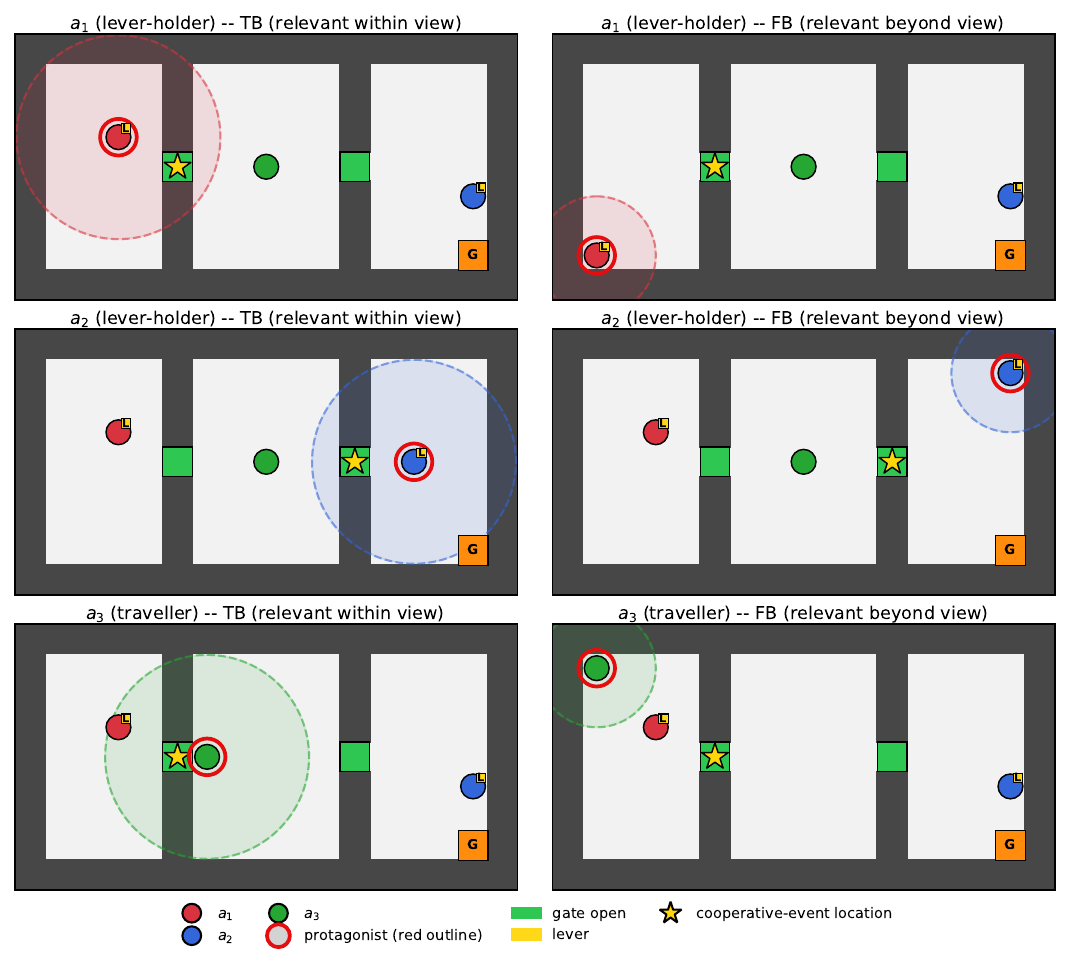}
    \caption{Relay Chain scenario layouts for each protagonist
      condition (rows: $a_1$ and $a_2$ lever-holders, $a_3$
      traveller) under True-Belief and False-Belief layouts
      (columns).  The protagonist is highlighted with a red
      outline; the shaded disk shows their field of view (radius
      $5$ cells; anything outside is fog of war and not
      perceptible).  In TB the protagonist stands next to its
      gate-controlling lever or just before the gate ahead; in FB
      it stands at the far lever or in a corner, so the
      field-of-view disk shifts with it and the cooperative event
      falls outside.  All panels show the success block (both
      gates open) so the contrast is purely
      perceptual.}
      \label{fig:relay_scenarios_appendix}
  \end{figure*}}

\newcommand{\cardAppendixFig}[2]{%
  \begin{figure*}[!t]
    \centering
    \includegraphics[width=0.95\textwidth]{figures/fig_card_#1.pdf}
    \caption{#2 -- full CARD heatmap per protagonist.  Red box
      = intent column.}
      \label{fig:card_#1}
  \end{figure*}}

\newcommand{\projSweepFig}{%
  \begin{figure}[!t]
    \centering
    \includegraphics[width=\columnwidth]{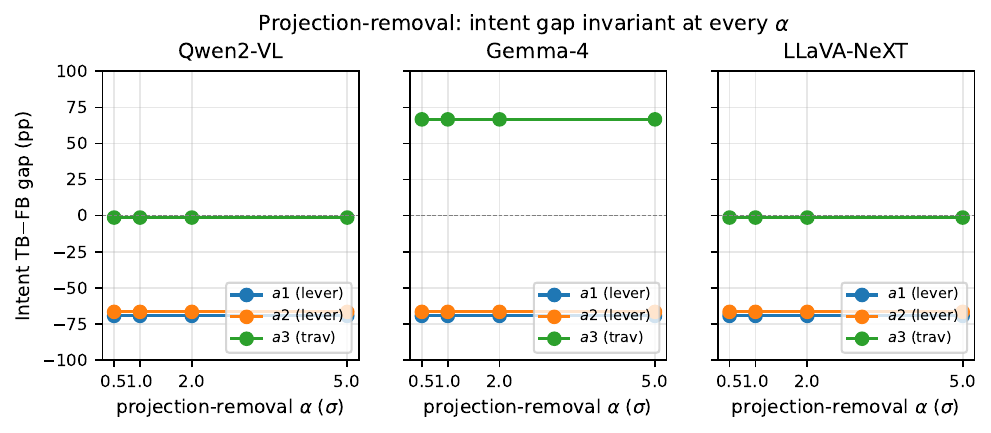}
    \caption{Projection-removal $\alpha$ sweep.  $x$-axis:
      $\alpha\in\{0.5, 1, 2, 5\}\sigma$ removal of the belief
      direction from the residual stream.  $y$-axis: intent TB$-$FB
      gap.  Three lines per VLM: $a_1$, $a_2$, $a_3$.  The intent
      gap is constant across $\alpha$ on every VLM and every
      protagonist.}
      \label{fig:proj_sweep}
  \end{figure}}

\newcommand{\figCardPerAgent}{%
  \begin{figure*}[!t]
    \centering
    \includegraphics[width=\textwidth]{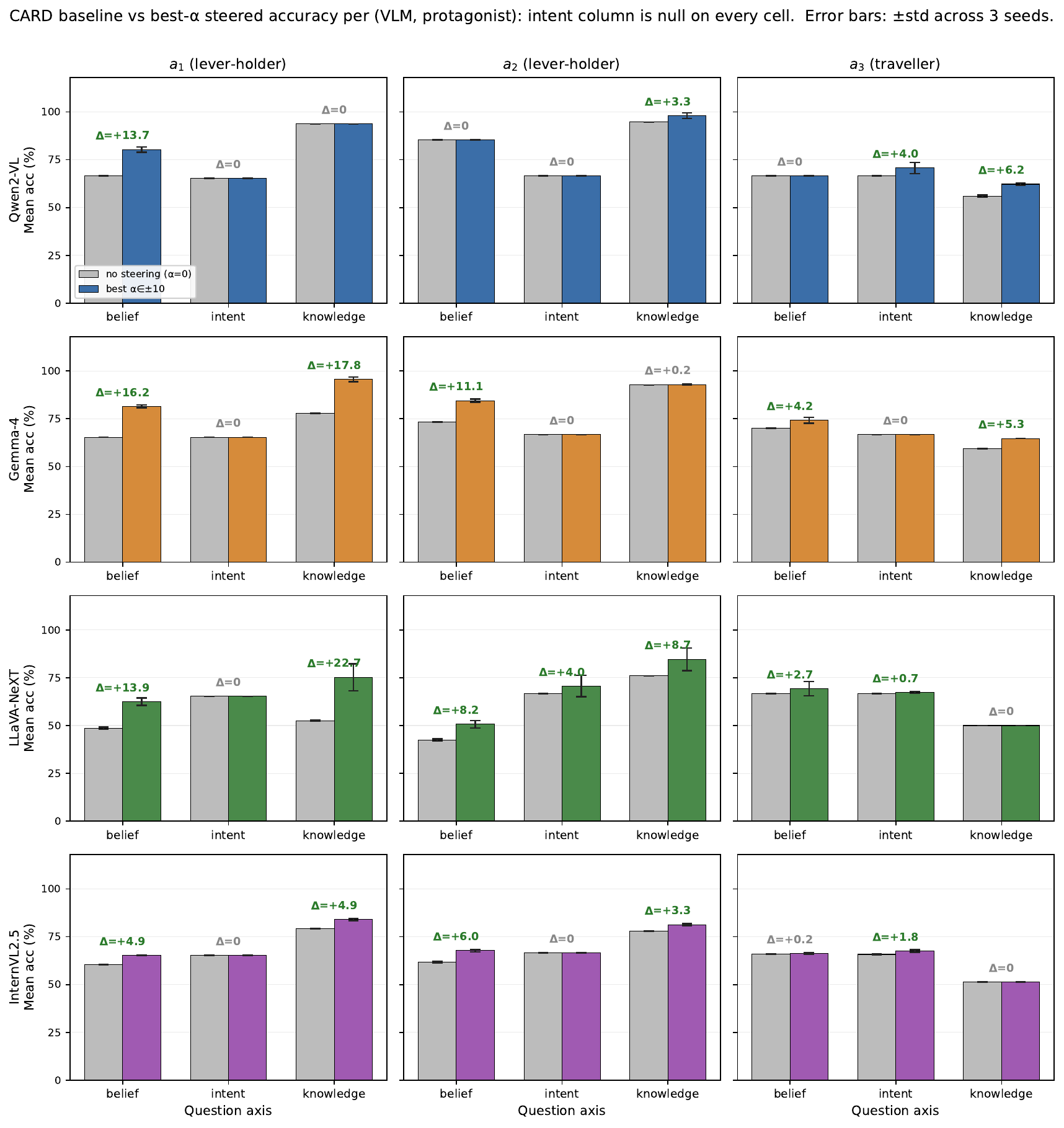}
    \caption{Per-protagonist breakdown of \autoref{fig:card_dual}.
      $4 \times 3$ grid: rows are VLMs, columns are protagonists
      ($a_1$, $a_2$ lever-holders; $a_3$ traveller).  Each cell
      shows mean TB/FB accuracy at $\alpha{=}0$ (grey) vs the
      best $\alpha\in\{-10,+10\}$ across probe directions (VLM
      color); $\Delta$ on top is best$-$baseline.  The intent
      column $\Delta$ is essentially $0$ in every (VLM,
      protagonist) cell ($12/12$), while sibling answer heads
      (belief, knowledge) move on the protagonists that have
      headroom.  The routing failure is therefore not
      protagonist-specific.}
      \label{fig:card_per_agent}
  \end{figure*}}

  \newcommand{\figTemporalProbe}{%
  \begin{figure*}[!t]
    \centering
    \includegraphics[width=0.95\textwidth]{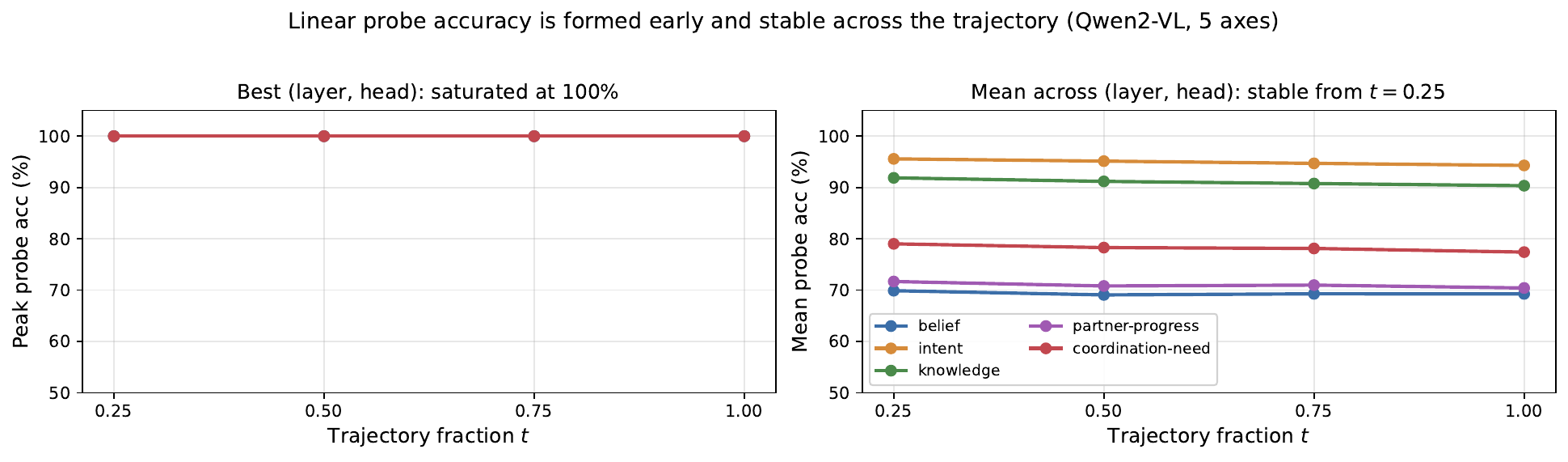}
    \caption{Linear-probe accuracy as a function of trajectory
      fraction $t\in\{0.25, 0.5, 0.75, 1.0\}$ on Qwen2-VL, across
      the three cooperative-ToM axes (belief, intent, knowledge).  \emph{Left:} peak (best
      layer/head) probe accuracy is saturated at $100\%$ from
      $t{=}0.25$ on every axis.  \emph{Right:} mean probe
      accuracy across all (layer, head) cells is already near its
      asymptotic value at $t{=}0.25$ and stable thereafter (slight
      monotone decrease).  The linear representation is formed
      early and maintained throughout the trajectory; the routing
      failure on intent cannot be attributed to
      delayed representation formation.}
      \label{fig:temporal_probe}
  \end{figure*}}
\newcommand{\tabFOne}{%
  \begin{table}[!h]\small\centering
    \begin{tabular}{lccc}
      \hline
      VLM & Probe acc. & TB-acc. & FB-acc. \\
      \hline
      Qwen2-VL    & \probeAccQwenBelief   & $100.0$ & $13.5$ \\
      Gemma-4     & \probeAccGemmaBelief  & $91.9$  & $18.9$ \\
      LLaVA-NeXT  & \probeAccLlavaBelief  & $81.8$  & $18.2$ \\
      InternVL2.5 & \probeAccInternBelief & $83.8$  & $2.7$ \\
      \hline
    \end{tabular}
    \caption{F1 positive control: linear-probe accuracy (chance
      $=50\%$) and behavioural TB/FB accuracy on the canonical
      belief question.}
      \label{tab:f1}
  \end{table}}

\newcommand{\tabPerScenario}{%
  \begin{table}[!h]\small\centering
    \setlength{\tabcolsep}{5pt}
    \begin{tabular}{l c c c}
      \hline
      Probe & Belief Q (\%) & Intent Q (\%) & Epist.\ Q (\%) \\
      \hline
      \multicolumn{4}{l}{\emph{Qwen2-VL}} \\
       \hline
      belief    & $5.6$ & $\mathbf{0.0}$ & $22.2$ \\
      intent    & $0.0$ & $\mathbf{0.0}$ & $0.3$  \\
      knowledge & $0.0$ & $\mathbf{0.0}$ & $8.3$  \\
      \hline
      \multicolumn{4}{l}{\emph{Gemma-4}} \\
      \hline
      belief    & $19.1$ & $\mathbf{0.0}$ & $22.2$ \\
      intent    & $13.9$ & $\mathbf{0.0}$ & $3.1$  \\
      knowledge & $0.0$  & $\mathbf{0.0}$ & $4.9$  \\
      \hline
      \multicolumn{4}{l}{\emph{LLaVA-NeXT}} \\
      \hline
      belief    & $51.7$ & $\mathbf{0.0}$ & $45.8$ \\
      intent    & $17.7$ & $\mathbf{0.0}$ & $21.9$ \\
      knowledge & $0.3$  & $\mathbf{0.0}$ & $18.1$ \\
      \hline
      \multicolumn{4}{l}{\emph{InternVL2.5}} \\
      \hline
      belief    & $27.1$ & $\mathbf{5.9}$ & $20.1$ \\
      intent    & $2.1$  & $\mathbf{0.0}$ & $2.1$  \\
      knowledge & $9.7$  & $\mathbf{0.3}$ & $5.6$  \\
      \hline
    \end{tabular}
    \caption{Per-scenario answer change rate under
      $\alpha{=}{\pm}10$ steering on lever-holder $a_1$ ($288$
      scenarios per cell).  Each cell reports the fraction of
      scenarios whose argmax answer differs between
      $\alpha{=}{+}10$ and $\alpha{=}{-}10$.  The intent column
      flip rate is $\mathbf{0/288}$ on $10$ of the $12$
      (VLM, probe) cells (Qwen, Gemma, LLaVA across all three
      probes; InternVL2.5 under the intent probe) and at most
      $5.9\%$ on the remaining two InternVL2.5 cells, where the
      flips are direction-balanced and cancel in net accuracy
      (intent $\Delta_{\rm acc}$ remains within
      $+0.0$--$+8.7$\,\%).  Sibling
      columns change up to $51.7\%$ of scenarios under the same
      hooks.  Without steering, Qwen $a_1$'s intent answer is
      mixed ($\sim 49\%$ yes / $51\%$ no), so the model is
      genuinely choosing between both options -- the null is
      not because the answer was already pinned to one side.}
      \label{tab:per_scenario}
  \end{table}}

  % -- Hewitt-Liang random-label probe control (appendix) --
\newcommand{\tabRandomLabel}{%
  \begin{table*}[!t]\small\centering
    \setlength{\tabcolsep}{10pt}
    \begin{tabular}{l c c c c}
      \hline
      VLM & Axis & True-label acc & Random-label acc & Selectivity \\
      \hline
      \multirow{3}{*}{Qwen2-VL}
        & belief    & $98.1$ & $47.6$ & $+50.5$ \\
        & intent    & $98.0$ & $47.4$ & $+50.6$ \\
        & knowledge & $98.0$ & $47.9$ & $+50.1$ \\
      \hline
      \multirow{3}{*}{Gemma-4}
        & belief    & $100.0$ & $50.7$ & $+49.3$ \\
        & intent    & $100.0$ & $52.0$ & $+48.0$ \\
        & knowledge & $100.0$ & $50.5$ & $+49.5$ \\
      \hline
      \multirow{3}{*}{LLaVA-NeXT}
        & belief    & $91.6$ & $48.6$ & $+43.0$ \\
        & intent    & $91.6$ & $48.0$ & $+43.6$ \\
        & knowledge & $91.7$ & $47.6$ & $+44.1$ \\
      \hline
    \end{tabular}
    \caption{Random-label probe control
      \citep{hewitt2019designing}.  For each (VLM, axis), we
      retrain logistic-regression probes on the same paired
      activations with randomly permuted labels and report mean
      held-out accuracy across the sampled (layer, head) cells.
      True-label probes attain $91.6$--$100\%$; random-label
      probes are at chance ($47.4$--$52.0\%$), giving selectivity
      $\ge 43.0$\,pp on every axis and every VLM.  The top
      single-cell random-label accuracy across all sampled cells
      is $56.0\%$, well below the true-label minimum of
      $91.6\%$.}
      \label{tab:random_label}
  \end{table*}}

% -- Temporal probe accuracy table (appendix) --
\newcommand{\tabTemporalProbe}{%
  \begin{table*}[!t]\small\centering
    \setlength{\tabcolsep}{5pt}
    \begin{tabular}{l c c c c c}
      \hline
      Axis & $t{=}0.25$ & $t{=}0.5$ & $t{=}0.75$ & $t{=}1.0$ & peak \\
      \hline
      belief             & $69.9$ & $69.1$ & $69.3$ & $69.3$ & $100.0$ \\
      intent             & $95.6$ & $95.1$ & $94.7$ & $94.3$ & $100.0$ \\
      knowledge          & $91.9$ & $91.2$ & $90.8$ & $90.4$ & $100.0$ \\
      \hline
    \end{tabular}
    \caption{Linear-probe held-out accuracy (\%) as a function of
      trajectory fraction $t$ on Qwen2-VL, for the three
      cooperative-ToM axes (belief, intent, knowledge).  ``Peak'' = best (layer, head) probe;
      the remaining columns are means across all (layer, head)
      cells.  The peak is saturated at $100\%$ from $t{=}0.25$
      onward.  Mean accuracy decreases very slightly
      ($\sim 1$\,pp) with $t$: while the \emph{best} head
      remains perfectly recoverable, fewer heads on average carry
      a clean axis-specific signal as later tokens accumulate
      downstream context.  The relevant headline is the saturated
      peak: the feature is present from early in the
      trajectory.}
      \label{tab:temporal_probe}
  \end{table*}}

\newcommand{\tabTAVariants}{%
  \begin{table*}[!t]\small\centering
    \setlength{\tabcolsep}{8pt}
    \begin{tabular}{l l p{0.60\textwidth}}
      \hline
      Variant & Family & Description \\
      \hline
      V0  & cost            & baseline; canonical intent question, no prefix \\
      V1  & cost            & implicit consequences: ``if released too early,
                              gate closes on traveller; team fails.'' \\
      V2  & cost            & explicit numerical cost: release$=$cost $2.0$ if
                              not crossed, hold$=$cost $1.0$ if crossed \\
      V3  & cost            & utility framing: ``what action MAXIMISES team
                              success?'' \\
      \hline
      V4  & CoT             & step-by-step: perception $\to$ belief $\to$ cost
                              $\to$ action \\
      V5  & CoT             & high-stakes framing: ``a wrong release may kill
                              the team'' \\
      V6  & CoT             & V4 chain-of-thought $+$ V2 numerical costs \\
      \hline
      V7  & perception-rule & ``State what the protagonist can see, then
                              answer.'' \\
      V8  & perception-rule & explicit rule: ``if you can see the crossing,
                              release; else hold.'' \\
      V9  & perception-rule & V7 perception-first $+$ V2 cost numbers \\
      \hline
      V10 & in-context      & one-shot worked example (Manhattan
                              distance $\le 5$ $\to$ release;
                              $> 5$ $\to$ hold) \\
      V11 & in-context      & two-shot: V10 plus a second example of opposite
                              polarity \\
      \hline
    \end{tabular}
    \caption{The $12$ T-A prompt variants used to evaluate
      prompt-resistance.  Each variant prepends a prefix to the
      canonical intent question; in-context variants additionally
      include worked examples.}
      \label{tab:ta_variants}
  \end{table*}}

% -- Vendor architecture and rigidity table (appendix) --
\newcommand{\tabVendorSpecs}{%
  \begin{table*}[!t]\small\centering
    \setlength{\tabcolsep}{5pt}
    \begin{tabular}{l r r r r c}
      \hline
      VLM & layers & heads & head\_dim & hidden & rigidity \\
      \hline
      Qwen2-VL-7B    & 28 & 28 & 128 & 3584 & hardest      \\
      LLaVA-NeXT-7B  & 32 & 32 & 128 & 4096 & middle       \\
      Gemma-4-E4B    & 34 &  8 & 320 & 2560 & softest      \\
      InternVL2.5-8B & 32 & 32 & 128 & 4096 & intermediate \\
      \hline
    \end{tabular}
    \caption{Text-decoder architecture specs and empirical
      rigidity ordering of the four VLMs.  \emph{Rigidity} =
      resistance of the intent constant prior to prompt
      perturbation in the T-A sweep: Qwen2-VL is
      \emph{hardest} (no V0--V11 cell breaks $30/100$ on $a_1$);
      LLaVA-NeXT is \emph{intermediate}; Gemma-4 is \emph{softest}
      (V2 cost-explicit breaks the prior on $a_2$ to $70/86$,
      though not toward belief-conditional behaviour).  The
      ordering matches loosely with head dimensionality
      (Gemma's $320$ vs.\ $128$ for the others), but the dataset
      ($4$ vendors) is too small to draw architectural
      conclusions; a quantitative regression of rigidity against
      pretraining mix and RLHF intensity is left to future
      work.}
      \label{tab:vendor_specs}
  \end{table*}}

% -- CARD per-cell breakdown table (appendix) --
\newcommand{\tabCardBreakdown}{%
  \begin{table*}[!t]\small\centering
    \setlength{\tabcolsep}{4pt}
    \begin{tabular}{l c c r r}
      \toprule
      VLM & Probe & Question & $\Delta_{\rm gap}$ (pp) & $\Delta_{\rm acc}$ (pp) \\
      \midrule
      \multirow{9}{*}{Qwen2-VL}
        & belief    & belief    & \cardBreakQwenBeliefBeliefGap & \cardBreakQwenBeliefBeliefAcc \\
        & belief    & \textbf{intent}    & \cardBreakQwenBeliefIntentGap & \cardBreakQwenBeliefIntentAcc \\
        & belief    & knowledge & \cardBreakQwenBeliefEpistGap  & \cardBreakQwenBeliefEpistAcc  \\
        & intent    & belief    & \cardBreakQwenIntentBeliefGap & \cardBreakQwenIntentBeliefAcc \\
        & intent    & \textbf{intent}    & \cardBreakQwenIntentIntentGap & \cardBreakQwenIntentIntentAcc \\
        & intent    & knowledge & \cardBreakQwenIntentEpistGap  & \cardBreakQwenIntentEpistAcc  \\
        & knowledge & belief    & \cardBreakQwenEpistBeliefGap  & \cardBreakQwenEpistBeliefAcc  \\
        & knowledge & \textbf{intent}    & \cardBreakQwenEpistIntentGap  & \cardBreakQwenEpistIntentAcc  \\
        & knowledge & knowledge & \cardBreakQwenEpistEpistGap   & \cardBreakQwenEpistEpistAcc   \\
      \midrule
      \multirow{9}{*}{Gemma-4}
        & belief    & belief    & \cardBreakGemmaBeliefBeliefGap & \cardBreakGemmaBeliefBeliefAcc \\
        & belief    & \textbf{intent}    & \cardBreakGemmaBeliefIntentGap & \cardBreakGemmaBeliefIntentAcc \\
        & belief    & knowledge & \cardBreakGemmaBeliefEpistGap  & \cardBreakGemmaBeliefEpistAcc  \\
        & intent    & belief    & \cardBreakGemmaIntentBeliefGap & \cardBreakGemmaIntentBeliefAcc \\
        & intent    & \textbf{intent}    & \cardBreakGemmaIntentIntentGap & \cardBreakGemmaIntentIntentAcc \\
        & intent    & knowledge & \cardBreakGemmaIntentEpistGap  & \cardBreakGemmaIntentEpistAcc  \\
        & knowledge & belief    & \cardBreakGemmaEpistBeliefGap  & \cardBreakGemmaEpistBeliefAcc  \\
        & knowledge & \textbf{intent}    & \cardBreakGemmaEpistIntentGap  & \cardBreakGemmaEpistIntentAcc  \\
        & knowledge & knowledge & \cardBreakGemmaEpistEpistGap   & \cardBreakGemmaEpistEpistAcc   \\
      \midrule
      \multirow{9}{*}{LLaVA-NeXT}
        & belief    & belief    & \cardBreakLlavaBeliefBeliefGap & \cardBreakLlavaBeliefBeliefAcc \\
        & belief    & \textbf{intent}    & \cardBreakLlavaBeliefIntentGap & \cardBreakLlavaBeliefIntentAcc \\
        & belief    & knowledge & \cardBreakLlavaBeliefEpistGap  & \cardBreakLlavaBeliefEpistAcc  \\
        & intent    & belief    & \cardBreakLlavaIntentBeliefGap & \cardBreakLlavaIntentBeliefAcc \\
        & intent    & \textbf{intent}    & \cardBreakLlavaIntentIntentGap & \cardBreakLlavaIntentIntentAcc \\
        & intent    & knowledge & \cardBreakLlavaIntentEpistGap  & \cardBreakLlavaIntentEpistAcc  \\
        & knowledge & belief    & \cardBreakLlavaEpistBeliefGap  & \cardBreakLlavaEpistBeliefAcc  \\
        & knowledge & \textbf{intent}    & \cardBreakLlavaEpistIntentGap  & \cardBreakLlavaEpistIntentAcc  \\
        & knowledge & knowledge & \cardBreakLlavaEpistEpistGap   & \cardBreakLlavaEpistEpistAcc   \\
      \midrule
      \multirow{9}{*}{InternVL2.5}
        & belief    & belief    & \cardBreakInternBeliefBeliefGap & \cardBreakInternBeliefBeliefAcc \\
        & belief    & \textbf{intent}    & \cardBreakInternBeliefIntentGap & \cardBreakInternBeliefIntentAcc \\
        & belief    & knowledge & \cardBreakInternBeliefEpistGap  & \cardBreakInternBeliefEpistAcc  \\
        & intent    & belief    & \cardBreakInternIntentBeliefGap & \cardBreakInternIntentBeliefAcc \\
        & intent    & \textbf{intent}    & \cardBreakInternIntentIntentGap & \cardBreakInternIntentIntentAcc \\
        & intent    & knowledge & \cardBreakInternIntentEpistGap  & \cardBreakInternIntentEpistAcc  \\
        & knowledge & belief    & \cardBreakInternEpistBeliefGap  & \cardBreakInternEpistBeliefAcc  \\
        & knowledge & \textbf{intent}    & \cardBreakInternEpistIntentGap  & \cardBreakInternEpistIntentAcc  \\
        & knowledge & knowledge & \cardBreakInternEpistEpistGap   & \cardBreakInternEpistEpistAcc   \\
      \bottomrule
    \end{tabular}
    \caption{CARD per-cell breakdown on lever-holder $a_1$,
      reporting both the gap shift
      $\Delta_{\rm gap} = {\rm gap}_B(\alpha{=}{+}10) -
      {\rm gap}_B(\alpha{=}{-}10)$ and the mean-accuracy shift
      $\Delta_{\rm acc} = \overline{\rm acc}_B(\alpha{=}{+}10) -
      \overline{\rm acc}_B(\alpha{=}{-}10)$, where
      $\overline{\rm acc}_B = \tfrac12({\rm TB}_B + {\rm FB}_B)$.
      Bold rows are the intent-question column: $9$ of the $12$
      cells are $\mathbf{0.00}$ on \emph{both} metrics across the
      four VLMs (Qwen2-VL and LLaVA-NeXT all three probes; Gemma-4
      and InternVL2.5 under intent and knowledge probes), and the
      remaining three (Gemma-4 and InternVL2.5 under belief
      steering, InternVL2.5 under knowledge steering) shift by at
      most $|\Delta_{\rm gap}|{=}25.0$\,pp / $|\Delta_{\rm acc}|{=}13.9$\,pp,
      ruling out both belief-conditional modulation
      ($\Delta_{\rm gap}$) and any uniform answer-flip
      ($\Delta_{\rm acc}$).  Sibling columns (belief, knowledge)
      show non-trivial shifts under the same probe directions.}
      \label{tab:card_breakdown}
  \end{table*}}

% -- Dataset counts table (appendix) --
\newcommand{\tabDatasetCounts}{%
  \begin{table*}[!t]\small\centering
    \setlength{\tabcolsep}{8pt}
    \begin{tabular}{l c c c c}
      \toprule
      Protagonist & Role & Success & Counter & Total \\
      \midrule
      $a_1$ & lever-operator & $100$ & $44$ & $144$ \\
      $a_2$ & lever-operator & $100$ & $50$ & $150$ \\
      $a_3$ & traveller    & $100$ & $50$ & $150$ \\
      \midrule
      \multicolumn{2}{l}{\textbf{Total}}
                           & $\mathbf{\nSuccess}$
                           & $\mathbf{\nCounter}$
                           & $\mathbf{\nTotalFour}$ \\
      \bottomrule
    \end{tabular}
    \caption{Relay Chain dataset breakdown by protagonist and
      world-state outcome.  All three axes (belief, intent,
      knowledge) share the same scenario base; each entry is
      a multi-frame trajectory with $4$--$8$
      keyframes.}
      \label{tab:dataset_counts}
  \end{table*}}

% -- D1 polarity-flip table --
\newcommand{\tabDOneFull}{%
  \begin{table*}[!t]\small\centering
    \begin{tabular}{l c c c}
      \toprule
      VLM & overall orig (pp) & overall flip (pp) & sign-preserved? \\
      \midrule
      Qwen2-VL     & \dOneOrigQwen   & \dOneFlipQwen   & yes (oracle) \\
      Gemma-4      & \dOneOrigGemma  & \dOneFlipGemma  & yes (oracle) \\
      LLaVA-NeXT   & \dOneOrigLlava  & \dOneFlipLlava  & weakened (drift) \\
      \bottomrule
    \end{tabular}
    \caption{D1 polarity-flip diagnostic.  Original and flipped
      question polarity TB$-$FB gap on the canonical $\nHoldoutF$
      hold-out.  An oracle-reasoning model preserves the sign and
      magnitude under flip; a constant-prior model collapses to a
      smaller or sign-flipped gap.}
      \label{tab:d1_polarity}
  \end{table*}}

% -- D2 full discrimination table — 2-column wide in appendix --
\newcommand{\tabDTwoFull}{%
  \begin{table*}[!t]\small\centering
    \setlength{\tabcolsep}{8pt}
    \begin{tabular}{l l c c c}
      \hline
      VLM & Axis & Lever $a_1$ & Lever $a_2$ & Trav.\ $a_3$ \\
      \hline
      \multirow{3}{*}{Qwen2-VL}   & belief    & \dTwoQwenBeliefALeverOne   & \dTwoQwenBeliefALeverTwo   & \dTwoQwenBeliefATrav   \\
                                  & intent    & \dTwoQwenIntentALeverOne   & \dTwoQwenIntentALeverTwo   & \dTwoQwenIntentATrav   \\
                                  & know.    & \dTwoQwenEpistemicALeverOne& \dTwoQwenEpistemicALeverTwo& \dTwoQwenEpistemicATrav\\
      \hline
      \multirow{3}{*}{Gemma-4}    & belief    & \dTwoGemmaBeliefALeverOne  & \dTwoGemmaBeliefALeverTwo  & \dTwoGemmaBeliefATrav  \\
                                  & intent    & \dTwoGemmaIntentALeverOne  & \dTwoGemmaIntentALeverTwo  & \dTwoGemmaIntentATrav  \\
                                  & know.    & \dTwoGemmaEpistemicALeverOne&\dTwoGemmaEpistemicALeverTwo&\dTwoGemmaEpistemicATrav\\
      \hline
      \multirow{3}{*}{LLaVA-NeXT} & belief    & \dTwoLlavaBeliefALeverOne  & \dTwoLlavaBeliefALeverTwo  & \dTwoLlavaBeliefATrav  \\
                                  & intent    & \dTwoLlavaIntentALeverOne  & \dTwoLlavaIntentALeverTwo  & \dTwoLlavaIntentATrav  \\
                                  & know.    & \dTwoLlavaEpistemicALeverOne&\dTwoLlavaEpistemicALeverTwo&\dTwoLlavaEpistemicATrav\\
      \hline
      \multirow{3}{*}{InternVL2.5}& belief    & \dTwoInternBeliefALeverOne & \dTwoInternBeliefALeverTwo & \dTwoInternBeliefATrav \\
                                  & intent    & \dTwoInternIntentALeverOne & \dTwoInternIntentALeverTwo & \dTwoInternIntentATrav \\
                                  & know.    & \dTwoInternEpistemicALeverOne&\dTwoInternEpistemicALeverTwo&\dTwoInternEpistemicATrav\\
      \hline
    \end{tabular}
    \caption{Full D2 four-tuple discrimination matrix
      (success-TB / success-FB / counter-TB / counter-FB) for every
      (VLM, axis, protagonist).}
      \label{tab:d2_full}
  \end{table*}}

  \newcommand{\tabClevrCard}{%
  \begin{table*}[!t]\small\centering
    \setlength{\tabcolsep}{5pt}
    \begin{tabular}{l l c c c c c}
      \hline
      VLM & Question & Baseline & P=count & P=shape & P=colour & best \\
      \hline
      \multirow{3}{*}{Qwen2-VL}
        & count  & $86$ & $+6$ & $+2$ & $+6$ & $\mathbf{+6}$ \\
        & shape  & $100$ & $+0$ & $+0$ & $+0$ & $+0$ (ceil) \\
        & colour & $100$ & $+0$ & $+0$ & $+0$ & $+0$ (ceil) \\
      \hline
      \multirow{3}{*}{Gemma-4}
        & count  & $20$ & $+26$ & $+60$ & $+42$ & $\mathbf{+60}$ \\
        & shape  & $30$ & $+24$ & $+58$ & $+52$ & $\mathbf{+58}$ \\
        & colour & $52$ & $+16$ & $+26$ & $+22$ & $\mathbf{+26}$ \\
      \hline
      \multirow{3}{*}{LLaVA-NeXT}
        & count  & $20$ & $+40$ & $+6$ & $+6$ & $\mathbf{+40}$ \\
        & shape  & $0$ & $+8$ & $+2$ & $+0$ & $\mathbf{+8}$ \\
        & colour & $6$ & $+6$ & $+8$ & $+10$ & $\mathbf{+10}$ \\
      \hline
      \multirow{3}{*}{InternVL2.5}
        & count  & $86$ & $+0$ & $+14$ & $+10$ & $\mathbf{+14}$ \\
        & shape  & $100$ & $+0$ & $+0$ & $+0$ & $+0$ (ceil) \\
        & colour & $96$ & $+0$ & $+2$ & $+2$ & $+2$ (ceil) \\
      \hline
      \multirow{3}{*}{SmolVLM-500M}
        & count  & $12$ & $+6$ & $+10$ & $+26$ & $\mathbf{+26}$ \\
        & shape  & $70$ & $+10$ & $+2$ & $+12$ & $\mathbf{+12}$ \\
        & colour & $64$ & $+0$ & $+0$ & $+10$ & $\mathbf{+10}$ \\
      \hline
    \end{tabular}
    \caption{CARD positive control on CLEVR (single image,
      $100$ scenarios per axis, seed $43$).  ``Baseline'' = mean
      accuracy at $\alpha{=}0$ on the corresponding within-axis
      cell; each \emph{probe} column = best-$\alpha\in\pm 10$
      accuracy gain.  Bold = the largest non-zero $\Delta_{\rm acc}$
      per (VLM, question).  At least one cross-axis cell shows
      $\ge +14$\,pp gain on every VLM that has headroom (Qwen and
      InternVL are at $\ge 96\%$ on shape/colour; Gemma and LLaVA
      are well below ceiling on every axis and steering produces
      $+6$ to $+60$\,pp gains).  The steering hook is functional;
      the Relay Chain intent null in F3 is not a hook-broken
      artefact.}
      \label{tab:clevr_card}
  \end{table*}}

\newcommand{\tabIntentDelta}{%
  \begin{table*}[!t]\small\centering
    \setlength{\tabcolsep}{6pt}
    \begin{tabular}{l c c c c}
      \toprule
      VLM & Protag.\ & Probe$=$belief & Probe$=$intent & Probe$=$knowledge \\
      \midrule
      \multirow{3}{*}{Qwen2-VL}
        & $a_1$ & $\intentDeltaQwenALeverOneBelief$ & $\intentDeltaQwenALeverOneIntent$ & $\intentDeltaQwenALeverOneEpist$ \\
        & $a_2$ & $\intentDeltaQwenALeverTwoBelief$ & $\intentDeltaQwenALeverTwoIntent$ & $\intentDeltaQwenALeverTwoEpist$ \\
        & $a_3$ & $\intentDeltaQwenATravBelief$ & $\intentDeltaQwenATravIntent$ & $\intentDeltaQwenATravEpist$ \\
      \midrule
      \multirow{3}{*}{Gemma-4}
        & $a_1$ & $\intentDeltaGemmaALeverOneBelief$ & $\intentDeltaGemmaALeverOneIntent$ & $\intentDeltaGemmaALeverOneEpist$ \\
        & $a_2$ & $\intentDeltaGemmaALeverTwoBelief$ & $\intentDeltaGemmaALeverTwoIntent$ & $\intentDeltaGemmaALeverTwoEpist$ \\
        & $a_3$ & $\intentDeltaGemmaATravBelief$ & $\intentDeltaGemmaATravIntent$ & $\intentDeltaGemmaATravEpist$ \\
      \midrule
      \multirow{3}{*}{LLaVA-NeXT}
        & $a_1$ & $\intentDeltaLlavaALeverOneBelief$ & $\intentDeltaLlavaALeverOneIntent$ & $\intentDeltaLlavaALeverOneEpist$ \\
        & $a_2$ & $\intentDeltaLlavaALeverTwoBelief$ & $\intentDeltaLlavaALeverTwoIntent$ & $\intentDeltaLlavaALeverTwoEpist$ \\
        & $a_3$ & $\intentDeltaLlavaATravBelief$ & $\intentDeltaLlavaATravIntent$ & $\intentDeltaLlavaATravEpist$ \\
      \midrule
      \multirow{3}{*}{InternVL2.5}
        & $a_1$ & $\intentDeltaInternALeverOneBelief$ & $\intentDeltaInternALeverOneIntent$ & $\intentDeltaInternALeverOneEpist$ \\
        & $a_2$ & $\intentDeltaInternALeverTwoBelief$ & $\intentDeltaInternALeverTwoIntent$ & $\intentDeltaInternALeverTwoEpist$ \\
        & $a_3$ & $\intentDeltaInternATravBelief$ & $\intentDeltaInternATravIntent$ & $\intentDeltaInternATravEpist$ \\
      \bottomrule
    \end{tabular}
    \caption{Best-$\alpha$ intent accuracy gain (pp) over the
      no-steering baseline ($\alpha{=}0$) for every (VLM,
      protagonist, probe direction) cell, reported as
      mean $\pm$ std across three seeds
      (seeds $\{42, 43, 44\}$; the $\pm$ is dropped when the
      std rounds to $0.0$).  Most cells are exactly $0$ on every
      seed; across the full multi-seed sweep,
      $\cardIntentMultiseedZeroCells / \cardIntentMultiseedTotalCells$
      cells are exactly $0$ and the largest gain anywhere is
      $\cardIntentMultiseedMaxGain$\,\%.  Compare with belief
      and knowledge accuracies, which improve by $+2.7$ to
      $+10.6$\,pp on the across-protagonist mean under the same
      hooks (\autoref{fig:card_dual}): the intent prediction
      does not use the belief direction usefully on any
      cell.
      \label{tab:intent_delta}}
  \end{table*}}

\begin{abstract}
Linear probes and activation steering have uncovered that vision-language models (VLMs) internally represent mental states such as agents' beliefs, knowledge, and intentions. However, it is unclear whether and how these representations
are used by downstream predictions along these axes. To close this gap, we introduce \textit{Cross-Axis Routing Diagnostic} (CARD), which steers activations along one axis while measuring the response of a \emph{different} axis's prediction. 
Applied to open-weight VLMs on
\textit{Relay Chain} -- a new cooperative grid-world benchmark we propose -- we diagnose a critical routing failure: models fail to incorporate belief representations into their next action prediction, effectively leaving valuable information about their partners unused.
\end{abstract}
\section{Introduction}\label{sec:intro}

Vision-language models (VLMs) are increasingly used in human-AI cooperation, such as embodied robots, AI coaches, or driving co-pilots, that require Theory of Mind (ToM; \citealp{premack1978does}) to infer what the interaction partner might see, know or want.
% \andreas{it remains (probably intentional?!) fuzzy if/how these diagnostic tools are linked/specific to collaborative settings}
To diagnose whether such models actually represent such mental states of their partners, two families of diagnostic tools have emerged:
\textit{Linear probing} tests if and where states are encoded in the models, while 
% \textit{activation steering} tests if they causally influence model behaviour.
\textit{activation steering} has primarliy been used to verify whether probed directions causally drive the model's prediction and to control its outputs.
% both are individually well-motivated, but together they leave one question unanswered: Does the model's internal representation of the partner's mental state actually drive its answer?
However, they leave one question unanswered: Does the model's internal representation of a partner's mental state actually drive its actions?

% Two families of diagnostic tools have emerged around such systems and are individually well-motivated, but together they leave one question unanswered: Does the model's internal representation of the partner's mental state actually drive its answer?
Existing behavioural benchmarks \citep{gu2024simpletom,bortoletto2025tom,shi2025muma,chen2024socialbench} score outputs along three axes, belief, knowledge and intentions but without inspecting and connecting their internal representations to the model predictions.
Thus, they cannot discriminate wrong actions that reflect an absent representation from those that reflect an unused one.
Alternatively, probes and prior work in activation steering \citep{alain2016understanding,belinkov2022probing,hewitt2019designing,elazar2021amnesic,bortoletto2024brittle,rauker2023toward,geiger2021causal,wu2023interpretability} access internal representations but apply both read-out and intervention only to the \emph{same} axis. %a probe confirms that a feature is encoded and a steering vector confirms the same feature's own prediction responds, yet neither shows whether the feature is read by a different downstream prediction.
For example, a probe might find a belief representation, and steering confirms its influence on next belief prediction (same-axis) but does not test its influence on action prediction (cross-axis).
Thus, both approaches do not diagnose the routing of mental representations to predictions across axes.
%This identifiability gap is consequential: An absent representation motivates richer pre-training, while a present-but-unread representation motivates intervention on the prediction that should consume it.
This matters especially in cooperative tasks, where the correct action depends on the mental state representation of the partner agent, which the VLM may encode internally, yet doesn't use when predicting its next action.
% \fabian{The next paragraph should introduce the cooperative game, for which GridToM is the inspiration. I would not call it "the closest multimodal effort" since multimodal was not mentioned at all yet and it makes your contribution seem smaller}

% The closest multimodal effort, GridToM~\citep{li2025black}, pairs probes with same-axis activation shifts on a grid-world Theory of Mind (ToM; \citealp{premack1978does}) dataset and inherits exactly this under-identification.

% \andreas{other contributions are not introduced/mentioned}

To address this limitation, we make the following contributions:

\begin{enumerate}[leftmargin=10px]                                                
    \item We introduce \textit{Cross-Axis Routing Diagnostic} (CARD),
    a novel approach to diagnose how mental state representations influence model predictions across 
    % belief, knowledge, and intention/action
    axes by steering representations in one axis and assessing the prediction of another.
    \item To evaluate CARD, we propose \textit{Relay Chain}, a new multi-agent cooperative grid-world benchmark that extends the grid-world format of GridToM~\citep{li2025black} with a cooperative action axis.
  \item Our experiments demonstrate critical routing failures across four open-weight VLM families, where the belief representation causally influences the belief and knowledge predictions but not the action prediction -- despite the belief being linearly decodable at $78$--$94\%$ from the same layer.
  % We validate this finding with multiple robustness checks.
  \end{enumerate}

\begin{figure*}[!t]
    \centering
    \includegraphics[width=\textwidth]{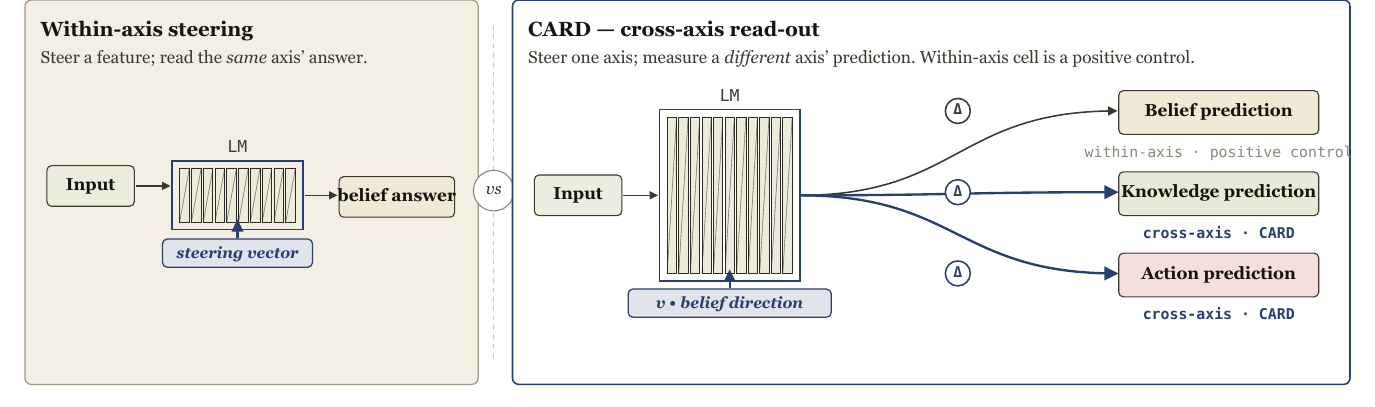}
    \caption{Overview of CARD. For each axis
      $a \in \{\text{belief, intent, knowledge}\}$ a linear probe
      on paired True-Belief / False-Belief activations at layer
      $L$ yields a unit-norm direction $v_a$.  At inference time, we
      hook the residual stream at the same layer and add
      $\alpha v_a$ (steering axis $a$, $\alpha \in \{-10, +10\}$),
      then read out the answer head for a \emph{different}
      question axis $b \neq a$.  A causally-routed feature moves
      the cross-axis read-out; an encoded-but-unread feature does
      not.  Within-axis cells ($a = b$, diagonal) serve as
      positive controls. 
      % \andreas{also hard to read as the font is too small}
    % \andreas{font is still a bit small. Rule of thumb is: font used in the figure should not be smaller than the caption font.}
      %The headline result: the intent column is uniformly null across four open-weight VLMs (see \autoref{fig:card3x3}), while sibling columns shift by tens of       \%.
      }
      \label{fig:card_teaser}
  \end{figure*}
\section{Related Work}\label{sec:related}

% \andreas{reconsider the odering of subsections -- the idea is to have the least closely related first and, as you go along, the subsections and their topics become increasingly relevant and related to the current work. Is this the case here?}

% \andreas{think of related work as a means to support your individual claims regarding novelty/contribution: A sequence of subsections that each covers one related area and that each includes 1-2 sentences at the end to relate to what you are doing in this paper and to clarify how your work is different/novel}

% \fabian{I would go in this order: (1) Overall ToM benchmarking (does not look inside, CARD does), (2) Probes and AS are tools to selectively inspect (but only single/same-axis, CARD tracks cross-axis routing), (3) Others have studied routing/circuits but not specifically on ToM agents}
\paragraph{Behavioural ToM benchmarks.}
% A large body of work evaluated ToM in foundation models by scoring answers to mental-state questions
A large body of work evaluates ToM in foundation models through mental-state questions
\citep{gu2024simpletom,bortoletto2025tom,shi2025muma,xu2024opentom,wu2023hi,chen2024socialbench,xiaotomvalley,chen2024tombench,jin2024mmtom,kim2023fantom,zhang2025mindpower}.
SimpleToM \citep{gu2024simpletom} demonstrated a gap between explicit and applied ToM on text LLMs while ToM-SSI \citep{bortoletto2025tom} extended to situated multi-agent interactions and reported a behavioural drop from percept to belief to intention.
MindPower \citep{zhang2025mindpower} showed that ToM-aware prompting improved cooperative success rates in embodied VLM settings.
CARD complements these by testing whether a model's internal mental-state representations actually drive its cooperative actions, not just whether its overall predictions are correct.

\paragraph{Probes and activation steering.}
Probes and standard (same-axis) activation steering to inspect models have a long history in text-only LLMs
\citep{alain2016understanding,belinkov2022probing,elazar2021amnesic,bortoletto2024brittle,rauker2023toward}:
A linear probe verifies that a feature is encoded in the residual stream, and steering amplifies the probe direction to test
whether the model's answer to the same feature shifts.
GridToM
\citep{li2025black} brought this protocol to multimodal grid-world
ToM and was the closest prior multimodal-ToM work pairing
probes with activation interventions.
CARD extends the protocol
from same-axis to a cross-axis matrix that tests
whether a probe direction is read by a \emph{different} prediction.
Like \citet{bortoletto2024brittle}, we use linear probes with
Hewitt--Liang random-label controls~\citep{hewitt2019designing}
to verify that the recovered belief direction is structured
rather than a probe-expressivity artefact.

% \paragraph{Mechanistic interpretability for VLMs.}
\paragraph{Studying routing in VLMs}
Mechanistic interpretability tools to study full circuits from inputs to outputs that drive predictions include activation patching and
causal mediation
\citep{vig2020investigating,meng2022locating},
activation-addition / steering vectors
\citep{turner2024activation,li2023inference,rimsky2024steering,zou2023representation},
direct logit attribution \citep{wang2022interpretability}, automated
circuit discovery \citep{conmy2023automated}, and linear-relation
decoding \citep{geva2023dissecting,hernandez2024linearity}.
Recent work extended these to VLMs
\citep{yang2026circuit,liu2025mechanistic,haon2025mechanistic}.
Unlike distributed alignment search~\citealp{geiger2021causal,wu2023interpretability}, which
gradient-searches a rotation aligning a high-level variable to a
subspace, CARD uses a fixed pre-trained probe direction and
conditions the read-out on a specific prediction. These techniques have largely targeted factual recall, refusal, or bias; CARD applies cross-axis routing diagnostics to mental-state representations in cooperative agents.
\section{Method}\label{sec:methods}

% Our \textit{Cross-Axis Routing Diagnostic} (CARD) has three components: a paired-statement linear probe that yields a per-axis steering direction; an additive last-token hook that injects that direction at inference and reads off the prediction of a \emph{different} axis; and a projection-removal control that subtracts the same direction.

Our \textit{Cross-Axis Routing Diagnostic} (CARD) takes three steps: (1) identify a steering vector for a mental representation in one axis, e.g. belief; (2) steer the model in its direction and measure the prediction of a \emph{different} axis, e.g. action; (3) repeat the steering in the opposite direction as a control.
% We evaluate it on the Relay Chain dataset (\autoref{sec:task}).

% 1. Probe Training
\paragraph{Paired-statement probe extraction.}
% For each axis $A \in \{\text{belief, intent, knowledge}\}$ and scenario $s$, we present the VLM with the keyframe clip and a pair of statements $\langle s_A^{+}, s_A^{-}\rangle$ that differ only in the truth value of the axis variable (e.g.\ for belief, ``the protagonist saw the traveller crossing'' vs.\ the negation).
% Let $\mathbf{z}(s) \in \mathbb{R}^d$ denote the per-(layer, head)
% activation at the final token; the training input is the
% difference
% \begin{equation}
% \Delta_A(s) \;=\; \mathbf{z}(s_A^{+}) - \mathbf{z}(s_A^{-}).
% \label{eq:probe-delta}
% \end{equation}
% A logistic-regression probe (\(L^2\))
% is trained on $\{\Delta^{\ell,h}_A(s)\}_s$ per (layer, head) to
% predict the binary axis label, yielding a unit-norm probe
% direction $u^{\ell,h}_A \in \mathbb{R}^{d_h}$ and a held-out
% accuracy.  We use the top-$K{=}56$ (layer, head) cells by
% held-out accuracy as the steering pool, and let
% $\sigma^{\ell,h}_A$ denote the training-set standard deviation
% of the scalar projection
% $u^{\ell,h}_A \cdot \mathbf{z}_{\ell,h}(s_A^{+})$.
% The FB variant uses an asymmetric oracle / protagonist-POV
% caption rule (see Appendix~\ref{app:probe-extract}).
For each axis $A \in \{\text{belief, intent, knowledge}\}$ and
scenario $s$, we present the VLM with the keyframe clip and a
pair of statements $\langle s_A^{+}, s_A^{-}\rangle$:
$s_A^{+}$ matches the ground-truth axis label (e.g.\ for belief,
``the protagonist saw the traveller crossing'' when the
protagonist did see it) and $s_A^{-}$ is its negation.
Let $\mathbf{z}^{\ell,h}(s) \in \mathbb{R}^{d_h}$ denote the
per-(layer, head) activation at the final input token under
each statement.  Per (layer, head), an $L^2$-regularised
logistic regression is fit on the $2N$ activations
$\{\mathbf{z}^{\ell,h}(s_A^{+}),\, \mathbf{z}^{\ell,h}(s_A^{-})\}_s$
with binary labels (1 for $s_A^{+}$, 0 for $s_A^{-}$), yielding
a unit-norm probe direction
$u^{\ell,h}_A \in \mathbb{R}^{d_h}$ that points from negated
toward matched, plus a held-out accuracy.  We use the top-$K{=}56$
(layer, head) pair ranked by held-out accuracy as the steering
pool ($\sim$7\% of $L\!\times\!H$ heads, following the
head-selection convention of \citealp{li2023inference}).
$\sigma^{\ell,h}_A$ is the per-head activation standard
deviation along $u^{\ell,h}_A$ on the training set.  For
False-Belief scenarios the matched-class caption is taken from
the protagonist's view rather than the world state
(see Appendix~\ref{app:probe-extract}).

% 2. Steering and Measuring
\paragraph{Cross-Axis Routing Diagnostic (CARD).}
Given the steering-axis probe set
$\{(u^{\ell,h}_A, \sigma^{\ell,h}_A)\}$ over the top-$K$ pool,
we register a forward hook on the self-attention output of
every layer that contains at least one selected
(layer, head) pair.  The hook adds, at the last-token position
only and only to the per-head slices of the selected pairs, a
scaled multiple of the probe direction:
\begin{equation}
\mathbf{z}_{\ell,h} \;\leftarrow\;
\mathbf{z}_{\ell,h} \,+\, \alpha\,\sigma^{\ell,h}_A\,
u^{\ell,h}_A,\;\; \alpha \in \{-10,\, 0,\, +10\},
\label{eq:card-hook}
\end{equation}
where $\alpha$ is unit-free (the additive magnitude is
$|\alpha|\,\sigma$ training-projection standard deviations) and
remaining heads are left untouched.  All $K$ heads are steered
simultaneously, so the intervention is multi-direction by
construction; the rest of the forward pass runs unchanged.  The VLM then answers a question on a \emph{different} axis $B \neq A$, yielding a
TB$-$FB accuracy gap $g_B(\alpha) =
\mathrm{acc}_B^{\mathrm{TB}}(\alpha) -
\mathrm{acc}_B^{\mathrm{FB}}(\alpha)$.  Steering with $\alpha{=}{+}10$
and $\alpha{=}{-}10$ should move this gap in opposite directions
if axis $A$'s direction causally influences the axis-$B$
prediction; we summarise the effect as the \emph{gap-shift} -- the
change in $g_B$ between the two steering polarities:
\begin{equation}
M_{A,B} \;=\; g_B(+10) - g_B(-10).
\label{eq:card-matrix-gap}
\end{equation}
We additionally report a mean-accuracy shift
$M^{\mathrm{acc}}_{A,B} = \max_{\alpha \in \{\pm 10\}}
\mathrm{acc}_B(\alpha) - \mathrm{acc}_B(0)$ to catch uniform
answer modulations a signed gap-shift would miss.  On Relay
Chain we populate the $3{\times}3$ matrix with $A, B \in
\{\text{belief, intent, knowledge}\}$: same-axis entries
($A=B$) serve as positive controls; off-diagonal entries test
whether the linearly-decodable $A$ direction is used by
the $B$ prediction.  A column $\{M_{A,B^{*}}\}_A \approx 0$
paired with at least one non-trivial row implies the $B^{*}$
prediction is \emph{decoupled from $A$'s linearly-decodable
direction} -- the routing failure.

\paragraph{Four-tuple signature.}

On the unified success+counter partition, for each
(VLM, protagonist, question axis) we record the four-tuple
$\big(\mathrm{acc}^{\mathrm{TB}}_{\rm succ},
\mathrm{acc}^{\mathrm{FB}}_{\rm succ},
\mathrm{acc}^{\mathrm{TB}}_{\rm cnt},
\mathrm{acc}^{\mathrm{FB}}_{\rm cnt}\big)$.  This signature
discriminates the three behaviours (constant-prior, world-state tracker, belief-conditional), which are indistinguishable on
success-only data.
%\endgroup

% 3. Negative control
\paragraph{Projection-removal control.}
% To rule out an amplitude-limited interpretation of an
% $M_{A,B} \approx 0$ cell, we instead \emph{subtract} a scaled
% projection of the per-head residual slice along
% $u^{\ell,h}_A$ at the same last-token position:
% \begin{equation}
% \mathbf{z}_{\ell,h} \,\leftarrow\,
% \mathbf{z}_{\ell,h} - c\,\sigma^{\ell,h}_A\,
% (u^{\ell,h}_A \!\cdot\! \mathbf{z}_{\ell,h})\,u^{\ell,h}_A,
% \;\; c \in \{0.5, 1, 2, 5\},
% \label{eq:proj-removal}
% \end{equation}
% where $c$ is a unit-free scalar.  At $c{=}1$ this reduces to
% canonical projection-removal scaled by training-$\sigma$;
% $c{>}1$ over-subtracts to test for amplitude-asymmetric coding.
% Removal at $c{=}5$ leaves the intent TB$-$FB gap unchanged
% ($|\Delta_{\rm gap}| < $ \projRemovalQwenAlphaFive\,pp).
As a negative-direction test, we subtract a scaled projection
of the per-head residual along $u^{\ell,h}_A$:
\begin{equation}
\mathbf{z}_{\ell,h} \leftarrow
\mathbf{z}_{\ell,h} - c\,\sigma^{\ell,h}_A\,
(u^{\ell,h}_A \cdot \mathbf{z}_{\ell,h})\,u^{\ell,h}_A,
\;\; c \in \{0.5, 1, 2, 5\}.
\label{eq:proj-removal}
\end{equation}
$c{=}1$ is canonical projection-removal; $c{>}1$ over-subtracts.

% \paragraph{Pre-registered prompt-resistance test.}
% We ask whether prompt engineering alone can recover
% belief-conditional action without an architectural change.
% Twelve prompt variants on top of the canonical intent question
% in four families (cost framing, chain-of-thought,
% perception-first / rule-based, in-context exemplars; full text
% in Appendix~\ref{app:ta-variants}) are evaluated across
% $\taTotalCells$ pre-specified cells.  The falsifier
% (existential, across the cell grid): a single cell on a
% lever-holder ($a_1$ or $a_2$) with both
% $\mathrm{acc}^{\mathrm{TB}} > 50\%$ and
% $\mathrm{acc}^{\mathrm{FB}} > 50\%$ under the same prompt --
% i.e., the model correctly releasing under TB and correctly
% holding under FB.

\paragraph{Prompt-resistance test.}
We ask whether prompt engineering alone can recover
belief-conditional action.  Twelve variants (cost framing,
chain-of-thought, perception-first / rule-based, in-context
exemplars; Appendix~\ref{app:ta-variants}) are evaluated across
$\taTotalCells$ cells.  A variant passes on a lever-holder
($a_1$ or $a_2$) if both $\mathrm{acc}^{\mathrm{TB}} > 50\%$
and $\mathrm{acc}^{\mathrm{FB}} > 50\%$ -- correctly releases
under TB and correctly holds under FB.

\paragraph{Evaluation protocol.}
Each axis question is presented as a binary choice and parsed
from JSON.  Probes, top-$K$ selection, and CARD evaluation share
the \emph{unified success+counter} partition ($444$ scenarios),
split $50/50$ into a disjoint \emph{probe-set} ($\nProbeSet$
scenarios; probe training and top-$K$ ranking) and
\emph{eval-set} ($\nEvalSet$ scenarios; CARD, projection-removal,
behavioural baselines), stratified by protagonist $\times$
world-state outcome (App.~\ref{app:probe-extract}).  Accuracy is
reported separately for TB vs FB and, where relevant, separated
into success vs counter blocks.
% Seed protocol: single-seed, two-seed T-A and
% CS-CARD-T (seeds $\{42, 43\}$).

\section{Dataset and Task}\label{sec:task}
% \fabian{I would start with defining your method "CARD", not the environment you evaluate it on.}
%\andreas{motivation for proposing and evaluating on this dataset?}

\paragraph{Cooperative Relay Chain dataset.}
We evaluate CARD (\autoref{sec:methods}) on \emph{Relay Chain} -- a novel three-agent cooperative grid-world environment (see \autoref{fig:relay_example}).
CARD's diagnostic logic -- steer one axis and read out a \emph{different} prediction -- requires a testbed in which a
partner's mental state causally drives an action prediction that is distinct from the belief prediction that it steers.
While GridToM~\citep{li2025black}
%, the closest multimodal-ToM resource that
pairs probes with activation interventions and supplies multi-agent scenarios and belief-attribution questions (first- and second-order beliefs), it does not offer a cooperative action axis: All questions only ask what an agent believes but not what the protagonist (the
ego agent whose mental state we probe) should do given the partner's belief.
%\andreas{there is an interesting link to Matteo's work (Bortoletto, Protom): He suggests, as one direction for future work, that one should move away from only ToM prediction to also action selection}
The belief-to-action CARD cell is therefore undefined -- not because a partner is absent, but because there is no action prediction that depends on the partner's mental state.

In contrast, on the Relay Chain, two agents ($a_1, a_2$) each hold a lever that opens one gate; a \emph{traveller} ($a_3$) must cross both gates in sequence to reach the goal.
The cooperative structure is a \emph{relay}: lever-operators must keep their gates open in sequence as the traveller passes through.
%, the partner's perceptual state directly determines the optimal action.
%Relay Chain's cooperative relay structure supplies exactly this missing axis: 
Crucially, as no single agent can solve this task alone, the traveller's optimal action is well-defined only relative to the partner's perception, turning belief-to-action into a well-defined test of routing.

\paragraph{Formal task description.}
% \andreas{I don't find this description particularly accessible. Introducing a formalism is fine/can make sense but then you also must explain each variable/part and how they are linked to each other.}
% A scenario $s$ specifies an initial layout $w_0$, a deterministic
% $T$-step cooperative trajectory $\tau = (w_0, \ldots, w_T)$ with
% $T \in [16, 32]$, and a protagonist \andreas{ego agent?} $a_p \in \{a_1, a_2, a_3\}$.
% The model input is a $k$-frame visual rendering subsampled from $\tau$ at relay-phase boundaries ($a_1$ acquires lever, $a_2$ acquires lever, $a_3$ reaches goal) plus midpoints, with $k \in [4,8]$ (see Appendix~\ref{app:gen}).
% We pair $s$ across True-Belief (TB) and False-Belief (FB) layouts $(s^{\mathrm{TB}}, s^{\mathrm{FB}})$ that hold the trajectory, captions, and agent identities fixed and flip only $a_p$'s perceptual access to the cooperative event $e \in \{0,1\}$; a \emph{counter}-pair replaces $s$ with $s'$ in which $e = 0$.
% For each axis $\alpha \in \{\text{belief, intent, knowledge}\}$ a
% textual question $q_\alpha(a_p)$ carries a binary label $y_\alpha(s, b)$ deterministically computed from $(w_0, \tau, e, b)$, where $b \in \{\mathrm{TB}, \mathrm{FB}\}$; the VLM $M$ produces a prediction $\hat{y}_{M,\alpha}(s, b)$.

%

We treat each scenario as a tuple
$s = (w_0, \tau, a_p)$.  $w_0$ is the initial grid layout
(agents, levers, gates, goal).
$\tau = (w_0, w_1, \ldots, w_T)$ with $T \in [16, 32]$ is the
deterministic cooperative trajectory -- the joint multi-agent
state at each step.
$a_p \in \{a_1, a_2, a_3\}$ is the protagonist -- the ego agent whose mental state we probe.
The VLM does not see $\tau$ directly: it receives a
$k$-frame visual rendering ($k \in [4,8]$) subsampled at
relay-phase boundaries ($a_1$ acquires lever, $a_2$ acquires
lever, $a_3$ reaches goal) plus midpoints
(see Appendix~\ref{app:gen}).

\emph{Belief variants.}  Each scenario appears in two layouts
that hold $\tau$ and agent identities fixed and differ only in
$a_p$'s observation of the cooperative event ($a_p$'s
partner releases/crosses): True-Belief
($s^{\mathrm{TB}}$, event within $a_p$'s field of view) and
False-Belief($s^{\mathrm{FB}}$, event in fog of war).
Let $e \in \{0,1\}$ indicate whether the event in fact
\emph{occurs} in the world.  The \emph{counter pair} replaces
$s$ with $s'$ where $e = 0$ (the lever is never released, the
gate never opens), preserving the same TB/FB perceptual
structure.

\emph{Questions and labels.}  For each axis
$A \in \{\text{belief}, \text{intent}, \text{knowledge}\}$ we
ask one binary question $q_A(a_p)$ whose ground-truth label
$y_A(s, b) \in \{0, 1\}$ is a deterministic function of
$(w_0, \tau, e, b)$, where $b \in \{\mathrm{TB}, \mathrm{FB}\}$
selects the belief variant.  \emph{Belief} asks whether the
protagonist \emph{perceived} the cooperative event; \emph{knowledge} asks whether they can \emph{justify} that it happened (perception plus integration of frame evidence); \emph{intent} asks the next cooperative action.
The VLM $M$ outputs a prediction $\hat{y}_{M,A}(s, b)$ that we score against $y_A$.  Per-axis label logic is in Appendix~\ref{app:groundtruth}.

Full construction details (scenario generator, paired layouts, counter-scenarios, per-axis ground-truth labels) can be found in Appendix~\ref{app:dataset}.

%\andreas{the following has to come earlier -- in short form already in the introduction where you motivate/introduce the contributions. But the following paragraph definitely right at the beginning of this section}

\paragraph{Counter scenarios.}

On success-only data (the typical ToM benchmark setup, where
the cooperative event always occurs), a model that emits the
safe default on every input -- ``don't release'' for
lever-operators, ``keep waiting'' for the traveller -- is
correct on the majority of cells, so high accuracy is not
evidence of belief-conditional reasoning.  The counter pair
($e = 0$) breaks this shortcut: the safe default becomes wrong
precisely where it was right under success, so constant-prior
and belief-conditional models produce visibly different
accuracy patterns.
Construction mechanics and per-axis counts can be found in
Appendix~\ref{app:counter}, \autoref{tab:dataset_counts}.
Figure~\ref{fig:relay_example} illustrates the overall layout.

\begin{figure*}[!t]
    \centering
    \includegraphics[width=0.80\textwidth]{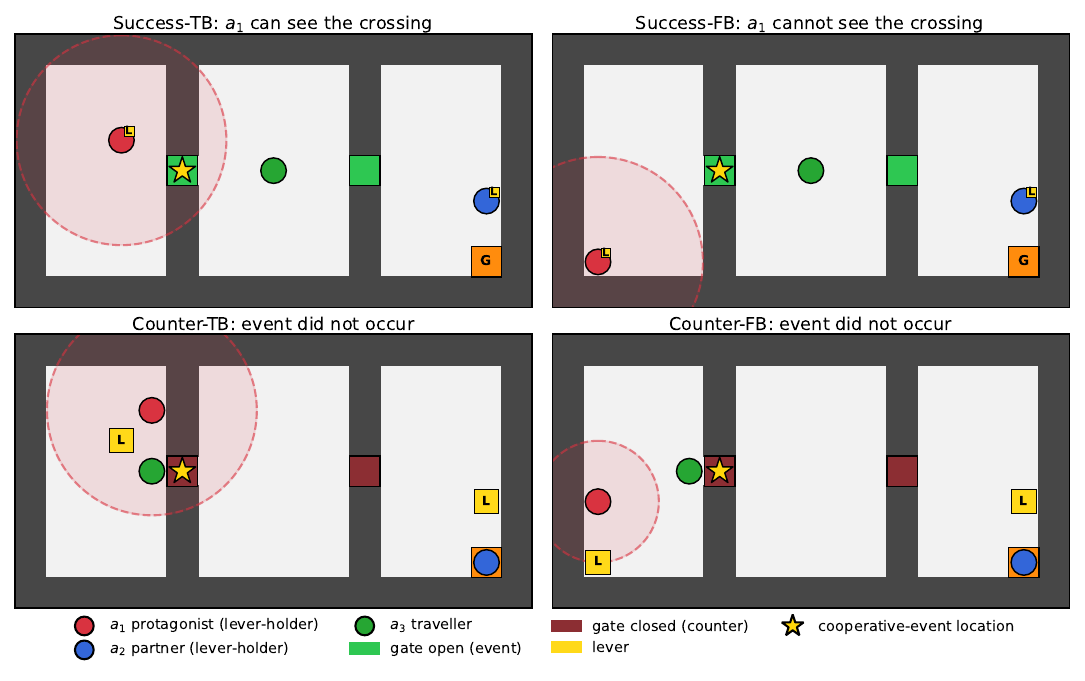}
    % \caption{\andreas{figure headings: How can a "lever see"?}
    % Relay Chain four-cell discrimination layout.  Each
    %   panel shows the protagonist $a_1$ (red, lever-operator) with a
    %   translucent fog-of-war disk \andreas{fog of war is exactly the opposite of the disk -- it's the area one cannot see while the disk is the visible area/field of view/vision}, partner $a_2$ (blue,
    %   lever-operator), traveller $a_3$ (green), the two gates (green
    %   when the cooperative event has opened them, dark-red when
    %   closed), the levers $L$, and the goal cell $G$.  Top row:
    %   \textbf{success} \andreas{don't use bold within text, whether captions or main text. Use italic instead but also that sparingly} scenarios in which the cooperative event
    %   occurs.  Bottom row: \textbf{counter} scenarios in which the
    %   lever-operator fails to release the lever at the cooperative
    %   deadline, so the gate never opens for the traveller to cross.  Left column: \textbf{True-Belief
    %   (TB)} layout ($a_1$'s lever within fog-of-war of the gate
    %   it controls); right column: \textbf{False-Belief (FB)} layout
    %   (lever beyond fog-of-war).  Captions, trajectories,
    %   identities, and the world-state event are held constant
    %   within each row; only lever placement differs across the
    %   pair.}
    %\label{fig:relay_example}
    \caption{
    Relay Chain four-tuple discrimination layout.
      Panel elements: protagonist $a_1$ (red, lever-operator)
      with field-of-view disk (radius $5$); partner $a_2$ (blue,
      lever-operator); traveller $a_3$ (green); gates (green
      when open, dark-red when closed); levers $L$; goal $G$.
      Top row: \emph{success} (event occurs).  Bottom row:
      \emph{counter} (lever-operator never releases; gate stays
      closed).  Left column: True-Belief (TB) -- $a_1$'s lever
      close enough to perceive the crossing.  Right column:
      False-Belief (FB) -- lever beyond $a_1$'s view.
      % Trajectories, identities, and the world-state event are
      % held constant within each row; only lever placement
      % differs across the pair.
      \label{fig:relay_example}
      }
  \end{figure*}

\paragraph{Three axes.}
Building on the belief--desire--intention framework
\citep{bratman1987intention} and the percept--belief--intention
(PBI) causal structure used in recent ToM benchmarks
\citep{bortoletto2025tom}, we instrument each scenario along three
axes:
\textit{belief} -- does the protagonist see the cooperative event
(first-order PBI percept; what content the agent
holds in mind?);
\textit{intent} -- in the BDI sense of \emph{present-directed
action-recommendation}: should the protagonist take the
cooperative action now (PBI intention);
\textit{knowledge} -- does the protagonist \emph{know whether}
the cooperative event occurred 
% (the \emph{knows-whether} attitude
% from epistemic logic, 
separate from the propositional content of
belief).
Knowledge serves as a control: within counter scenarios (where
the world did not change), it is the only axis whose label still
flips between TB and FB -- with observation alone, while belief
and intent labels stay constant.  Including knowledge in the
probe-extraction pool forces the learned direction to
track observation rather than the world-state outcome.
Appendix~\ref{app:cos} verifies that the three probe directions are
neither collinear nor orthogonal.

% \paragraph{Auxiliary dataset.}
% For the hook positive-control we re-used the same CARD
% pipeline on CLEVR~\citep{johnson2017clevr}, a
% single-image visual-reasoning benchmark, with three
% object-property axes \{count, shape, colour\} replacing the ToM
% axes (see~\ref{tab:clevr_card} in the Appendix).

\paragraph{Models.}
Four open-weight VLMs: Qwen2-VL-7B \citep{Yang2024Qwen2TR},
Gemma-4-E4B-it \citep{gemma4_2026}, LLaVA-NeXT-Video-7B-hf
\citep{zhang2024llavanextvideo}, and InternVL2.5-8B
\citep{chen2024expanding}.  We used multi-frame inputs ($4$--$8$
keyframes per scenario).

\section{Results}\label{sec:results}

We conducted a series of experiments that together
show a \emph{routing failure} on the cooperative-action
axis.  
% \andreas{what is F?}
% \andreas{avoid references to sections in these kind of papers (and ideally in scientific papers in general unless they are super long)}
% \andreas{if at all then refer to concrete figures or tables instead of Fx and sections}
% \todo{remove}F1 (\S\ref{sec:f1}) verifies belief is linearly
% decodable and used on the belief question (positive control).
% \todo{remove} (\S\ref{sec:}) establishes a behavioural dissociation: the
% same model that reads belief on the belief question collapses
% to a constant per-protagonist prior on the intent question.
% \todo{remove}F3 (\S\ref{sec:f3}) supplies the mechanistic evidence: CARD
% steering along the linearly-decodable belief direction
% substantially shifts the belief and knowledge predictions but
% leaves the cooperative-action prediction essentially unchanged.
% Three ablations (\todo{remove}\S\ref{sec:ablation}) close the natural
% alternative explanations: \todo{remove}A1 (projection-removal,
% \S\ref{sec:f4}) rules out amplitude limitation, \todo{remove}A2
% (pre-registered prompt-resistance, \S\ref{sec:f5}) rules out a
% prompting artefact, and \todo{remove}A3 (CLEVR, \S\ref{sec:a3}) rules out a
% globally-broken steering hook.
First, we verify belief is linearly decodable and used on
the belief question (positive control).
We then establish a behavioural dissociation: the same model
that reads belief on the belief question collapses to a
constant per-protagonist prior on the intent question.
CARD steering along the linearly-decodable belief direction
substantially shifts the belief and knowledge predictions but
leaves the cooperative-action prediction essentially unchanged
-- the mechanistic evidence. Three ablations close the natural alternative explanations: projection-removal rules out amplitude limitation, a prompt-resistance test rules out a prompting artefact, and CLEVR rules out a globally-broken steering hook.

\paragraph{Belief is decodable and used on the belief question.}
\label{sec:f1}

% We confirm belief is encoded and used on its canonical
% question.
% \andreas{what is the canonical belief question?}
% A linear probe on the TB-vs-FB activation difference
% reaches held-out accuracy well above chance on every VLM
% (\autoref{tab:f1}).  We rule out spurious correlations with a
% Hewitt--Liang random-label control
% \citep{hewitt2019designing}, which collapses probe accuracy to
% chance (Appendix \autoref{tab:random_label}).
% Behaviourally, every VLM answers the canonical belief question
% near ceiling on the TB layout and much lower on FB
% (\autoref{tab:f1}) -- consistent with the prediction
% reading event occurrence but not gating by the protagonist's
% field of view.  We thus rule out the linearly-absent-representation
% null: any later failure on the intent question cannot be
% blamed on missing or unreadable belief.
We confirm belief is encoded and behaviourally used on the
belief question -- a binary read-out of whether the protagonist
saw the cooperative event (``Does $a_p$ see the partner
release / cross?'').  A linear probe on the
TB-vs-FB activation difference reaches held-out accuracy well
above chance on every VLM (see \autoref{tab:f1}).  We rule out
spurious correlations with a Hewitt--Liang random-label control
\citep{hewitt2019designing}, which collapses probe accuracy to
chance (see \autoref{tab:random_label} in the Appendix).
Behaviourally, every VLM answers the same question near ceiling
on the TB layout and much lower on FB (see \autoref{tab:f1}) --
consistent with the prediction reading event occurrence but
not gating by the protagonist's field of view.  We thus rule
out the possibility that belief is simply not represented: any
later failure on the intent question cannot be blamed on
missing or unreadable belief.

\begin{table}[t]\small\centering
    \begin{tabular}{lccc}
      \toprule
      \textbf{Model} & \textbf{Probe acc.} & \textbf{TB-acc.} & \textbf{FB-acc.} \\
       \midrule
      Qwen2-VL    & $\probeAccQwenBeliefMS$   & $\fOneBeliefTbQwen$   & $\fOneBeliefFbQwen$   \\
      Gemma-4     & $\probeAccGemmaBeliefMS$  & $\fOneBeliefTbGemma$  & $\fOneBeliefFbGemma$  \\
      LLaVA-NeXT  & $\probeAccLlavaBeliefMS$  & $\fOneBeliefTbLlava$  & $\fOneBeliefFbLlava$  \\
      InternVL2.5 & $\probeAccInternBeliefMS$ & $\fOneBeliefTbIntern$ & $\fOneBeliefFbIntern$ \\
      \bottomrule
    \end{tabular}
    \caption{Positive control: linear-probe accuracy (mean
      $\pm$ std across three seeds; chance $=50\%$) and
      behavioural TB/FB accuracy on the belief
      question.}
    \label{tab:f1}
  \end{table}

\paragraph{Belief is used on belief and knowledge but ignored on intent questions.}
\label{sec:f2}

We find a three-way behavioural dissociation on Relay Chain:
the three axis questions, asked of the same model on the same
scenarios with the same activations, produce qualitatively
distinct accuracy patterns across all four VLMs (see the per-VLM
matrix in \autoref{tab:d2_full} in the Appendix).

On the \emph{belief question} every VLM reproduces the
\emph{world-state tracker} signature: across the four
(success/counter) $\times$ (TB/FB) cells, the answer is
correct on three (success-TB, counter-TB, counter-FB) and
wrong only on success-FB, where the event occurred but outside
the protagonist's field of view (per-VLM breakdown in
\autoref{tab:d2_full} in the Appendix).  We read this as the prediction using
event occurrence but not gating by observation.

On the \emph{knowledge question} the same models become
behaviourally belief-conditional in the success block: TB is
answered ``knows'' and FB is answered ``doesn't know'',
correctly tracking the protagonist's observation rather
than the world state alone.  We conclude observation is
at least partially consulted for this question.

On the \emph{intent question} the same models collapse to a
\emph{constant per-protagonist prior} (``don't release'' on
every cell, incorrect only on success-TB where releasing is
the right action).  The constant-prior signature is
indistinguishable across the four VLMs -- a striking
convergence given otherwise divergent probe accuracies and
architectures, and the strongest sign that intent answers are
generated without consulting either event occurrence or
observation.

We thus observe three distinct signatures on the same model
and the same activations: belief-Q reads world state,
knowledge-Q reads observation, and intent-Q reads
neither.  The intent question alone is behaviourally
belief-insensitive -- the lonely outlier among the three
axes.  We next probe this dissociation mechanistically with
CARD, asking whether the structural cause is upstream (belief
unreachable from the intent prediction) rather than behavioural
(belief read but ignored).

  \begin{table}[t]\small\centering
    \setlength{\tabcolsep}{5pt}
    \begin{tabular}{l c c c}
      \toprule
      \textbf{Probe} & \textbf{Belief Q (\%)} & \textbf{Intent Q (\%)} & \textbf{Epist.\ Q (\%)} \\
      \midrule
      \multicolumn{4}{l}{\emph{Qwen2-VL}} \\
       \midrule
      belief    & \flipRateQwenBeliefBelief & \flipRateQwenIntentBelief & \flipRateQwenEpistBelief \\
      intent    & \flipRateQwenBeliefIntent & \flipRateQwenIntentIntent & \flipRateQwenEpistIntent \\
      knowledge & \flipRateQwenBeliefEpist  & \flipRateQwenIntentEpist  & \flipRateQwenEpistEpist  \\
      \midrule
      \multicolumn{4}{l}{\emph{Gemma-4}} \\
      \midrule
      belief    & \flipRateGemmaBeliefBelief & \flipRateGemmaIntentBelief & \flipRateGemmaEpistBelief \\
      intent    & \flipRateGemmaBeliefIntent & \flipRateGemmaIntentIntent & \flipRateGemmaEpistIntent \\
      knowledge & \flipRateGemmaBeliefEpist  & \flipRateGemmaIntentEpist  & \flipRateGemmaEpistEpist  \\
      \midrule
      \multicolumn{4}{l}{\emph{LLaVA-NeXT}} \\
      \midrule
      belief    & \flipRateLlavaBeliefBelief & \flipRateLlavaIntentBelief & \flipRateLlavaEpistBelief \\
      intent    & \flipRateLlavaBeliefIntent & \flipRateLlavaIntentIntent & \flipRateLlavaEpistIntent \\
      knowledge & \flipRateLlavaBeliefEpist  & \flipRateLlavaIntentEpist  & \flipRateLlavaEpistEpist  \\
      \midrule
      \multicolumn{4}{l}{\emph{InternVL2.5}} \\
      \midrule
      belief    & \flipRateInternBeliefBelief & \flipRateInternIntentBelief & \flipRateInternEpistBelief \\
      intent    & \flipRateInternBeliefIntent & \flipRateInternIntentIntent & \flipRateInternEpistIntent \\
      knowledge & \flipRateInternBeliefEpist  & \flipRateInternIntentEpist  & \flipRateInternEpistEpist  \\
      \bottomrule
    \end{tabular}
    \caption{
    % Per-scenario answer change rate under
    %   $\alpha{=}{\pm}10$ steering on lever-operator $a_1$ ($288$
    %   scenarios per cell).  Each cell reports the fraction of
    %   scenarios whose argmax answer differs between
    %   $\alpha{=}{+}10$ and $\alpha{=}{-}10$.  The intent column
    %   flip rate is $\mathbf{0/288}$ on $10$ of the $12$
    %   (VLM, probe) cells (Qwen, Gemma, LLaVA across all three
    %   probes; InternVL2.5 under the intent probe) and at most
    %   $5.9\%$ on the remaining two InternVL2.5 cells, where the
    %   flips are direction-balanced and cancel in net accuracy
    %   (intent $\Delta_{\rm acc}$ remains within
    %   $+0.0$--$+8.7$\,\%).  Sibling
    %   columns change up to $51.7\%$ of scenarios under the same
    %   hooks.  Without steering, Qwen $a_1$'s intent answer is
    %   mixed ($\sim 49\%$ yes / $51\%$ no), so the model is
    %   genuinely choosing between both options -- the null is
    %   not because the answer was already pinned to one side.
    %   \label{tab:per_scenario}
    Per-scenario flip rate on $a_1$: of $144$ scenarios, the
      fraction whose answer differs between $\alpha{=}{+}10$ and
      $\alpha{=}{-}10$ steering (i.e., whether steering moves the
      model \emph{at all}, before changes average out).  
      In the intent column (middle), $9$ of $12$ (VLM, probe) entries are
      exactly $0$; only three are above zero (Gemma-4 and
      InternVL2.5 on the belief probe, InternVL2.5 on the
      knowledge probe; ${\le}15.3\%$), with direction-balanced
      flips that cancel in net accuracy
      (\autoref{tab:intent_delta} in the Appendix).  The belief and
      knowledge columns flip many scenarios under the same hook.
      \label{tab:per_scenario}
      }
  \end{table}
  
  \begin{figure*}[!t]
    \centering
    \includegraphics[width=0.8\textwidth]{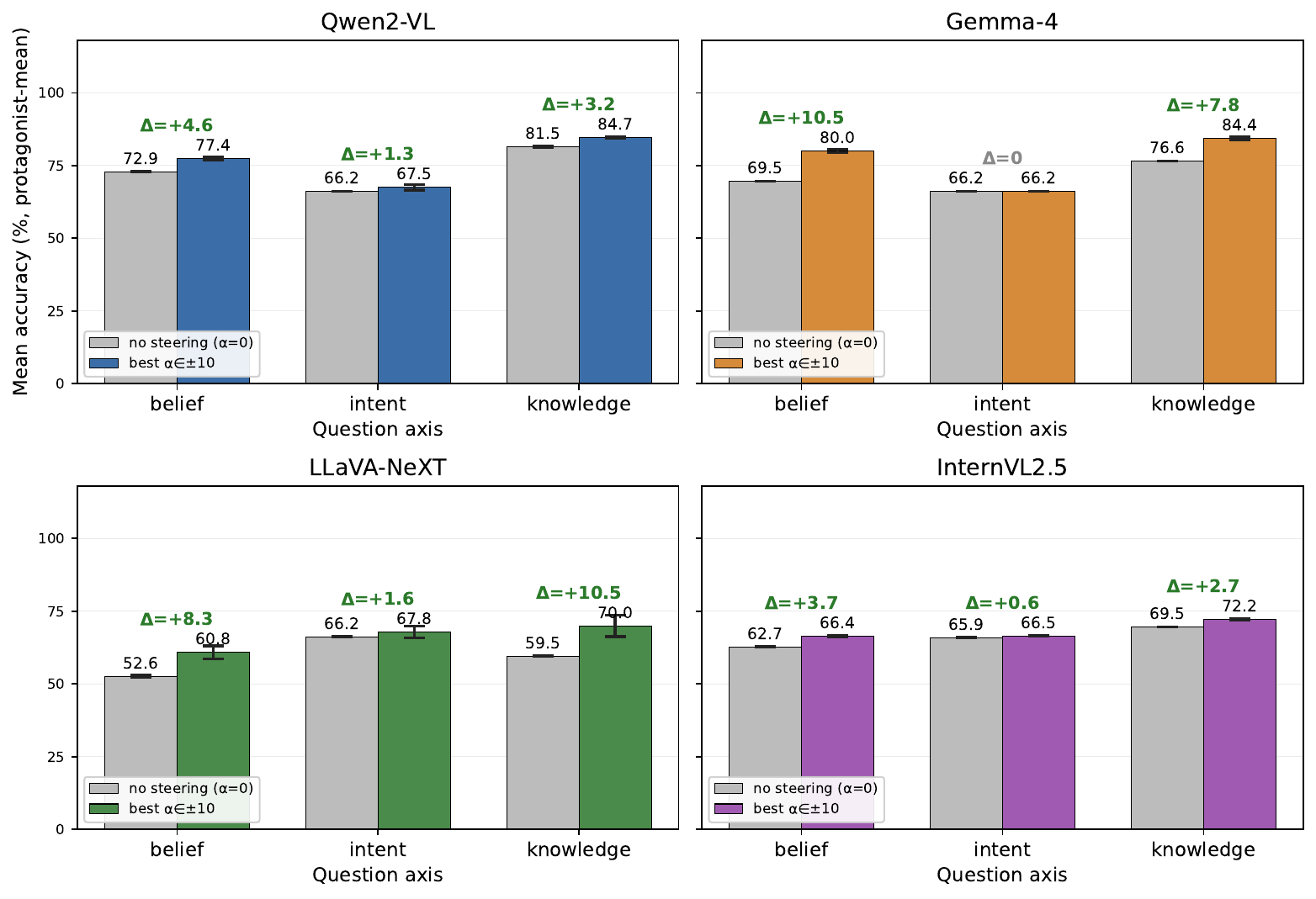}
    \caption{Steering improves accuracy on the belief and
      knowledge predictions, but never on the intent answer
      prediction, averaged across the three protagonists ($a_1$,
      $a_2$, $a_3$).  For each (VLM, question axis) we report
      the mean TB/FB accuracy with no steering ($\alpha{=}0$,
      grey) and with the best-case steering
      ($\alpha\in\{-10,+10\}$ across all probe directions, in
      VLM color).  $\Delta$ on top is best$-$baseline.  Belief
      and knowledge improve by $+2.7$ to $+10.6$\,pp; the
      intent $\Delta$ is essentially $0$ on every VLM.  The
      identical intent baseline ($\sim 66\%$) across VLMs
      reflects the  constant-prior signature.  Per-protagonist
      breakdown in \autoref{fig:card_per_agent} in the Appendix.}
    \label{fig:card_dual}
  \end{figure*}

\paragraph{CARD intent column is uniformly null (routing failure).}
\label{sec:f3}

We find a clean three-way dissociation under CARD steering.
Across the full $4{\times}3{\times}3 = 36$-cell sweep (four VLMs
$\times$ three protagonists $\times$ three probe directions; see 
\autoref{tab:intent_delta} in the Appendix), the best-$\alpha$ intent
accuracy stays essentially at its no-steering baseline on
nearly every cell, while the same steering hooks substantially
shift the belief and knowledge predictions.  Averaged across
protagonists (see \autoref{fig:card_dual}), intent is flat on
every VLM and the belief and knowledge axes move; the
per-(VLM, protagonist) breakdown (see \autoref{fig:card_per_agent} in the Appendix)
shows the intent null holds across all 12 cells, not just the
target lever-operator $a_1$. The gap-shift matrix on $a_1$
(see \autoref{fig:card3x3}) tells the same story at the gap-shift
level.

We further show the null is per-scenario tight.  On $a_1$ the
intent argmax is invariant to $\alpha{=}{\pm}10$ steering on
the overwhelming majority of scenarios across all four VLMs
(see \autoref{tab:per_scenario}); the small fraction that do flip
(InternVL2.5 and Gemma-4) are direction-balanced and cancel in
net accuracy, while the belief and knowledge questions flip a
substantial fraction of scenarios under the same hooks.  We
rule out the possibility that the intent answer was already
pinned to one side before steering: without steering, the
intent baseline distribution is mixed, so the null cannot be
explained that way. Analysis confirms the null is statistically tight:
across $\cardSignTestIntentNTotal$ intent measurements, the largest improvement in mean intent accuracy is bounded at $\cardSignTestIntentMaxDeltaHi$\,pp at $95\%$ confidence, versus $\cardSignTestBeliefMaxDeltaHi$\,pp on the belief
positive control (Appendix~\ref{app:stats}).

% Across all intent measurements (4 VLMs
% $\times$ 3 protagonists $\times$ 3 probes $\times$ 3 seeds),
% steering improves intent accuracy by at most
% $\cardSignTestIntentMaxDeltaHi$\,pp ($95\%$ paired-bootstrap
% upper bound), vs $\cardSignTestBeliefMaxDeltaHi$\,pp on the
% belief control.  The intent answer remains invariant to
% $\alpha$ in
% $\cardSignTestIntentNZero/\cardSignTestIntentNTotal$ of these;
% in the $42$ that do shift, the shift skews toward the
% constant-prior default ($\cardSignTestIntentNNeg$ vs
% $\cardSignTestIntentNPos$,
% Wilcoxon $p = \cardSignTestIntentWilcoxonP$) -- steering
% pushes intent toward its default, not toward belief-conditional
% behaviour.

We interpret this as a structural \emph{routing failure}: the
linearly-decodable belief direction is encoded, readable on the
belief and knowledge predictions, but not used in the action prediction.
\figCardThreeByThree

\paragraph{Ablation studies.}\label{sec:ablation}
% \andreas{you write four but then only three "subsections" follow -- and their labels are different from how you call them here. $\rightarrow$ unify}
We finally evaluated our findings against four alternative explanations:
amplitude-limited steering (projection-removal control), a prompting artefact, cross-axis steering on a non-ToM task (CLEVR), and a last-token-locus artefact (distributed-token steering).

\subsubsection*{Projection-removal control}
\label{sec:f4}

An alternative is that our additive steering simply does not push the belief subspace hard enough to register at the intent prediction, and a stronger or differently-shaped perturbation would.  We test the opposite limit: instead of adding belief, we \emph{remove} it from the residual stream entirely by subtracting its $c\,\sigma$-scaled projection from the per-head slice (Eq.~\ref{eq:proj-removal}) at $c \in \{0.5, 1, 2, 5\}$. The intent TB$-$FB gap stays essentially flat at every $c$ on every VLM with a complete sweep, while the same removal shifts the \emph{belief}-question gap toward chance as expected (positive control).  The intent null survives both directions of perturbation, so it is structural, not a consequence of under-driving the belief subspace.

\subsubsection*{No prompt achieves belief-conditional action}
\label{sec:f5}
% Another obvious artifact is that the intent prediction already
% routes belief but only when the prompt scaffolds it:
% chain-of-thought, perception-first cues, an explicit cost
% reminder, or in-context examples could be the missing
% ingredient, and the null we observed would then reflect prompt
% design rather than a structural gap.  We
% test this with a pre-registered $\taTotalCells$-cell sweep
% covering exactly these families (12 prompt variants spanning
% CoT, perception-first, rule-based, high-stakes cost, and 1/2-shot
% in-context; $\times$ four VLMs $\times$ four protagonist conditions;
% \autoref{fig:ta_heatmap} in the Appendix).  Only $\taLeverHolderCells$
% seed-$42$ cells clear the strict
% $\{\mathrm{acc}^{\rm TB} > 50,\, \mathrm{acc}^{\rm FB} > 50\}$
% falsification threshold on a lever-operator, and the largest
% TB$>$FB margin we observe is an order of magnitude below the
% belief-Q baseline on the same scenarios.  A second-seed
% replication erases even those: both borderline seed-$42$ cells
% fail to reproduce and no new cell passes.  No prompt in our
% sweep recovers belief-conditional action.
Another possibility is that the intent prediction routes
belief only when scaffolded by the right prompt.  We test this
with a pre-registered sweep of 12 prompt variants (CoT,
perception-first, rule-based, high-stakes cost, 1/2-shot
in-context) $\times$ four VLMs $\times$ four protagonist
conditions ($\taTotalCells$ total; \autoref{fig:ta_heatmap}
in the Appendix).  
% Only $\taLeverHolderCells$ seed-$42$ cells clear
% the $\{\mathrm{acc}^{\rm TB} > 50,\, \mathrm{acc}^{\rm FB} > 50\}$
% threshold on a lever-operator, and a second-seed replication
% erases even those. 
No prompt recovers belief-conditional action.

\subsubsection*{Cross-axis Steering on Non-ToM task}
\label{sec:a3}

% Finally, the steering hook itself is
% the problem -- too weak, mis-wired, or unable to register at
% any prediction -- and the intent null is diagnosing the hook
% rather than the model.  We test this on a different task family
% where the target axis is unambiguous.  Applying CARD to
% CLEVR \citep{johnson2017clevr} (single-image
% visual-reasoning; axes \{count, shape, colour\}), the same hook
% on the same four VLMs produces substantial best-$\alpha$
% accuracy gains on every VLM with headroom (see
% \autoref{tab:clevr_card} in the Appendix; the only flat cells are ceiling
% effects on Qwen2-VL and InternVL2.5 shape/colour).  If the
% hook were globally broken, CLEVR would also be flat; it is not.
% The intent null is therefore a target-axis property, not a
% hook artefact.

Finally, the hook itself could be at fault -- too weak to
register anywhere.  Applying the same hook to CLEVR on the
same models (\autoref{tab:clevr_card} in the Appendix) yields
substantial best-$\alpha$ accuracy gains where there is
headroom; the few flat cells reflect ceiling effects on shape
and colour.  The intent null is therefore a target-axis
property, not a hook artefact.

\subsubsection*{Distributed-token steering}
\label{sec:cs-card-t}

A further possibility is that the intent prediction reads
belief at a different token position than our
last-token hook.  We re-run CARD with the steering
hook fired at two additional positions (\textit{vision-end} and
a \textit{vision-plus-last} slice covering both) in addition to
the \textit{last-token} target, using three probe
sources: the paired-statement belief and intent
probes plus a per-position condition probe trained at
\textit{vision-end}. 
Sweeping $\alpha \in \{-10, +10\}$ yields $18$
(probe-source $\times$ target-position) measurements. 
Across all, the maximum intent $|\Delta_{\rm acc}|$ vs the $\alpha{=}0$ baseline is $0.3$\,pp; maximum shift is around
$1$\,pp (see \autoref{fig:cscardt} in the Appendix).  The result is therefore not an artefact of the last-token hook position.

\paragraph{Summary.}
We established that belief is linearly decodable and used on
the belief question, then a behavioural dissociation between
belief- and intent-question signatures, then the causal
dissociation under intervention, and finally closed the three
alternative explanations (amplitude, prompt, hook).
Belief direction is encoded, readable on the belief and
knowledge predictions, and not used by the
action prediction -- a routing failure, not an absence.
\section{Discussion}\label{sec:discussion}

The simplest account is that the cooperative-action prediction
does not use the linear belief direction.  Additive steering
and projection-removal both shift the belief and knowledge
predictions but leave the intent prediction essentially
unchanged on every VLM and every protagonist; the same
prediction defaults to a constant per-protagonist prior that
the visual evidence does not modulate.  Two observations rule
out the simpler alternative that the prediction is just broken
or saturated.  First, the prediction \emph{does} vary across
inputs --- its prior differs by protagonist (``don't release''
on $a_1, a_2$; ``go'' on $a_3$) --- so it is not globally
stuck on one answer.  Second, the belief signal \emph{is}
present in the activations: the belief and knowledge
predictions, exposed to the same residual stream, do use it.
The prediction is therefore capable of varying and the
information it would need is available; the failure to use
belief is a routing fact specific to the intent prediction,
not a general inability to respond. 

% Read-out fidelity varies across VLMs (the LLaVA belief probe
% is the noisiest), but the intent null is invariant to this:
% it holds on every VLM, including the noisiest read-out, so the
% null is not contingent on having a clean belief decoder.

\section{Conclusion and Future work}\label{sec:conclusion}
%\andreas{past tense}

%\andreas{people don't remember the A1/A2, F3 etc I generally would advise against using them}
In this work we introduced CARD as a new cross-axis routing diagnostic that disambiguates "feature absent" from "feature encoded but unused by the action prediction". 
Using this new diagnostic we showed in a cooperative task that current VLMs encode a partner's belief well enough to answer the belief question, yet none usefully use that belief when asked what to do.
This represents a structural routing failure visible in current VLMs.
% Two natural extensions remain open: distributed-token interventions would test whether routing operates outside our last-token locus, and nonlinear probes would test whether the action prediction reads a subspace our linear probes miss.
% As such, our results show that verbal-ToM alignment does not imply action-ToM alignment -- training alone that aligns verbalised belief but leaves the action head's loss untouched cannot close this gap.
One natural extension remains open: nonlinear probes would test whether the action prediction reads a subspace our linear probes miss. As such, our results suggest that verbal-ToM alignment does not imply action-ToM alignment: alignment training that updates only verbalised belief predictions, without an explicit objective on the action prediction, would leave this routing gap unaddressed.

%Additive steering, subtractive projection-removal, a temporal sweep across vision-end / frame-end / last-token positions, and a pre-registered $\taTotalCells$-cell prompt-resistance sweep all converge on the same null at the intent prediction while the belief and knowledge predictions move.
% \andreas{do we need the following?}
%\paragraph{Connection to two-systems theory.}
%The dissociation mirrors the verbal-ToM / action-ToM split in cognitive science: children pass false-belief tasks verbally years before they reliably act on those beliefs in helping paradigms~\citep{wimmer1983beliefs,apperly2009minimal}.
%Our finding is a system-level analogue: the description-supporting representation is intact, but the action-supporting computation does not use it.

\section*{Limitations}\label{sec:limitations}

% We demonstrate the routing failure on
% one task family -- Relay Chain, a 2D grid-world --
% and four open-weight VLMs at the $4$--$8$B scale.  CARD and
% projection-removal require activation-stream access, so we
% cannot subject frontier closed-source VLMs to the mechanistic
% arms. We do not claim generality to larger model scales
% ($40$--$400$B, where applied-ToM behaviour is known to
% improve~\citep{gu2024simpletom,bortoletto2024brittle}), 
% % \andreas{missing references}
% non-cooperative ToM, or real-world video
% (\citet{li2025black} report a similar grid-vs-video gap).
% Finally, we report the intent-column null against
% \emph{linear} probe directions; a nonlinearly encoded subspace
% that the action head consumes would not be detected by CARD or
% projection-removal.
We demonstrate the routing failure on
one task family -- Relay Chain, a 2D grid-world --
and four open-weight VLMs.  CARD and
projection-removal require activation-stream access, so we
cannot subject closed-source VLMs to the mechanistic
arms; We do not claim generality to larger model scales
($40$--$400$B, where applied-ToM behaviour is known to
improve~\citep{gu2024simpletom,bortoletto2024brittle}), 
% \andreas{missing references}
non-cooperative ToM, or real-world video.

\section*{Ethical considerations}
% \fabian{They are optional for ACL ARR, but might increase acceptance chances. They also do not count towards the page limit: \url{https://aclrollingreview.org/cfp\#long-papers}}
% % \andreas{are these needed in the initially submitted version? I don't think so but best check}
The work uses publicly available open-weight VLMs and a
synthetic grid-world task with no human subjects or
personally-identifying data.  CARD reports an internal property
of the model and surfaces a limitation that deployers should be
aware of; we see no direct misuse pathway.  Datasets, prompts,
and code will be released under permissive licenses.

\bibliography{custom}

\clearpage

\appendix
\section{Dataset construction}\label{app:dataset}

\subsection{Dataset breakdown}\label{app:dataset-counts}
Relay Chain has $444$ unique scenario bases distributed across
three protagonist conditions ($a_1, a_2$ as lever-holders and
$a_3$ as traveller).  We render each base in two layouts that
preserve the trajectory but differ in the protagonist's
lever-to-gate Manhattan distance -- True-Belief
($\le 5$, lever within $a_p$'s field of view of the gate) and
False-Belief ($>5$, lever beyond)
-- yielding $888$ evaluation cases per axis
(\autoref{tab:dataset_counts}).  Scenarios split into
\emph{success} blocks where the cooperative event happens and
\emph{counter} blocks where it is omitted, so the same
protagonist appears once with and once without the cooperative
outcome on identical perceptual access.  The full $444$-scenario
partition is split $50/50$ into a disjoint probe-set and
eval-set (App.~\ref{app:probe-extract}).
\tabDatasetCounts

\subsection{Relay Chain scenario generator}\label{app:gen}
We render scenarios as $11\!\times\!11$ grids with three agents
($a_1, a_2, a_3$), two gates, and one goal cell.  Each scenario is
a $16$--$32$-step trajectory; we subsample $4$--$8$ keyframes
covering the cooperative event window.
Keyframes anchor on four relay-phase boundaries:
(i) episode start;
(ii) $a_1$ acquires lever A;
(iii) $a_2$ acquires lever B;
(iv) $a_3$ reaches the goal.
We then insert midpoints between consecutive boundaries to yield
up to $7$ distinct frames per scenario, following GridToM's
keyframe helper.  When a phase is never reached, the last
available step is used as a fallback so the boundary list stays
strictly monotonic.  Agents move on Manhattan-distance shortest
paths; lever-state transitions are scripted to produce the
target cooperative chain.  Random seeds are saved with each
scenario for reproducibility.

\subsection{Perceptual-access minimal pair}\label{app:paired}
For each scenario we generate two layouts that share trajectory
and event sequence but differ in protagonist's lever position
relative to the gate it controls.  TB: lever placed at
Manhattan distance $\le 5$ from the gate (within field of view).  FB:
lever at distance $> 5$.  All other state (agent identities, gate
positions, goal, traveller path) is identical.  We verify pair
matching by hashing the trajectory and confirming bit-exact
agreement on the non-protagonist state slice.

\subsection{Example scenario layouts}\label{app:scenario-examples}
Figure~\ref{fig:relay_scenarios_appendix} shows one True-Belief
and one False-Belief layout for each protagonist condition.  For
the lever-holders ($a_1, a_2$), TB places the protagonist's own
lever within Manhattan distance $5$ of the gate it controls
(inside the field-of-view disk); FB places it beyond.  For the
traveller ($a_3$), TB places the agent within view of the gate
ahead so the gate-state is perceivable; FB places it beyond.
All panels show the success block (both gates open), so the
contrast across columns is purely perceptual.
\figRelayScenariosAppendix

\subsection{Counter-story generator}\label{app:counter}
We construct the counter pair for each scenario by re-running
the scripted trajectory but \emph{omitting} the cooperative
event: the lime traveller stalls $3$ cells before the gate
(it never crosses), or the pink lever-holder fails to release
the lever at the cooperative deadline (the gate never opens
for the cooperative crossing).  We preserve the protagonist's
perceptual access (TB/FB layout), rewrite captions with the
asymmetric oracle-vs-pov rule (\S\ref{sec:methods}), and
recompute the binary correct answer by enumeration.

\subsection{Axis-specific ground-truth labels}\label{app:groundtruth}
We compute ground-truth labels deterministically from the
layout and event-outcome pair.  Belief: ``does the
protagonist see the event''.  Intent: ``should the
protagonist take the cooperative action now'', mapped to
release/keep-holding for lever-holders and continue/wait for
the traveller.  Knowledge: ``does the protagonist
have enough perceptual access to determine whether the
cooperative event occurred'' -- for lever-holders this is
``can the protagonist see the traveller'', for the traveller
it is ``can the protagonist see at least one lever-holder''.
The knowledge label depends only on perceptual visibility, not
on whether the event in fact occurred; success and counter
blocks therefore share the same knowledge label on each layout
(TB $\to$ TRUE, FB $\to$ FALSE), so a knowledge-conditional
answer is forced to track perceptual access rather than the
world-state outcome.

\section{Probe + behavioral protocol details}\label{app:protocol}

\subsection{Probe extraction}\label{app:probe-extract}
For each axis we extract per-(layer, head) activations at the
final token under each paired statement.  The difference
$\Delta_A = a(s_A^{+}) - a(s_A^{-})$ is the training input for
the linear probe; labels are $\pm 1$ for matched/mismatched
statements.  We use logistic regression with default $L^2$
regularisation ($C{=}1$).

\paragraph{K-selection / CARD-eval protocol.}
To prevent K-selection bias from probe training, top-$K$ ranking,
and CARD evaluation sharing scenarios, we split the unified
success+counter partition ($444$ scenarios) into a fully disjoint
\emph{probe-set} and \emph{eval-set} stratified $50/50$ by
protagonist $\times$ world-state outcome ($\nProbeSet$ scenarios
each).  The probe-set is used for probe training (single
$75/25$ stratified train/val split per (layer, head), seeded for
reproducibility) and top-$K{=}56$ cell ranking by held-out val
accuracy; CARD is evaluated only on the disjoint eval-set.  To
quantify probe-training variance we re-run with three seeds
$\{42, 43, 44\}$ for the internal $75/25$ split and report the
mean across seeds.

\subsection{CARD steering hook}\label{app:hook}
We implement the steering hook as a forward hook on each chosen
layer's self-attention output: at the last-token position only,
$\mathbf{z}_{\ell,h} \leftarrow \mathbf{z}_{\ell,h} +
\alpha\,\sigma^{\ell,h}_A\,u^{\ell,h}_A$
(Eq.~\ref{eq:card-hook}), with $u^{\ell,h}_A$ the unit-norm
probe direction at that (layer, head), $\sigma^{\ell,h}_A$ the
training-set standard deviation of the scalar projection
$u^{\ell,h}_A \cdot \mathbf{z}_{\ell,h}(s_A^{+})$, and
$\alpha \in \{-10, 0, +10\}$.  We steer the top-$56$
per-(layer, head) cells ranked by held-out probe accuracy
(cells selected by $\operatorname{argsort}(\text{val\_acc})$
descending), the same selection used by the canonical
probe-extraction sweep.  We do \emph{not} normalise by sequence
position or apply position-specific scaling.

\subsection{Projection-removal}\label{app:proj}
At each (layer, head) we subtract
$c\,\sigma^{\ell,h}_A\,(u^{\ell,h}_A \cdot \mathbf{z}_{\ell,h})\,u^{\ell,h}_A$
from $\mathbf{z}_{\ell,h}$ at the last-token position
(Eq.~\ref{eq:proj-removal}), with $c \in \{0.5, 1, 2, 5\}$.
$\sigma^{\ell,h}_A$ is the per-(layer, head) standard deviation
of $u^{\ell,h}_A \cdot \mathbf{z}_{\ell,h}(s_A^{+})$ across the
unified-partition training set.  Under projection-removal
LLaVA-NeXT occasionally appends commentary tokens after the
JSON answer; we extract the answer with a regex on the
\emph{yes/no} field in addition to the canonical JSON-parse
path.  Recomputed accuracies match the raw answer-token
distribution and are reported in \S\ref{sec:f4}.

\subsection{Temporal probe accuracy (Qwen2-VL)}\label{app:temporal-probe}
We report linear-probe held-out accuracy at four trajectory
fractions $t\in\{0.25, 0.5, 0.75, 1.0\}$ on Qwen2-VL, for the
three cooperative-ToM axes (\autoref{tab:temporal_probe}).
Peak (best layer/head) accuracy is saturated at $100\%$ from the
earliest timestep; mean accuracy varies by axis but is stable to
within $\sim 1$\,pp.  The representation is formed early.
We restrict the temporal sweep to Qwen2-VL on compute-budget
grounds (each timestep is a full per-(layer, head) probe-extraction
pass over $\nTotalFour$ scenarios); we treat it as a single-VLM
ablation, not as a generality claim.
\tabTemporalProbe

\subsection{Pre-registration record}\label{app:prereg}
We pre-registered the T-A falsifier before running the V0--V11
sweep: a cell passes only if both TB and FB accuracy exceed
$50\%$ on a lever-holder.  The hypothesis space (12 variants
$\times$ 4 VLMs $\times$ 4 protagonist conditions) was fixed
before any data was collected.

\subsection{T-A prompt variants}\label{app:ta-variants}
\autoref{tab:ta_variants} summarises the $12$ prompt variants
used in the T-A pre-registered sweep, grouped into four families
(cost framing, chain-of-thought, perception-first / rule-based,
in-context exemplars).  Each variant prepends a prefix to the
canonical intent question; the verbatim prefix text for every
variant is in \S\ref{app:ta-prompts} below.
\tabTAVariants

\subsection{T-A prompt prefixes (verbatim)}\label{app:ta-prompts}
We reproduce here the exact prefix appended before the canonical
intent question for each T-A variant.  Variants whose prefix
differs by protagonist role list both lever-holder and traveller
text.

\paragraph{V0 (baseline).}
No prefix.  The canonical intent question is asked unchanged.

\paragraph{V1 (cost: implicit consequences).}
{\small\begin{quote}\itshape
\textbf{Lever-holder:} Reminder of cooperative consequences: if
the lever-holder releases while the traveller has not yet crossed
the gate, the gate closes on the traveller and the team fails the
task. If the lever-holder keeps holding after the traveller has
crossed, the team is delayed but does not fail.

\textbf{Traveller:} Reminder of cooperative consequences: if the
traveller continues toward a closed gate, it gets stuck and the
team fails. If the traveller waits when the gate is open, the
team is delayed but does not fail.
\end{quote}}

\paragraph{V2 (cost: explicit numerical).}
{\small\begin{quote}\itshape
\textbf{Lever-holder:} Cost framing: action `Yes'
(release/no-longer-hold) $\to$ cost $2.0$ if the traveller has
not yet crossed, else cost $0$. Action `No' (keep holding)
$\to$ cost $1.0$ if the traveller has already crossed, else
cost $0$. Choose the action that minimises expected cost given
what the protagonist can observe.

\textbf{Traveller:} Cost framing: action `Yes' (continue toward
gate) $\to$ cost $2.0$ if the gate is currently closed, else
cost $0$. Action `No' (stop and wait) $\to$ cost $1.0$ always
(delay but no failure). Choose the action that minimises
expected cost given what the protagonist can observe.
\end{quote}}

\paragraph{V3 (utility framing).}
{\small\begin{quote}\itshape
Question framing: what action would MAXIMISE the team's chance
of completing the cooperative task successfully, given what the
protagonist can currently observe?
\end{quote}}

\paragraph{V4 (chain-of-thought).}
{\small\begin{quote}\itshape
Think step by step: (1) FIRST identify what the protagonist can
perceive given the protagonist's position and fog-of-war range.
(2) THEN consider what the optimal action is conditional on what
the protagonist actually knows. (3) FINALLY produce a single Yes
or No answer. The protagonist may be partly ignorant about
events outside its fog. Only choose the answer the protagonist
would optimally pick given its actual perceptual state.
\end{quote}}

\paragraph{V5 (high-stakes safety framing).}
{\small\begin{quote}\itshape
Safety-critical reminder: this is a cooperative team task in
which a wrong action by the lever-holder may TRAP the traveller
permanently (team failure), and a wrong action by the traveller
may STRAND the agent in a closed-gate room (team failure). Only
choose actions that minimise the risk of catastrophic failure
given what the protagonist can actually observe.
\end{quote}}

\paragraph{V6.} V4 chain-of-thought prefix concatenated with the
V2 explicit-cost prefix (lever-holder or traveller, by role).

\paragraph{V7 (perception-first).}
{\small\begin{quote}\itshape
BEFORE answering, identify what the protagonist agent can
actually perceive. The protagonist has a fog-of-war view of
radius $5$ cells from its current position; cells beyond that
radius are NOT visible to the protagonist. The lever-holder's
view of the gate they control depends on the lever's distance
from the gate. If the lever sits next to the gate (Manhattan
distance $\le 5$), the lever-holder CAN see who crosses. If the
lever sits far from the gate (Manhattan distance $>5$), the
lever-holder CANNOT see who crosses. Use only what the
protagonist actually observes to choose the answer.
\end{quote}}

\paragraph{V8 (rule-based).}
{\small\begin{quote}\itshape
Rule: a lever-holder should release its lever ONLY IF it can
directly perceive that the agent it is enabling has finished
crossing the gate. Otherwise it must keep holding. A traveller
should continue toward a gate ONLY IF it can perceive that the
gate is open. Apply this rule given what the protagonist
actually observes (fog-of-war radius $5$).
\end{quote}}

\paragraph{V9.} V7 perception-first prefix concatenated with the
V2 explicit-cost prefix (by role).

\paragraph{V10 (one-shot in-context).}
{\small\begin{quote}\itshape
EXAMPLE: a lever-holder agent's lever sits $2$ cells away from a
gate (within fog-of-war radius $5$). The lever-holder CAN see the
gate. The lever-holder observes the traveller crossing the gate.
Question: should the lever-holder release the lever now? Answer:
Yes, because the lever-holder has perceived the traveller crossing
and the gate is no longer needed.

Now answer the following with the same belief-conditional
reasoning. If the protagonist CAN perceive the relevant event,
recommend the action that responds to that perception. If the
protagonist CANNOT perceive the event, recommend the conservative
action.
\end{quote}}

\paragraph{V11 (two-shot in-context).}
{\small\begin{quote}\itshape
EXAMPLE 1 (TrueBelief case): a lever-holder agent's lever sits
$2$ cells away from a gate (within fog-of-war radius $5$). The
lever-holder CAN see the gate and observes the traveller
crossing. Question: should the lever-holder release? Answer: Yes,
because the protagonist perceived the crossing.

EXAMPLE 2 (FalseBelief case): a lever-holder agent's lever sits
$10$ cells away from a gate (beyond fog-of-war radius $5$). The
lever-holder CANNOT see the gate and does not know whether the
traveller has crossed. Question: should the lever-holder release?
Answer: No, because the protagonist has NOT perceived the
crossing and releasing would risk trapping the traveller.

Now answer the following with the same belief-conditional
reasoning. Distinguish: does the protagonist actually perceive
the event in question, or not? Choose the action that matches
the protagonist's actual perceptual access.
\end{quote}}

\section{Full results tables}\label{app:results}

\subsection{Per-VLM, per-protagonist, per-axis behaviour matrix}\label{app:d2}
We report the full per-cell behavioural accuracies that
underlie the three-way dissociation in F2.  The matrix covers
$4 \text{ VLMs} \times 3 \text{ axes} \times 3 \text{ protag}
\times 2 \text{ outcomes} \times 2 \text{ beliefs}$ = $144$
cells (\autoref{tab:d2_full}).  Readers should look for the
\emph{shape} contrast: belief-Q cells track the world-state
outcome, knowledge-Q cells track TB/FB (perceptual access),
intent-Q cells stay on the per-protagonist prior.
\tabDTwoFull

\subsection{Intent $\Delta_{\rm acc}$ per (VLM, protag, probe)}\label{app:intent-delta}
For every $(\text{VLM} \times \text{protagonist} \times
\text{probe direction})$ cell in the $4 \times 3 \times 3 = 36$-cell
CARD sweep on the intent question, we report best-$\alpha$
accuracy gain over the no-steering baseline
(\autoref{tab:intent_delta}).  This is the load-bearing
evidence for the generalised routing-failure claim in
\S\ref{sec:f3}.
\tabIntentDelta

\subsection{CARD per-cell breakdown ($\Delta_{\rm gap}$ and
  $\Delta_{\rm acc}$)}\label{app:card-breakdown}
For lever-holder $a_1$, we report both the gap shift and the
mean-accuracy shift for every (probe, question, VLM) cell.
Across the full $4 \text{ VLMs} \times 3 \text{ probes} = 12$
intent-question cells, $9$ are exactly $\mathbf{0.00}$ on both
metrics.  The three nonzero cells: Gemma-4 under belief-direction
steering ($|\Delta_{\rm gap}|{=}9.7$\,pp, $|\Delta_{\rm acc}|{=}4.9$\,pp); InternVL2.5 under belief-direction steering
($|\Delta_{\rm gap}|{=}8.3$\,pp, $|\Delta_{\rm acc}|{=}5.6$\,pp);
and InternVL2.5 under knowledge-direction steering
($|\Delta_{\rm gap}|{=}25.0$\,pp, $|\Delta_{\rm acc}|{=}13.9$\,pp,
direction-balanced flips that cancel in net accuracy).  This rules out both belief-conditional modulation and uniform answer-flip
across the matrix.
\tabCardBreakdown

\subsection{Per-protagonist CARD breakdown}\label{app:card-per-agent}
Figure~\ref{fig:card_per_agent} breaks down the protagonist-mean
of \autoref{fig:card_dual} into a $4 \times 3$ grid over (VLM,
protagonist).  The intent column $\Delta$ is essentially $0$ in
every cell, confirming that the routing failure is not
protagonist-specific.

\figCardPerAgent

\subsection{CIs and Wilcoxon sign test on the intent null}\label{app:stats}
Across all $\cardSignTestIntentNTotal$ intent measurements (4
VLMs $\times$ 3 protagonists $\times$ 3 probes $\times$ 3
seeds), steering improves intent accuracy by at most
$\cardSignTestIntentMaxDeltaHi$\,pp ($95\%$ paired-bootstrap
upper bound), vs $\cardSignTestBeliefMaxDeltaHi$\,pp on the
belief control.  The intent answer remains invariant to
$\alpha$ in
$\cardSignTestIntentNZero/\cardSignTestIntentNTotal$ of these;
the remainder skews toward the constant-prior default
($\cardSignTestIntentNNeg$ vs $\cardSignTestIntentNPos$,
Wilcoxon $p = \cardSignTestIntentWilcoxonP$) -- steering pushes
intent toward its default, not toward belief-conditional
behaviour.

\subsection{CARD $3\!\times\!3$ heatmaps per VLM}\label{app:card}
For each VLM we show one panel per protagonist; each panel is a
$3{\times}3$ grid in which rows are the direction we steered
along (belief / intent / knowledge) and columns are the question
we then asked.  The cell number is how much the TB$-$FB gap
\emph{moves} between $\alpha{=}{+}10$ and $\alpha{=}{-}10$, in
pp.

A large cell value means the gap moved
a lot -- not that the model got more accurate: a TB-correct /
FB-wrong pattern can flip to TB-wrong / FB-correct under
steering, giving a large gap shift with mean accuracy unchanged.
The intent column (red box) is the case we care about: even
where it lights up, the companion accuracy-improvement matrix
(\autoref{tab:intent_delta}) is essentially zero on every
cell.  The largest values sit on the diagonal (steer-belief
$\to$ ask-belief, etc.) -- these are the expected positive
controls, where steering an axis strongly moves its own answer
head.

Figures~\ref{fig:card_qwen2vl}--\ref{fig:card_internvl_2_5_8b}
show all four VLMs.
\cardAppendixFig{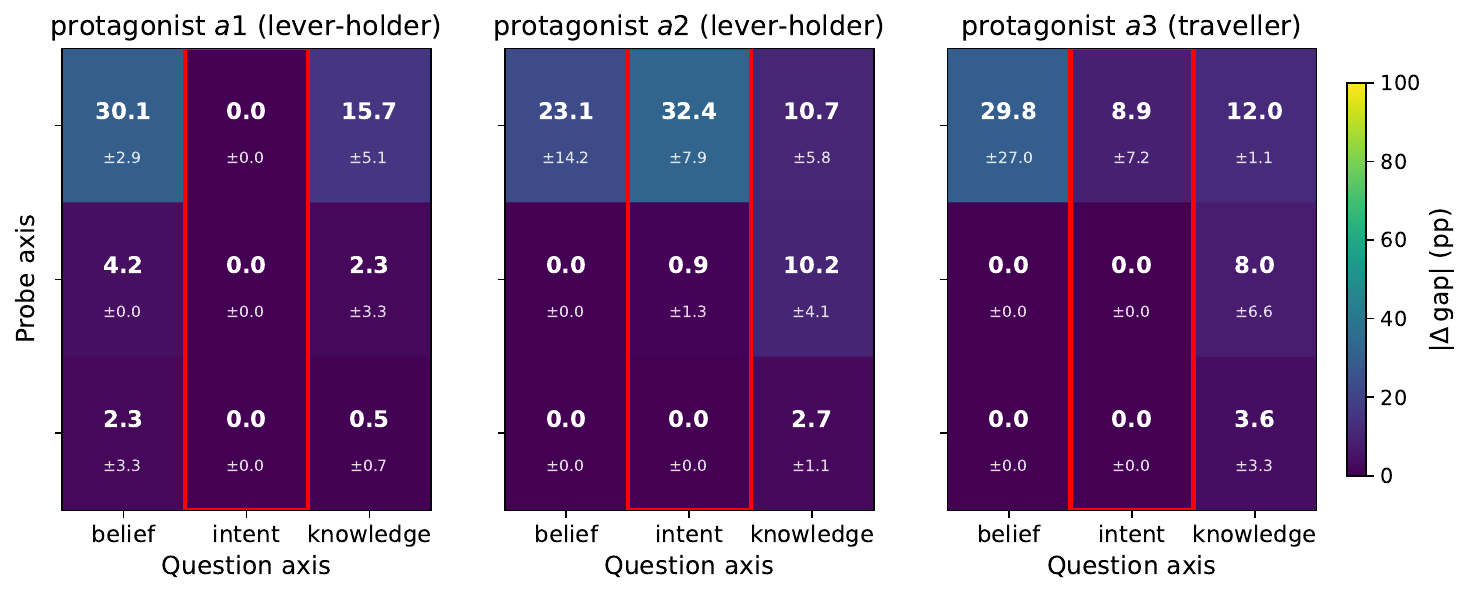}{Qwen2-VL-7B}
\cardAppendixFig{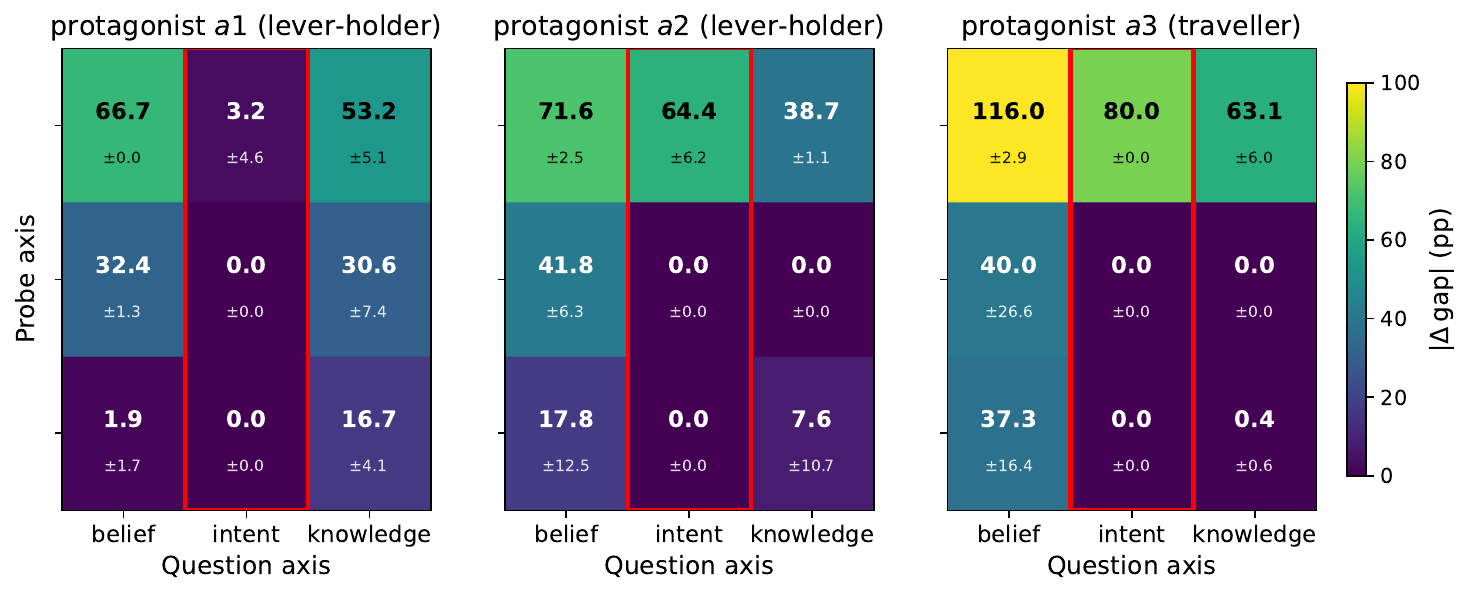}{Gemma-4-E4B}
\cardAppendixFig{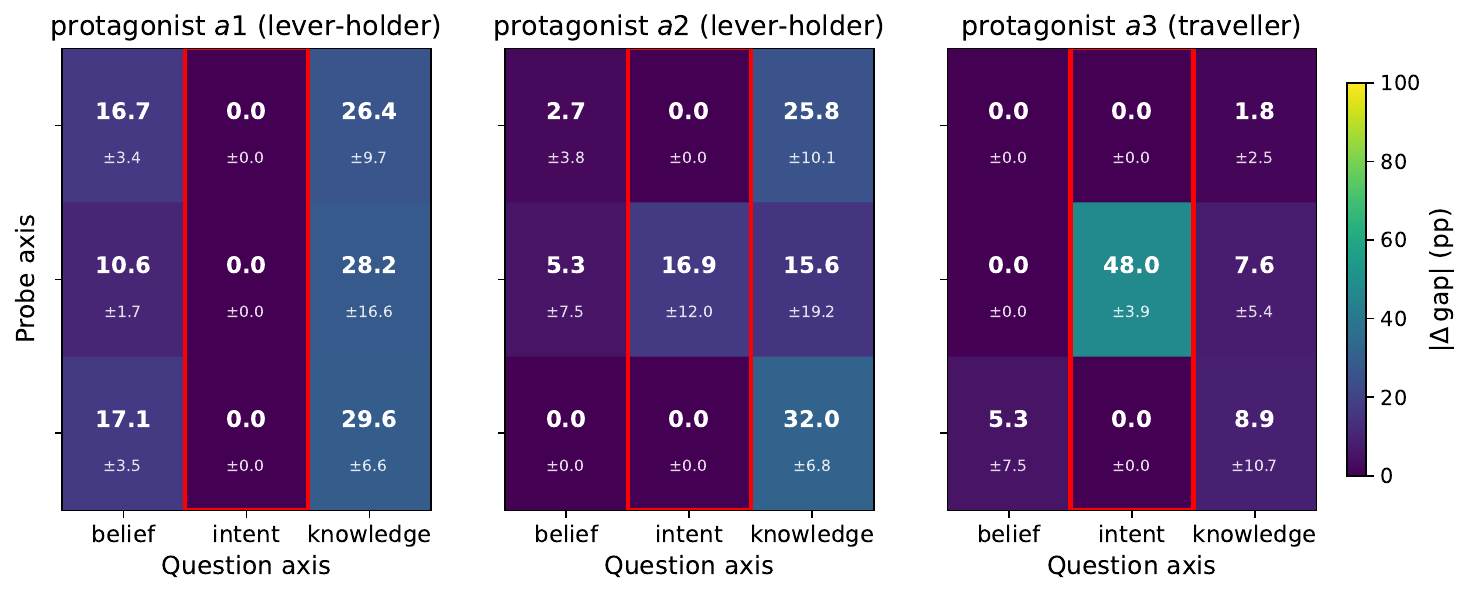}{LLaVA-NeXT-Video-7B}
\cardAppendixFig{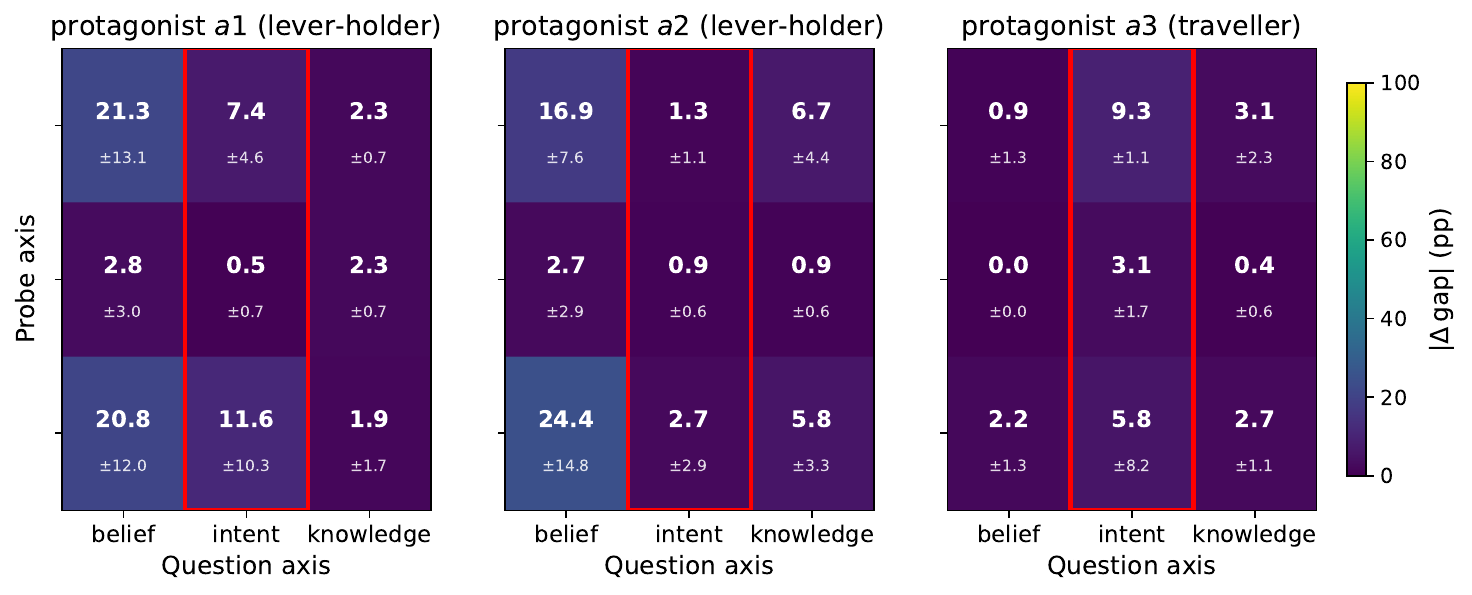}{InternVL2.5-8B}

\subsection{T-A $\taTotalCells$-cell heatmap}\label{app:ta}
For each (variant V0--V11, VLM, protag $\in
\{\text{overall, a1, a2, a3}\}$) we report TB accuracy, FB
accuracy, and their gap (Figure~\ref{fig:ta_heatmap}).  Readers
should look for cells in which a lever-holder column exceeds
the pre-registered $\{\text{TB}>50, \text{FB}>50\}$ threshold --
no cell does.
\figTAHeatmap

\subsection{Projection-removal per-VLM sweep}\label{app:proj-sweep}
For each VLM and each $\alpha \in \{0.5, 1, 2, 5\}\sigma$ we
report the intent TB$-$FB gap on the unified partition
(Figure~\ref{fig:proj_sweep}).  The flat curves -- nearly
horizontal across $\alpha$ on every VLM -- are the visual form
of the A1 amplitude-invariance argument.
\projSweepFig

\subsection{Vendor architecture and rigidity ordering}\label{app:vendor-specs}
The four VLMs differ in text-decoder architecture and in their
empirical resistance to prompt perturbation
(\autoref{tab:vendor_specs}); the more rigid decoders are the
ones whose intent answers were hardest to move under any T-A
prompt in our sweep.
\tabVendorSpecs

\section{Additional controls and diagnostics}\label{app:methods-notes}

\subsection{Off-diagonal cosine measurements}\label{app:cos}
Probe directions on the top-$56$ (layer, head) cells are
moderately aligned, not orthogonal.  On the three steered axes
(belief, intent, knowledge), pairwise cosines fall in
$\meanOffDiagCos$ across the four VLMs.  The belief--intent
cosine is $\cosBeliefIntentQwen$ (Qwen2-VL),
$\cosBeliefIntentGemma$ (Gemma-4), $\cosBeliefIntentLlava$
(LLaVA-NeXT), and $\cosBeliefIntentIntern$ (InternVL2.5).

This non-orthogonality is what makes the intent column's null
informative.  Steering along belief is partially also steering
along intent, yet the intent prediction does not move.  If the
two directions were nearly orthogonal, a null cross-axis effect
on intent would be trivial; at the observed alignment, the null
reflects that the intent prediction does not consume the shared
subspace.  The non-null sibling columns we observe in CARD
(\S\ref{sec:f3}) confirm that the linear belief direction
\emph{is} causally usable by the belief and knowledge
predictions, ruling out an ``information-not-present''
interpretation.

\subsection{CLEVR positive-control breakdown}\label{app:clevr}
\autoref{tab:clevr_card} reports the full per-(VLM, question,
probe) accuracy-gain matrix for the CLEVR positive control
(\S\ref{sec:a3}).  The sweep covers $100$ scenarios per axis and
three probe axes (count, shape, colour) on five VLMs: the four
Relay Chain backbones plus SmolVLM-500M (Idefics-3 family).  We
added SmolVLM-500M explicitly because Qwen2-VL and InternVL2.5
sit at ceiling on shape and colour, and we wanted a backbone
with abundant headroom on every axis.
CLEVR's True/False variants share the same correct answer by
construction, so the gap-shift metric is uninformative here
($\Delta_{\rm gap}=0$ in every cell).  The accuracy-shift metric
is the right summary: it shows that the steering hook moves
accuracy by up to $+60$\,pp when the model has headroom.
\tabClevrCard

\subsection{Random-label probe control}\label{app:random-label}
Following \citet{hewitt2019designing} and
\citet{bortoletto2024brittle}, we retrain logistic-regression probes
on the same paired-statement activations with \emph{randomly
permuted} TB/FB labels.  If the original probe is exploiting
spurious correlations rather than capturing a structured belief
representation, the permuted-label probe should also achieve
high accuracy.  Across all (VLM, axis) combinations
(\autoref{tab:random_label}), random-label probe accuracy is at
chance ($47.4$--$52.0\%$) while true-label probe accuracy is
$91.6$--$100\%$, giving selectivity $\ge 43.0$\,pp on every cell.
The probes are not exploiting superficial structure.
\tabRandomLabel

\subsection{Lexical-ablation control}\label{app:lex}
A linear probe trained on paired-statement activations could in
principle latch onto three asymmetries that co-vary with belief:
a \emph{lexical} ignorance clause inserted into FB belief-true
captions, a \emph{visual} contrast between protagonist-POV
(fog-occluded) and oracle frames, and a \emph{distributional}
gap between those two view sets.  We re-train probes on three
controlled ablation arms that selectively remove these
confounds:
\begin{description}
\item[Arm A (symmetric clause).] We add a counterfactual
``able-to-see'' clause to FB belief-false captions, eliminating
the lexical asymmetry while leaving views unchanged.
\item[Arm B (oracle-only).] We use the oracle view and the
original caption for both belief-true and belief-false
activations, removing both lexical and visual asymmetries so
the probe must rely on the belief-true vs.\ belief-false text
contrast alone.
\item[Arm C (clause artefact).] We inject the ignorance clause
into TB captions only (no real perspective shift), measuring
how much val accuracy a probe can achieve from the clause alone.
\end{description}
On Arm B the top-cell probe accuracy stays within $\pm 2$\,pp
of the canonical extraction across all VLMs, so the probe is
not riding the lexical or view confound.  Arm C accuracy stays
near chance, ruling out the clause-alone exploit.

% \subsection{Distributed-token steering control (CS-CARD-T)}\label{app:cs-card-t}
% We test whether routing operates outside the last-token
% position our hook targets.  On Qwen2-VL we re-run
% CARD with the steering hook fired at three additional positions
% (\textit{vision-end}, \textit{frame-4-end}, and a
% \textit{multi-position late} slice covering all three) in
% addition to the \textit{last-token} target, using
% four probe sources: the canonical paired-statement belief and
% intent probes, plus two per-position condition probes trained
% at \textit{vision-end} and \textit{frame-4-end} via the
% per-image-token pipeline.  Sweeping $\alpha \in \{-10, +10\}$
% yields $32$ (probe-source $\times$ target-position) cells.
% Across all $32$ cells, the maximum intent
% $|\Delta_{\rm acc}|$ vs the $\alpha{=}0$ baseline is
% $0.3$\,pp; no cell shifts by more than $1$\,pp
% (\autoref{fig:cscardt}).  The intent null is therefore not an
% artefact of the last-token hook position.

\begin{figure}[!h]
  \centering
  \includegraphics[width=0.95\columnwidth]{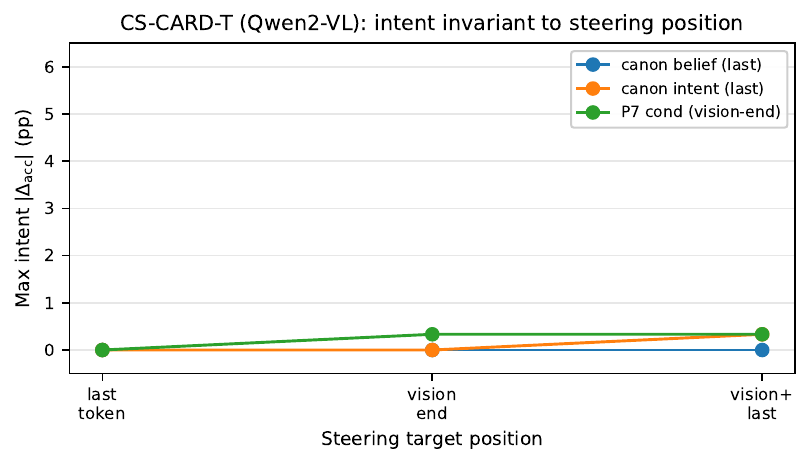}
  \caption{CS-CARD-T on Qwen2-VL: intent $|\Delta_{\rm acc}|$
    vs steering target position, one line per probe source.
    All nine (source, position) entries sit at $\le 0.3$\,pp.
    \label{fig:cscardt}}
\end{figure}

\subsection{Polarity-flip diagnostic}\label{app:d1}
We ran a separate diagnostic in which the question polarity is
flipped (``should the protagonist \emph{not} release?'' instead
of ``release?'').  An oracle reasoner produces the same TB$-$FB
gap shape under the flip; a constant prior collapses.  We
report all (VLM, protagonist) cells in
\autoref{tab:d1_polarity}; the contrast between Qwen $a_1$
(preserves the $100{\to}100$ pattern -- oracle signature) and
LLaVA $a_2$ (collapses -- constant-prior signature) is
illustrative of the two regimes observed across the table.
\tabDOneFull

\subsection{Scale ablation: Qwen2.5-VL-32B}\label{app:scale-32b}
To check that the intent-column null is not specific to the
$7$--$8$B Relay Chain backbones, we ran the same CARD pipeline
on Qwen2.5-VL-32B ($\approx 32$B parameters; $64$
layers, $40$ heads, head\_dim $128$): identical probe
extraction, top-$56$ steering pool, and
$\alpha \in \{-10, 0, +10\}$ sweep across the three probes.
Peak held-out probe accuracy is $98.2\%$ (belief), $95.9\%$
(intent), $97.5\%$ (knowledge) -- on par with or above the
$7$--$8$B backbones.  On lever-operator $a_1$ the intent answer
is invariant to steering across all nine (probe, $\alpha$)
combinations: TB accuracy is $30.6\%$ and FB accuracy is
$100\%$ at every $\alpha$ on every probe direction -- the same
constant-prior signature observed at $7$--$8$B.  The belief and
knowledge questions move under the same hooks (e.g.\ belief-Q
FB accuracy on $a_3$ shifts from $96.7\%$ at $\alpha{=}{-}10$ to
$98.7\%$ at $\alpha{=}{+}10$ on the belief probe).  The routing
failure does not vanish at $4\times$ parameter scale, even
though the probe recovers the belief direction at ceiling.

\section{Models, infrastructure, and computational budget}\label{app:compute}

\paragraph{Models and parameter counts.}
The four open-weight VLMs used in the main Relay Chain
experiments span three vision-encoder lineages and three
language-decoder lineages at the $4$--$8$B scale:
Qwen2-VL-7B-Instruct ($\approx 7.6$B parameters;
\citealp{Yang2024Qwen2TR}),
Gemma-4-E4B-it (effective $\approx 4$B activated
parameters in the E$N$B mixture-of-experts variant;
\citealp{gemma4_2026}),
LLaVA-NeXT-Video-7B-hf ($\approx 7.1$B;
\citealp{zhang2024llavanextvideo}), and
InternVL2.5-8B ($\approx 8.1$B;
\citealp{chen2024expanding}).
The CLEVR positive control additionally uses
SmolVLM-500M ($\approx 500$M parameters, Idefics-3
family) \citealp{marafioti2025smolvlm}. We also use Qwen2.5VL-32B-Instruct (\citealp{Yang2024Qwen25TR}) for a model scale ablation study on CARD.

\paragraph{Hardware.}
All experiments ran on NVIDIA H100 NVL GPUs
($96$\,GB HBM) on an internal SLURM-managed cluster.  Each CARD,
projection-removal, or T-A sweep occupies one GPU; multi-seed
runs and the CLEVR chain were parallelised across up to four
GPUs.  Probe training is CPU-bound (logistic regression on
saved activations).

\paragraph{Compute budget.}
End-to-end cost for the full pipeline -- activation extraction,
probe training, the CARD $3{\times}3$ sweep, and all ablations
across the four open-weight VLMs -- is approximately $25$
GPU-hours on a single H100 NVL.  In practice we parallelised
across two H100 GPUs, so wall-clock time was about half a day.

\paragraph{Software.}
VLM inference uses \texttt{transformers}~v5.5 (Hugging Face)
with \texttt{bfloat16} weights and either \texttt{flash-attention-2}
or \texttt{sdpa} as the attention backend.  Logistic-regression
probes are trained with \texttt{scikit-learn} ($L^2$
regularisation, default solver) on saved activations.
Paired bootstrap CIs, the binomial sign test, and the Wilcoxon
signed-rank test are computed with \texttt{scipy.stats}; the
bootstrap RNG is fixed at seed $20260523$ for reproducibility.

\subsection{Artifact licenses and intended use}\label{app:licenses}
All four open-weight VLMs we evaluate are publicly released:
Qwen2-VL-7B under Apache 2.0, Gemma-4-E4B-it under the Gemma
License Agreement, LLaVA-NeXT-Video-7B under Apache 2.0, and
InternVL2.5-8B under MIT.  CLEVR~\citep{johnson2017clevr},
used as the positive-control auxiliary dataset, is distributed
under CC~BY~4.0.  GridToM~\citep{li2025black}, cited as the
grid-world ToM format inspiration, is released by its authors
for academic use.  We use all models and datasets for
non-commercial academic research, consistent with their
intended use.  Our CARD code and the Relay Chain dataset will
be released under MIT (code) and CC~BY~4.0 (data) upon
acceptance.

\end{document}